\documentclass{article}
\usepackage{jfrExamplee}
\usepackage[utf8]{inputenc}
\usepackage[T1]{fontenc}
\usepackage{lmodern}
\usepackage[hidelinks]{hyperref}
\usepackage{graphicx}
\usepackage[export]{adjustbox}
\usepackage{natbib}
\usepackage{apalike}
\usepackage{setspace}
\usepackage{url}
\usepackage{amsmath,amssymb}
\usepackage{textcomp}
\usepackage{epstopdf}
\usepackage{multicol}
\usepackage{color}
\usepackage{pifont}
\usepackage{balance}
\usepackage{footmisc}
\usepackage{comment}
\usepackage{subcaption}
\usepackage{layouts}
\usepackage{cancel}
\usepackage{bm}
\usepackage{booktabs}
\usepackage{tabularx}
\usepackage{paralist}
\usepackage{xspace}
\usepackage{tikz}
\usepackage{threeparttable}

\graphicspath{{./}}
\newlength{\twosubht}
\newsavebox{\twosubbox}

\newcommand{\ie}{i.e.,\ }
\newcommand{\eg}{e.g.,\ }
\newcommand{\cf}{cf.\xspace}
\newcommand{\reffig}[1]{Fig.~\ref{#1}}
\newcommand{\reftab}[1]{Tab.~\ref{#1}}
\newcommand{\refsec}[1]{Sec.~\ref{#1}}

\DeclareMathOperator{\atantwo}{atan2}
\usepackage{booktabs}
\usepackage{threeparttable}

\usepackage{amsmath}
\usepackage{tensor}

\usepackage{placeins}

\usepackage[hidelinks]{hyperref}
\usepackage[capitalize]{cleveref}

\usepackage[draft]{fixme}
\fxsetup{theme=color, layout={inline,index}}

\title{Target Chase, Wall Building, and Fire Fighting:\\Autonomous UAVs of Team NimbRo at {MBZIRC} 2020}

\author{Marius Beul, Max Schwarz, Jan Quenzel, Malte Splietker, Simon Bultmann,\\
\textbf{Daniel Schleich, Andre Rochow, Dmytro Pavlichenko, Radu Alexandru Rosu, Patrick Lowin,}\\
\textbf{Bruno Scheider, Michael Schreiber, Finn S\"{u}berkr\"{u}b, and Sven Behnke}\\
Autonomous Intelligent Systems Group \\
University of Bonn \\
Bonn, Germany \\
\texttt{mbeul@ais.uni-bonn.de}\\
}

\begin{document}

\maketitle

\begin{tikzpicture}[overlay, remember picture]
  \path (current page.north) ++(0.0,-0.5) node[draw = black] {Accepted for Field Robotics, to appear 2022};
\end{tikzpicture}
\vspace{-0.5cm}

\begin{abstract}
The Mohamed Bin Zayed International Robotics Challenge (MBZIRC) 2020 posed diverse challenges for unmanned aerial vehicles (UAVs). We present our four tailored UAVs, specifically developed for individual aerial-robot tasks of MBZIRC, including custom hardware- and software components. 

In Challenge~1, a target UAV is pursued using a high-efficiency, onboard object detection pipeline to capture a ball from the target UAV. A second UAV uses a similar detection method to find and pop balloons scattered throughout the arena.

For Challenge~2, we demonstrate a larger UAV capable of autonomous aerial manipulation:
Bricks are found and tracked from camera images. Subsequently, they are approached, picked, transported, and placed on a wall.

Finally, in Challenge~3, our UAV autonomously finds fires using LiDAR and thermal cameras. It extinguishes the fires with an onboard fire extinguisher.

While every robot features task-specific subsystems, all UAVs rely on a standard software stack developed for this particular and future competitions. We present our mostly open-source software solutions, including tools for system configuration, monitoring, robust wireless communication, high-level control, and agile trajectory generation.

For solving the MBZIRC 2020 tasks, we advanced the state of the art in multiple research areas like machine vision and trajectory generation. We present our scientific contributions that constitute the foundation for our algorithms and systems and analyze the results from the MBZIRC competition 2020 in Abu Dhabi, where our systems reached second place in the Grand Challenge. Furthermore, we discuss lessons learned from our participation in this complex robotic challenge.
\end{abstract}

\section{Introduction}
\label{sec:Introduction}
The Mohamed Bin Zayed International Robotics Challenge (MBZIRC) aims to advance the state of the art in autonomous mobile robotics through robot competitions. Unique to this challenge is the focus on field robotics and the diversity of tasks that need to be carried out. In the 2nd MBZIRC competition, participants were required to track, follow, and interact with a target unmanned aerial vehicle (UAV), find and pop balloons, build walls using a ground/aerial robot fleet, and finally detect and fight fires in indoor and outdoor environments.

\begin{figure}
  \sbox\twosubbox{%
    \resizebox{\dimexpr.95\textwidth-1em}{!}{%
      \includegraphics[height=3cm,clip,trim=100 000 100 000]{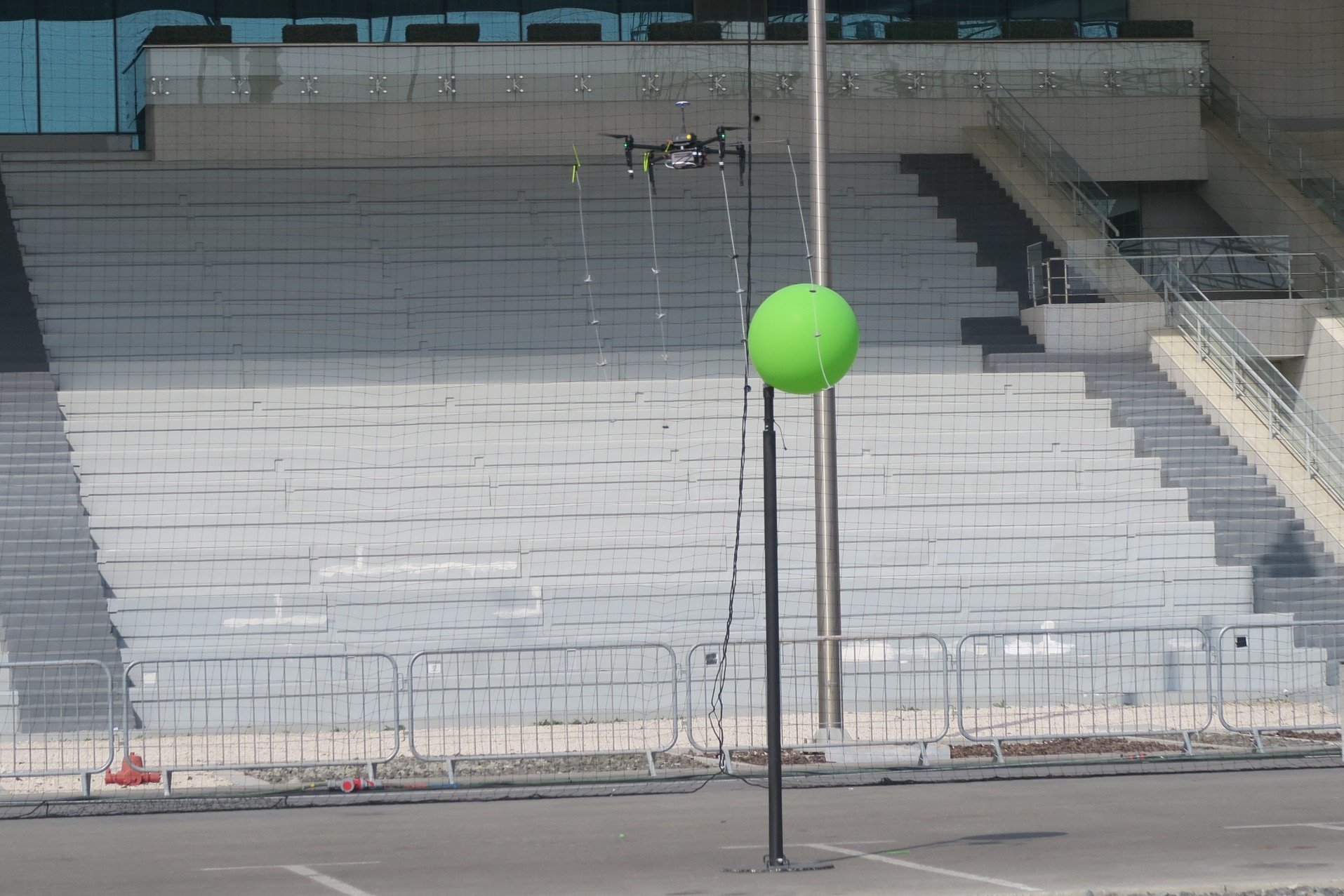}%
      \includegraphics[height=3cm,clip,trim=000 000 030 000]{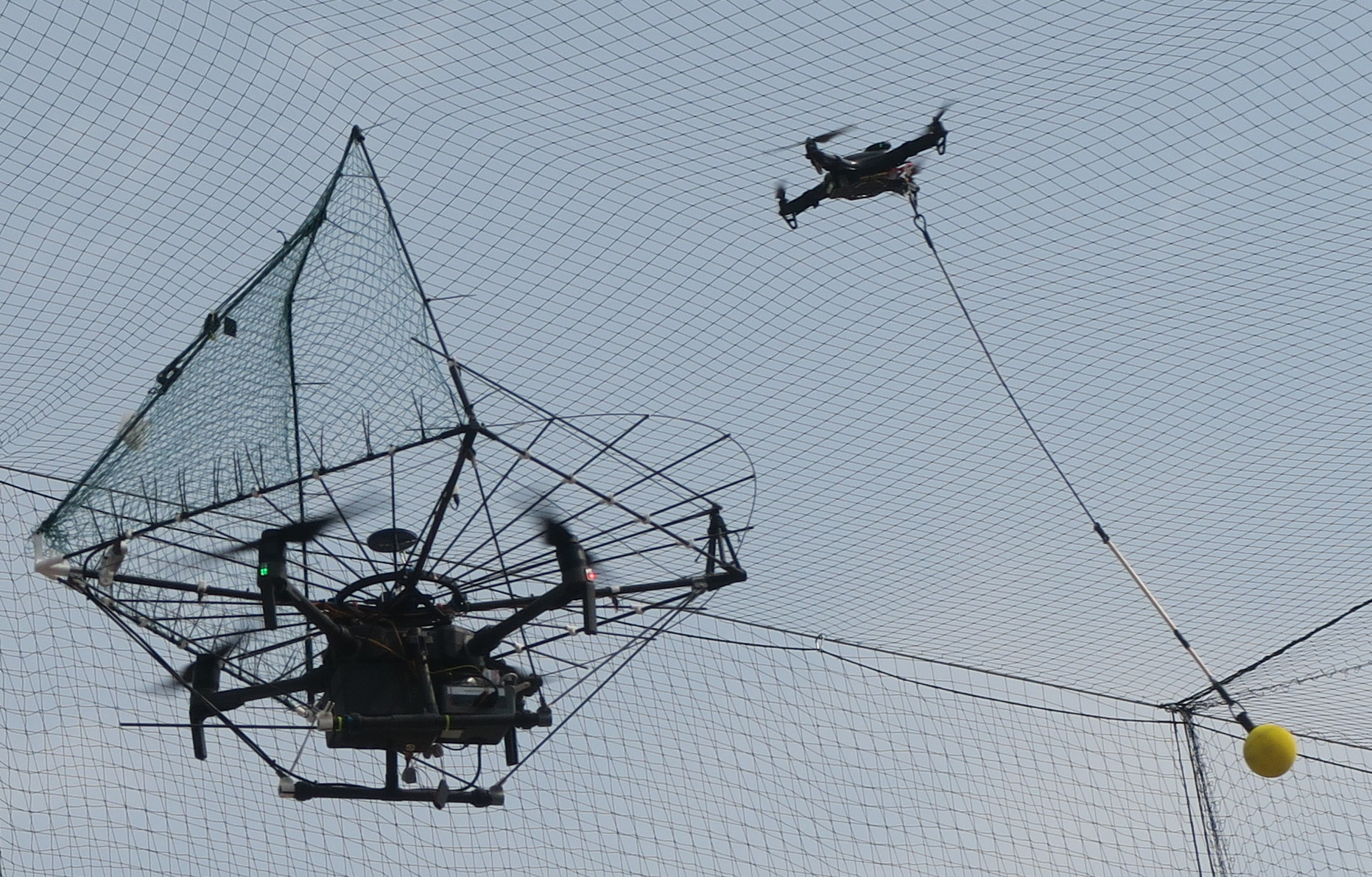}%
      \includegraphics[height=3cm,clip,trim=200 000 100 150]{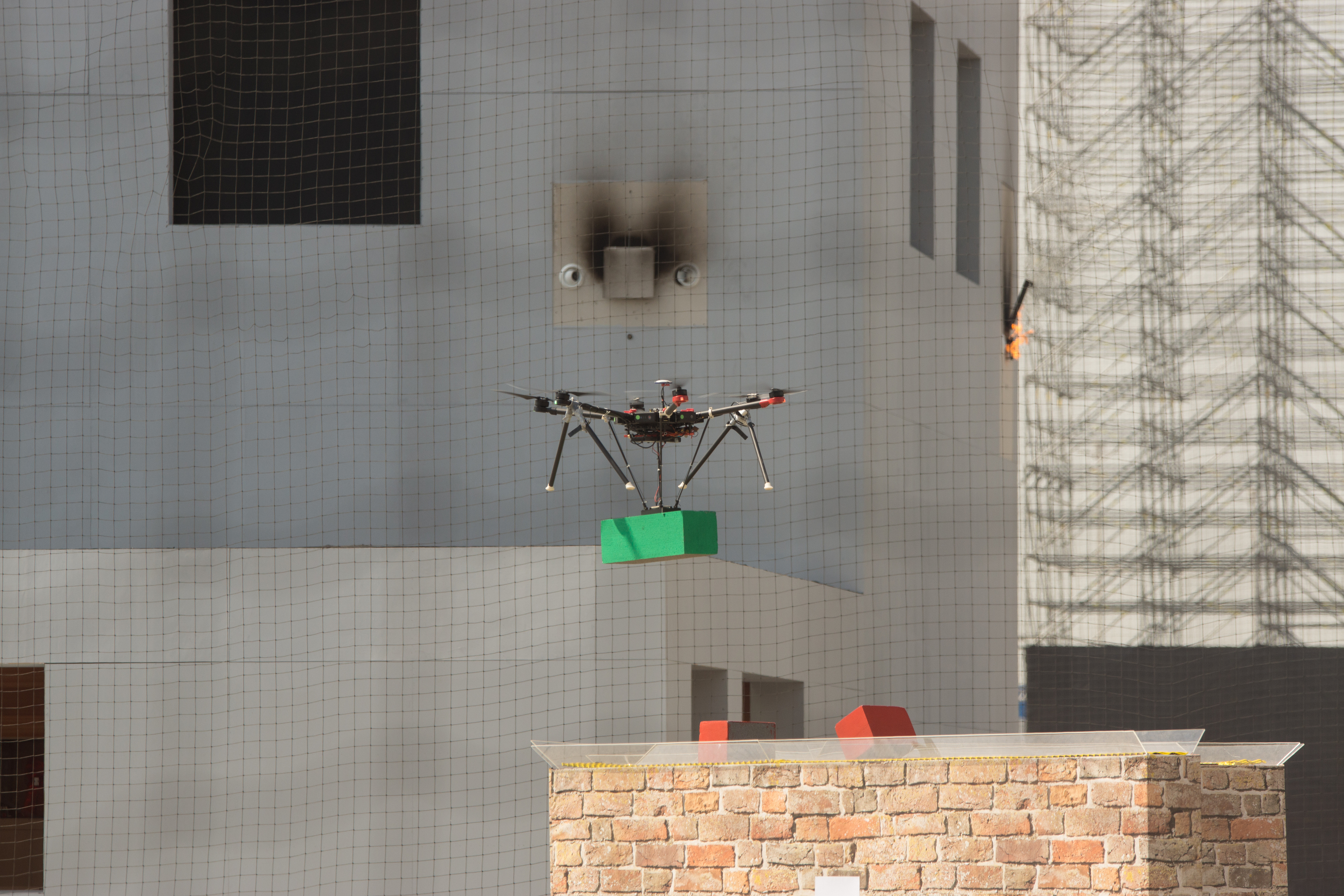}%
      \includegraphics[height=3cm,clip,trim=000 000 100 000]{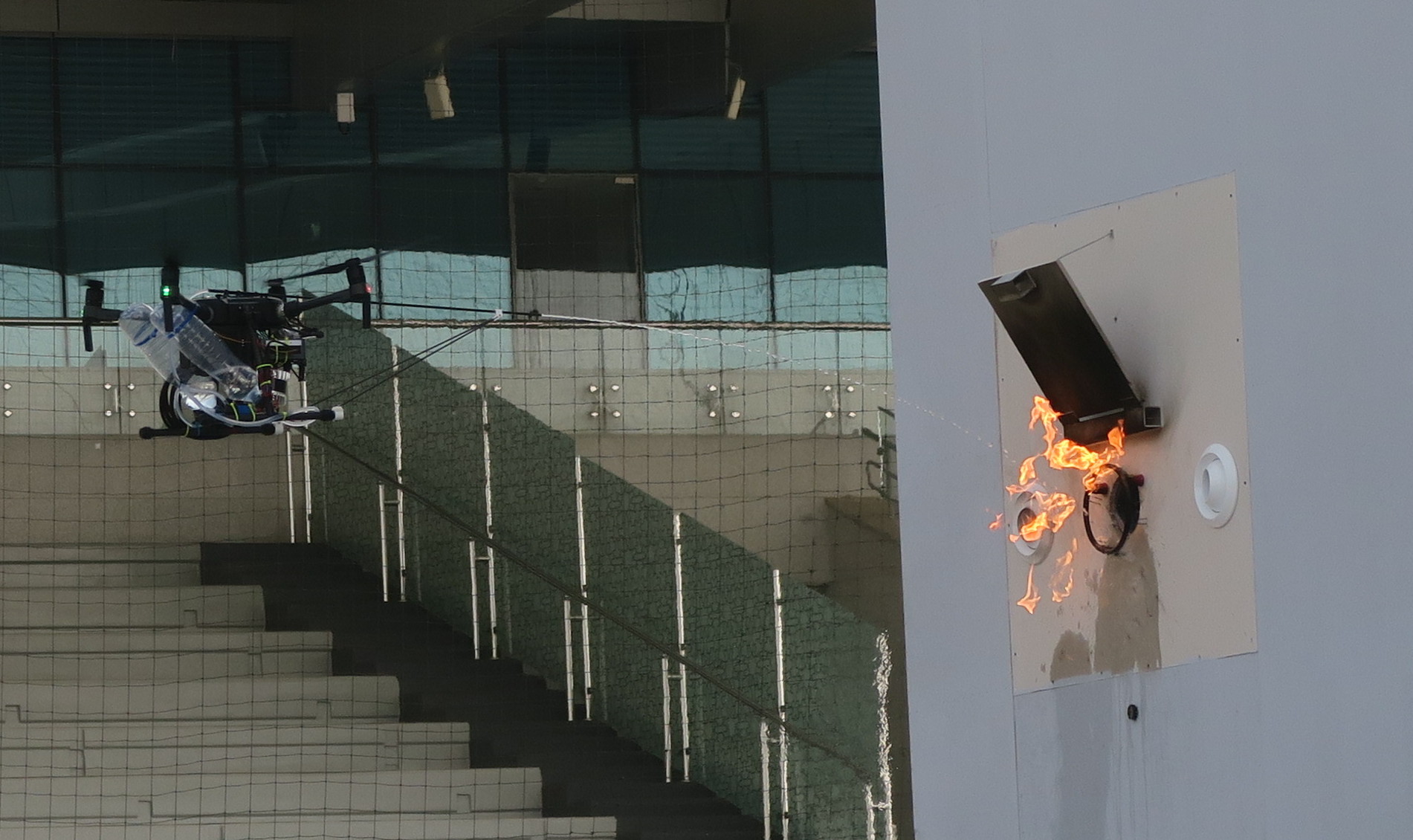}%
    }%
  }
  \setlength{\twosubht}{\ht\twosubbox}
  \centering
  \subcaptionbox{Jelly\label{fig:Jelly}}{%
    \includegraphics[height=\twosubht,clip,trim=100 000 100 000]{IMG5629cut.JPG}%
  }
  \subcaptionbox{Chaser\label{fig:Chaser}}{%
    \includegraphics[height=\twosubht,clip,trim=000 000 030 000]{IMG5642crop.JPG}%
  }
  \subcaptionbox{Lofty\label{fig:Lofty}}{%
    \includegraphics[height=\twosubht,clip,trim=200 000 100 150]{MBZIRC2020137_low.jpg}%
  }
  \subcaptionbox{Splasher\label{fig:Splasher}}{%
    \includegraphics[height=\twosubht,clip,trim=000 000 100 000]{IMG5610.JPG}%
  }
  \caption{Our UAVs in action during the MBZIRC 2020.}
  \label{fig:teaser}
\end{figure}

In this article, we will describe the approach of Team NimbRo to the aerial-robot tasks of the MBZIRC 2020, detailing the fleet of UAVs that was developed (see \reffig{fig:teaser}). We will focus on the three sub-challenges individually, highlighting the specific solutions found for the complex problems they posed, but also discuss our overall approach and identify lessons learned from our successful participation in the competition---Team NimbRo reached second place in Challenge~2 (wall building) and second place in the combined Grand Challenge.

In addition to the overall system integration, our contributions include:
\begin{compactitem}
 \item Custom-tailored UAV hardware designs for the individual challenges,
 \item a common \emph{software stack} for mission control and system monitoring, including a hardened FSM framework with accompanying design and visualization tools, robust network communication and remote supervision tools, as well as a universal model predictive controller,
 \item real-time \emph{onboard} target perception and trajectory control for Challenge~1 (balloon hunt),
 \item onboard object detection, 6D pose estimation, and aerial manipulation control for Challenge~2 (wall building), and
 \item fully \emph{autonomous} fire detection and extinguishing for Challenge~3 (fire fighting).
\end{compactitem}

Many of the employed algorithms are based on state-of-the-art techniques, and their development advanced their field significantly. Since detailing every approach would go beyond the scope of this article, we highlight key aspects of our methods and refer to corresponding papers where applicable.

We evaluate our approach with real-robot experiments and report results from the MBZIRC 2020 competition.
A description of our unmanned ground vehicle (UGV) approaches for MBZIRC 2020 can be found in \citet{lenz2021jfr}.

\section{Related Work}
\label{sec:Related_Work}

To our knowledge, no comprehensive analysis or review of the MBZIRC 2020 competition has yet been published. Still, many of the sub-problems that appeared in the individual challenges are highly active research areas, and we provide an overview of the current state of the art.

\textbf{Aerial Search \& Rescue:}
\citet{https://doi.org/10.1002/rob.21436} report on field experiments with UAVs and UGVs inside earthquake-damaged buildings. The authors provide details about their collaborative mapping approach and report results from the experiments in the form of maps generated by the individual robots and as a team.
\citet{RoboCup_2019} use a convolutional neural network (CNN) to detect circular and other pre-trained objects in real time. Circular objects can be easily detected and differentiated from non-circular ones based on the shape of their contour. \citet{object_contour_2016} propose an encoder-decoder structure for contour detection of generic foreground objects from the PascalVOC dataset~\citep{pascalvoc_2010}. \citet{conv_boundaries_2018} use a CNN architecture to detect object contours at multiple scales together with their orientations, based on a generic backbone CNN, like ResNet~\citep{he_deep_2016}. We follow a similar approach but train our contour detection network to detect only contours of one class of objects---the balloons (see~\refsec{sec:Balloon_Perception}).

Lightweight computer vision models that can be executed efficiently also on mobile or embedded systems with limited computational power have been of increasing research interest during recent years. The MobileNet architectures \citep{mobilenet_2017,mobilenetv2_2018}, for example, significantly reduce the number of parameters in a CNN by replacing standard convolutions with depthwise-separable convolutions. For our vision system, we employ a standard ResNet architecture but with very few layers (\cf\refsec{sec:Balloon_Perception}), keeping the number of parameters and the necessary computational power low. Furthermore, specialized inference accelerators like the Google Edge TPU can be used for efficient processing with a limited size and energy budget. To make a trained CNN model compatible with the Edge TPU, weights and activations need to be quantized to 8-bit integer values, \eg using the quantization scheme described by \citet{quantization_2018}.

Also, fast real-time trajectory generation and control is an active area of research. Specifically, as a result of the MBZIRC 2017, various groups presented advanced control approaches for UAVs.
\citet{Baca2017} report the approach to landing on a moving platform during the MBZIRC 2017, employed by their team including the CTU Prague, UPenn and UoL.

Similarly, \citet{Cantelli2017} and \citet{Battiato2017} from the University of Catania report their systems, including their control approach. \citet{Falanga_SSRR2017} from the University of Z\"{u}rich plan jerk-minimizing trajectories using a fast analytic polynomial generation method similar to ours. Outside of the MBZIRC, many groups employ polynomial trajectories for UAV control. For a comparison of these approaches, see the work of \citet{Ezair2014}.

\textbf{Aerial Manipulation:} Aerial manipulation has been investigated for some time \citep{ruggiero2018aerial}.
Complex systems with fully actuated multi-DoF robotic arms have been built \citep{huber2013first,kim2013aerial}.
\cite{Cooperative_Manipulation} as well as \cite{9124698} demonstrated carrying large items with multiple UAVs.
\citet{lindsey2012construction} demonstrated the assembly of structures with teams of small UAVs. This work relied on an external motion capture system and self-locking magnetic part connectors.
\citet{goessens2018feasibility} present a feasibility study of constructing real-scale structures with UAVs based on self-aligning Lego-like brick shapes.

\citet{baca2020arxiv} present the UAV system for the MBZIRC 2020 wall building task of the team from CTU Prague, UPenn, and NYU. The team won the Wall-building as well as the Grand Challenge.
Their approach for the treasure-hunt challenge in MBZIRC 2017 is described in~\citep{8276269}.

\citet{real2021access} demonstrates the approach from the University of Seville (GRVC) together with Tecnico Lisboa and CATEC. The authors placed 14nd in the Wall-building challenge and 4th in the Grand Challenge, together with Tecnico Lisboa.

A predecessor of our work is Challenge~3 of the MBZIRC 2017, where a team of UAVs was supposed to collect discs. Our entry \citep{beul2019team} was quite successful and reached third place in this challenge, behind ETH Z\"{u}rich \citep{bahnemann2019eth} and CTU Prague, UPenn, and UoL \citep{spurny2019cooperative}.

In contrast, the 2020 edition of the MBZIRC featured much heavier and larger objects, which could only be grasped on a specific spot and had to be placed in a specified pose. To this end, we designed a magnetic gripper that is guided using visual servoing and has five passive DoFs that allow flexibility during grasping but facilitate precise placement.

\Citet{krizmancic2020cooperative} describe a planning system for UAV-UGV cooperative wall building. Their planner allows exploitation of the unique characteristics of each platform. In contrast, our UAV for wall building uses a simple greedy behavior for selecting the next task---which was sufficient for MBZIRC 2020 since we only had one UAV and the UAV/UGV walls were separate.

\textbf{Fire Fighting:} Cooperative monitoring and detection of forest and wildfires with autonomous teams of UAVs \citep{BailonRuiz2020icuas} or UGVs \citep{Ghamry2016icuas} gained significant attention in recent years \citep{Delmerico2019jfr}.
While UGVs are able to carry larger amounts of extinguishing agents or drag a fire hose \citep{Liu2016icac}, payload limitations impede the utility of UAVs.
\citet{Aydin2019drones} investigated the deployment of fire-extinguishing balls by a UAV.
\citet{Ando2018ral} developed an aerial hose pipe robot using steerable pressurized water jets for robot motion and fire extinguishing.

MBZIRC 2020 competitors \citet{spurny2021access,jindal2021design} from CTU Prague, UPenn and NYU present their fire-fighting system. Their UAV is capable to autonomously approach and extinguish fires in GNSS-denied environments after detecting a suitable building entrance. The authors report results from simulations, field tests, and from the MBZIRC 2020 competition where they (together with UPenn and NYU) placed 4th in the fire-fighting and won the Grand Challenge.

In urban environments, thermal mapping \citep{cho2015survey} is commonly used for building inspection. It relies on simultaneously captured color and thermal images from different poses and employs standard photogrammetry pipelines. \citet{schonauer2013live} provide real-time assistance for firefighters via thermal augmentation of live images within room-scale environments.
In contrast, \citet{borrmann2013thermal} mounted a terrestrial LiDAR, a thermal camera, and a color camera on a UGV to obtain colorized point clouds, enabling automated detection of windows and walls \citep{demisse2015interpreting}.
\citet{fritsche2017fusion} cluster high-temperature points from fused 2D LiDAR and mechanically pivoting radar to detect and localize heat sources.
Similarly, \citet{Rosu2019ssrr} acquire a thermal textured mesh using a UAV equipped with LiDAR and thermal camera and estimate the 3D position and extent of heat sources.

New challenges arise where robots have to operate close to structures. Therefore, UAVs are often equipped with cameras and are remote-controlled by first responders. In contrast, autonomous execution was the goal for Challenge~3 of the MBZIRC 2020.
Team Skyeye \citep{SkyeyeMBZIRC} used DJI Matrice~600 and DJI Matrice~210~v2 UAVs equipped with color and thermal cameras, GPS, and LiDAR. A map is prebuilt from LiDAR, IMU, and GPS data to allow online Monte Carlo localization and path planning with Lazy Theta$^{*}$. Fires are detected via thresholding on thermal images. The fire location is estimated with an information filter from either projected LiDAR range measurements or the map. A water pump for extinguishing is mounted on a pan-tilt unit on the UGV while being fixed in place on the UAV.

Although our general approach is similar to team Skyeye, we rely more heavily upon relative navigation for aiming at the target after initial detection and less on the quality of our map and localization.
In comparison, the protruding nozzle on our UAV allows for a safer distance from fire and walls while aiming.
Furthermore, where distance measurements are unavailable, we use the known heat source size for target localization. We detect the holes of outdoor facade fires in LiDAR measurements and fuse these with thermal detections, which allows us to spray precisely on target, where the surrounding flames would offset thermal-only aiming.

\section{Common Hardware and Software}
\label{sec:common}
Instead of starting development from scratch, we based our UAVs on our systems built for the MBZIRC 2017 competition~\citep{beul2019team}.
We equip commercially available DJI platforms with specific sensors, actuators, and computing power for their particular task. Depending on the required payload, we chose a DJI product of suitable size and lift capability, such as the DJI Matrice~100, DJI Matrice~210~v2, and DJI Matrice~600. Locking into a particular vendor allowed us to share design resources and necessary equipment such as batteries, chargers, and remote controls. Furthermore, our trained safety pilots can switch between the different robots without problems, significantly increasing flexibility during testing and in the actual competition.
For detailed descriptions of the hardware platforms for each challenge, we refer to \cref{sec:Hardware,sec:Moving_Target_Interception,sec:lofty:hw,sec:ch3:hw}.

Our software components are also an evolution of our MBZIRC 2017 entry. We built our software on top of the ROS middleware~\citep{ROS}, the de-facto standard framework for robotic applications. In addition to the wealth of available components in the ROS ecosystem, we developed several essential tools and libraries that allowed us to quickly implement robust software for the competition.

\subsection{Robust Network Communication}
A key component in UAV fleets is communication. Robots may communicate with each other about object detections, goal and task assignments, their current position, and a multitude of other data. Furthermore, since the robots operate under human supervision, they will send monitoring data to the supervisors.

ROS offers built-in network transparency, but it is ill-suited to unreliable wireless networks due to its reliance on TCP and handshaking protocols. Our group has developed the \texttt{nimbro\_network}\footnote{\url{https://github.com/AIS-Bonn/nimbro_network}} package as a robust replacement for ROS network transparency. Our robots are separate ROS systems (a so-called multimaster approach), which allows them to operate even without a network connection. The \texttt{nimbro\_network} stack creates statically-configured connections between the robots, which forward the chosen ROS topic over the network connection. It uses forward error correction (FEC) rather than resending to deal with packet loss, which reduces latency.
All robots stream monitoring data to a central operator station on the ground. The mission specialists then connect to this operator station through Ethernet using standard ROS network transparency. For a detailed list detailing the advantages and disadvantages of \texttt{nimbro\_network}, see the referenced website.

\subsection{Runtime Control and Supervision}
\label{sec:Runtime_Control_and_Supervision}

\begin{figure}
  \sbox\twosubbox{%
    \resizebox{\dimexpr.95\textwidth-1em}{!}{%
      \includegraphics[height=3cm,clip,trim=000 000 250 000]{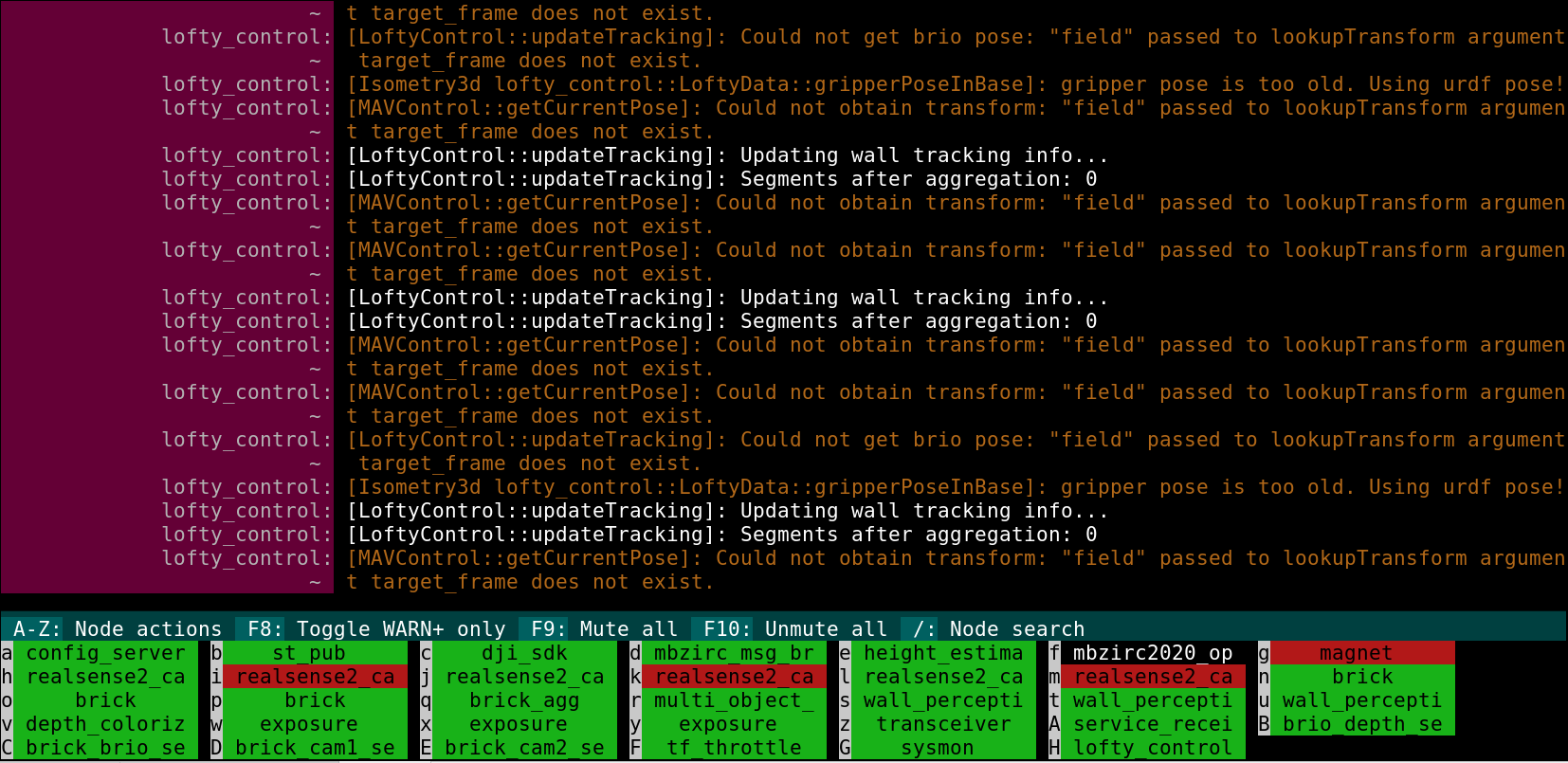}%
      \includegraphics[height=3cm,clip,trim=000 000 000 000]{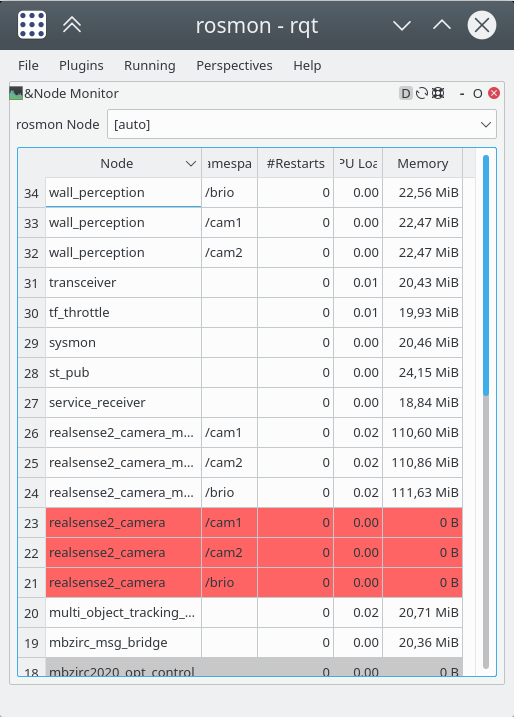}%
      \includegraphics[height=3cm,clip,trim=000 000 000 000]{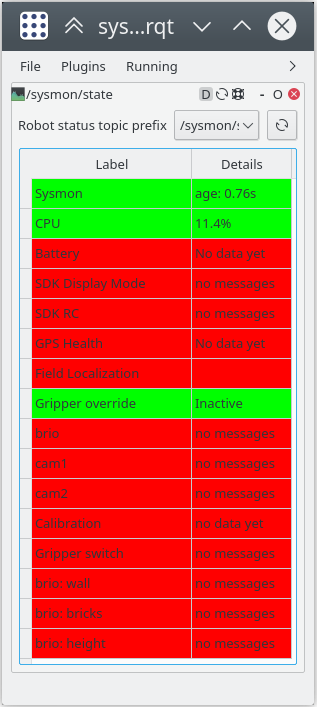}%
    }%
  }
  \setlength{\twosubht}{\ht\twosubbox}
  \centering
  \subcaptionbox{\texttt{rosmon} terminal interface\label{fig:rosmon_terminal}}{%
    \includegraphics[height=\twosubht,clip,trim=000 000 250 000]{rosmontui.png}%
  }
  \subcaptionbox{\texttt{rosmon} GUI with process overview\label{fig:rosmon_GUI}}{%
    \includegraphics[height=\twosubht,clip,trim=000 000 000 000]{rosmongui.png}%
  }
  \subcaptionbox{System monitor GUI with high-level checks\label{fig:System_monitor}}{%
    \includegraphics[height=\twosubht,clip,trim=000 000 000 000]{sysmon.png}%
  }
  \caption{Process and system monitoring interfaces. In all views, the red color indicates process or component failures.}
  \label{fig:rosmon_sysmon}
\end{figure}

Running a complex robot software stack, especially from remote, is a complicated task. We use a combination of a local, onboard launch system and remote high-level system monitor. Our launch system, \texttt{rosmon}\footnote{\url{https://github.com/xqms/rosmon/}}, is a robust replacement for the standard \texttt{roslaunch} tool, which offers several useful features for remote robot operation:
A ROS interface for process status and process control, terminal and remote GUIs for process control (see \reffig{fig:rosmon_sysmon}), automatic log, and process core dump collection in case of failures.

While \texttt{rosmon} allows the supervisors to ascertain whether all required processes are still running, it is impossible to see if the components are running \textit{correctly}. Does the camera driver produce images? Is the GPS signal stable? For such questions, we built a flexible monitoring system consisting of a local ROS node that runs predefined checks every second and a remote GUI component that shows the results to the user in an intuitive way (see \reffig{fig:System_monitor}).

\subsection{Finite State Machines}
\begin{figure}[t!]
\centering
  \begin{subfigure}[c]{0.6\textwidth}
    \includegraphics[trim=00 00 00 00,clip,width=1.0\linewidth]{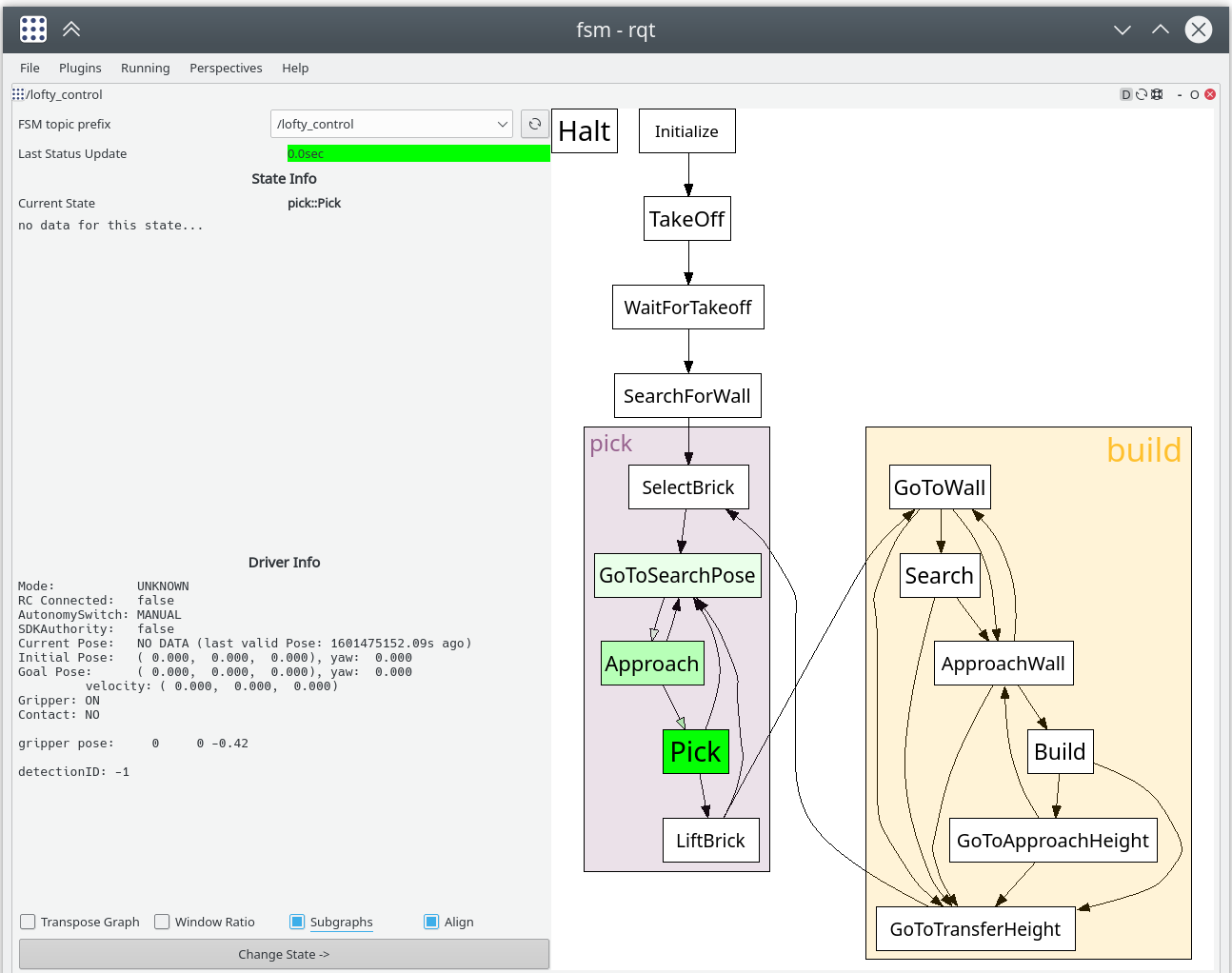}~
    \subcaption{FSM GUI with state-specific information such as robot and brick poses during picking.}
  \end{subfigure}
  \par\bigskip
  \begin{subfigure}[c]{0.6\textwidth}
    \includegraphics[trim=00 00 450 00,clip,width=1.0\linewidth]{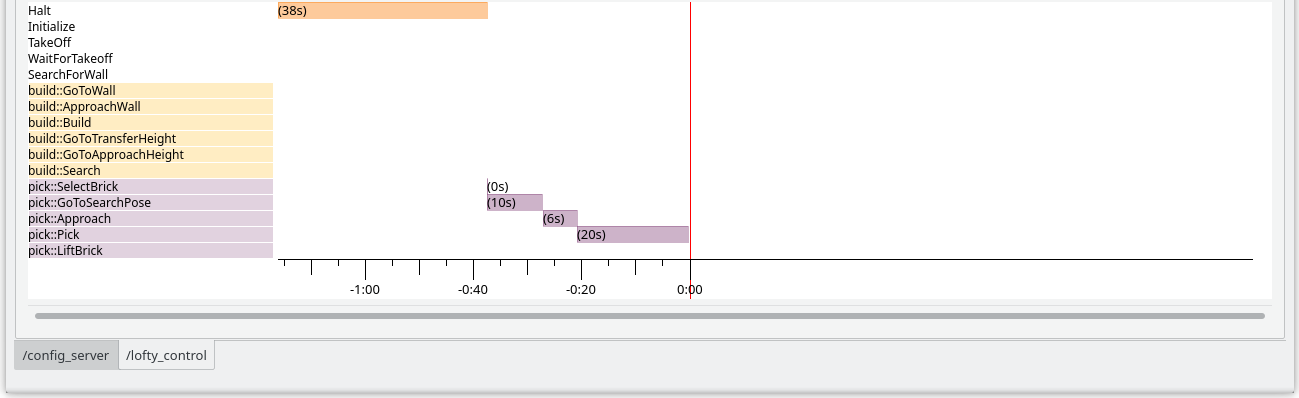}~
    \subcaption{Time-line plot with state activations.}
  \end{subfigure}
  \caption{\texttt{nimbro\_fsm2} GUI showing the state machine of our UAV Lofty (see \refsec{sec:lofty}). The currently active state ``Pick'' is shown in dark green, while the previous two states are shown in lighter shades of green.}
  \label{fig:nimbro_fsm2}
\end{figure}

Finite state machines are a standard tool for the definition and control of robot behavior. For relatively constrained tasks such as the ones defined by MBZIRC, they allow fast construction of behaviors. Instead of working with standard ROS tools such as \texttt{SMACH}, a Python-based FSM framework, we decided to develop our own \texttt{nimbro\_fsm2} library with a focus on compile-time verification. Since testing time on the real robots is limited and simulation can only provide a rough approximation of the real systems, the robot will likely encounter untested situations during a competition run. We trade the ease-of-use of dynamically typed languages and standard frameworks against compile-time guarantees to guard against unexpected failures during runtime.

The \texttt{nimbro\_fsm2} library supports FSM definition in C++. The entire state graph (see \reffig{fig:nimbro_fsm2}) is discovered and verified at compile time using C++ metaprogramming features. \texttt{nimbro\_fsm2} is open-source\footnote{\url{https://github.com/AIS-Bonn/nimbro_fsm2}}.

Our library also automatically publishes monitoring data so that a human supervisor can see the current status. An accompanying GUI displays this data and can trigger manual state transitions, which is highly useful during testing.

Metaprogramming techniques for FSMs, including compile-time optimizations, have been investigated by, \eg \citet{Juhasz2008}. Unfortunately, the authors do \emph{not} provide an implementation, and in contrast to our approach, no user interface is provided.

The combination of our tools for multi-robot communication, transparent state execution, and easy supervision is groundbreaking. Although we developed these tools for the tasks posed in robotic competitions, their potential use reaches far beyond these demonstrations. With robotic systems becoming more complex, featuring more degrees of freedom, and robots communicating with each other, keeping the overview over these systems becomes increasingly difficult. Our monitoring and communication shaping approaches help to not get lost in complexity and to scale to systems of systems with a large number of robots.

\subsection{Trajectory Generation and Control}
\label{sec:Trajectory_Generation_and_Control}
Our UAVs need to be fast and agile in order to score competition points. At the same time, they need to precisely arrive at target positions. Our trajectory generation and control method\footnote{\url{https://github.com/AIS-Bonn/TopiCo}} is based on the method that already reliably worked during MBZIRC 2017 \citep{ecmr2017_c1,ecmr2017_c3}. For the MBZIRC 2020, we only changed parameters to adapt the behavior to the specific challenges. The parameters for the horizontal x- and y-Axis and the vertical z-Axis are presented in \reftab{tab:Parameters_used_at_MBZIRC_2020}.
Our method is described in detail by \citet{beul2016icuas} with the extensions from \citet{beul2017icuas}. For brevity, in this section, we cover only the most essential aspects of the algorithm.

\begin{table}
\small
\caption{Trajectory parameters at MBZIRC 2020.}
\label{tab:Parameters_used_at_MBZIRC_2020}
\vspace{-1.0em}
\begin{center}
\setlength{\tabcolsep}{1.9mm}
\begin{tabular}{ccl|ccl}
  \toprule
  Parameter         & Axis & Value        & Parameter & Axis & Value                   \\
  \midrule
  $v_{max}$         & X, Y & 5.0\,m/s     & $v_{max}$ & Z    & $\hphantom{0}$1.0\,m/s \\
  $a_{max}$         & X, Y & 4.0\,m/s$^2$ & $a_{max}$ & Z    & 10.0\,m/s$^2$           \\
  $j_{max}$         & X, Y & 5.0\,m/s$^3$ & $j_{max}$ & Z    & 50.0\,m/s$^3$           \\
  \bottomrule
\end{tabular}
\end{center}
\end{table}

Based on a simple triple integrator model, our method analytically generates third-order time-optimal trajectories that satisfy input ($j_{min} \leq j \leq j_{max}$) and state constraints ($a_{min} \leq a \leq a_{max}$, $v_{min} \leq v \leq v_{max}$). Trajectories are computed from the current state $(p,v,a)_{UAV}^\intercal$ to the target state $(p,v,a)_{target}^\intercal$. The x, y, and z-axis are synchronized to arrive at the target state at the same time. By doing so, the UAV flies on a relatively straight path.

We directly use this trajectory generation method as a model predictive controller (MPC), running in a closed loop at 50\,Hz on all UAVs. As stated above, our MPC generates jerk commands, but the low-level flight controllers expect pitch resp. roll commands. We therefore assume pitch and roll to directly relate to $\theta = \atantwo(a_x,g)$ and $\phi = \atantwo(a_y,g)$. Thus, we send smooth pitch $\theta$ and roll $\phi$ commands for horizontal movement and smooth climb rates $v_z$ instead.

Although an arbitrary number of axes can be controlled by the above-mentioned method, we do \emph{not} consider the yaw-axis to be synchronized with the x, y, and z-axis. For simplicity, we use proportional control for the yaw-axis.

Like our tools for robot communication and supervision, also our universal MPC contributed to making the effort of operating multiple very diverse UAVs tractable. Due to its inherent dynamic model independence, we used it on all UAVs without having to keep track of individual parameter sets like PID gains. It also reduced the control system complexity since we didn't have to model challenge-specific dynamics like UAVs' changing weight when dropping a brick.

We see our MPC as a prime example of how the scientific contribution of a novel method leads to added value in the real world. By releasing our tools as open-source to the public, we transfer this value to the robotics community and, therefore, advance the entire field which adds value to our \emph{own} systems in turn.

\section{Balloon Hunt}
\label{sec:Balloon_Hunt}

The first task of the MBZIRC 2020 Challenge~1 required teams to find and pop balloons in an outdoor arena of size 90\,$\times$\,40\,m. Five green balloons with approximately 60\,cm diameter were randomly placed inside on top of 2.5\,m long poles. Although the total challenge time was set to 15\,min, the task had to be completed much faster and autonomously to receive a high score.

The task mainly promoted the development of advanced (visual) perception methods, multi-UAV planning, and aerial manipulation capabilities that are beneficial for real-world applications like \eg crop dusting.

Up to three UAVs could be used to complete the challenge, but we found it sufficient to use only one UAV.
While global navigation satellite system (GNSS) positioning was available, the use of differential GNSS was penalized. For this task, we designed the UAV ``Jelly'' including fast TPU-based perception, robust filtering of sensor data, and fast trajectory generation and control \citep{beul2020ssrr_mbzirc}.

\subsection{Hardware}
\label{sec:Hardware}

\begin{figure}
  \centering
  %\resizebox{0.62\linewidth}{!}{\input{mav.pgf}}
  \includegraphics[width=0.62\linewidth]{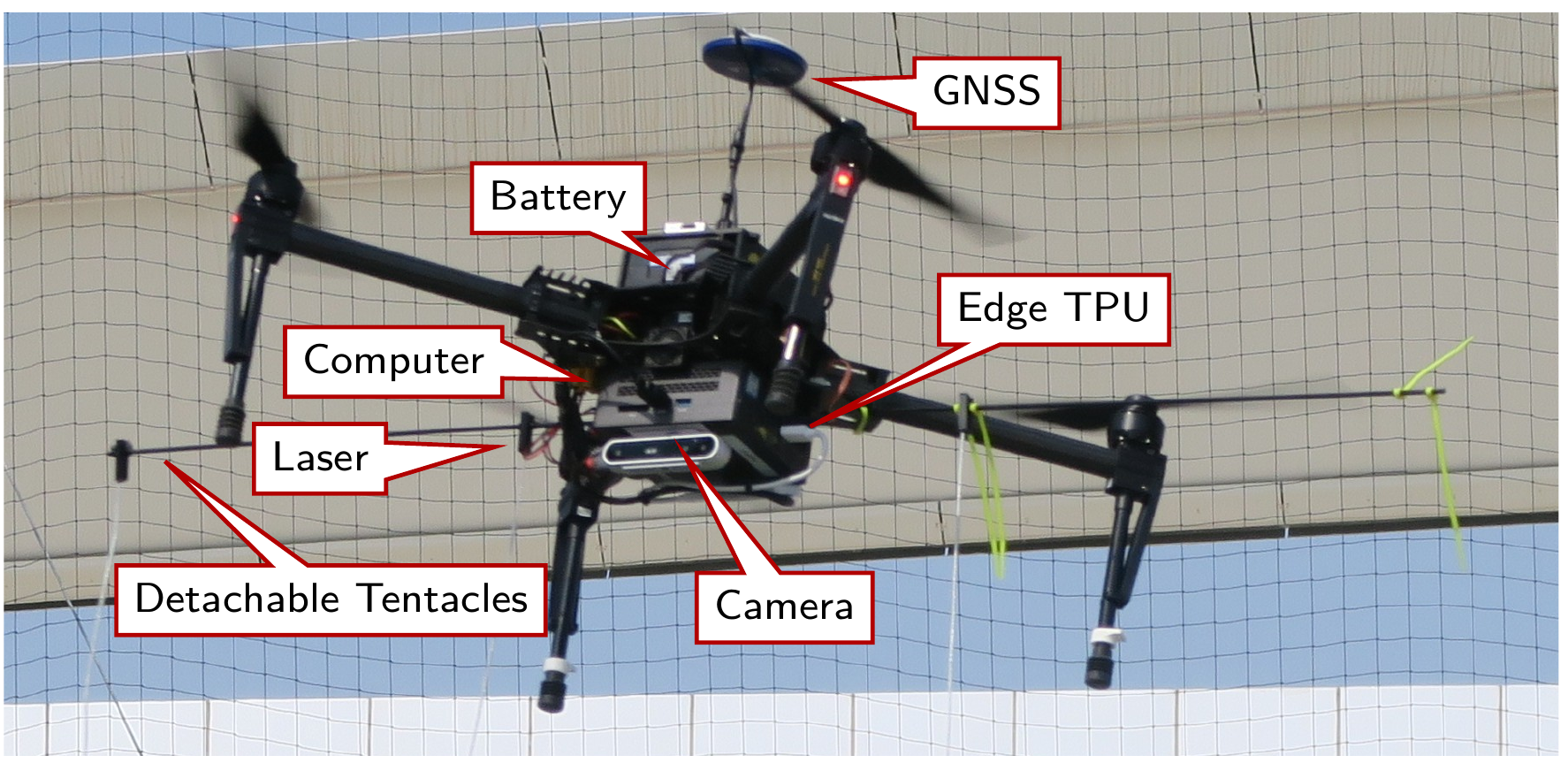}~
  \caption{Design of our UAV ``Jelly'' equipped with four detachable spiked tentacles, an Intel\textsuperscript{\textregistered} RealSense\textsuperscript{\texttrademark} D415 camera, a Google Edge TPU, a laser height sensor, and a lightweight but powerful onboard computer.}
  \label{fig:mav}
\end{figure}

\begin{figure}
  \centering
  %\resizebox{0.62\linewidth}{!}{\input{tentacles.pgf}}
  \includegraphics[width=0.62\linewidth]{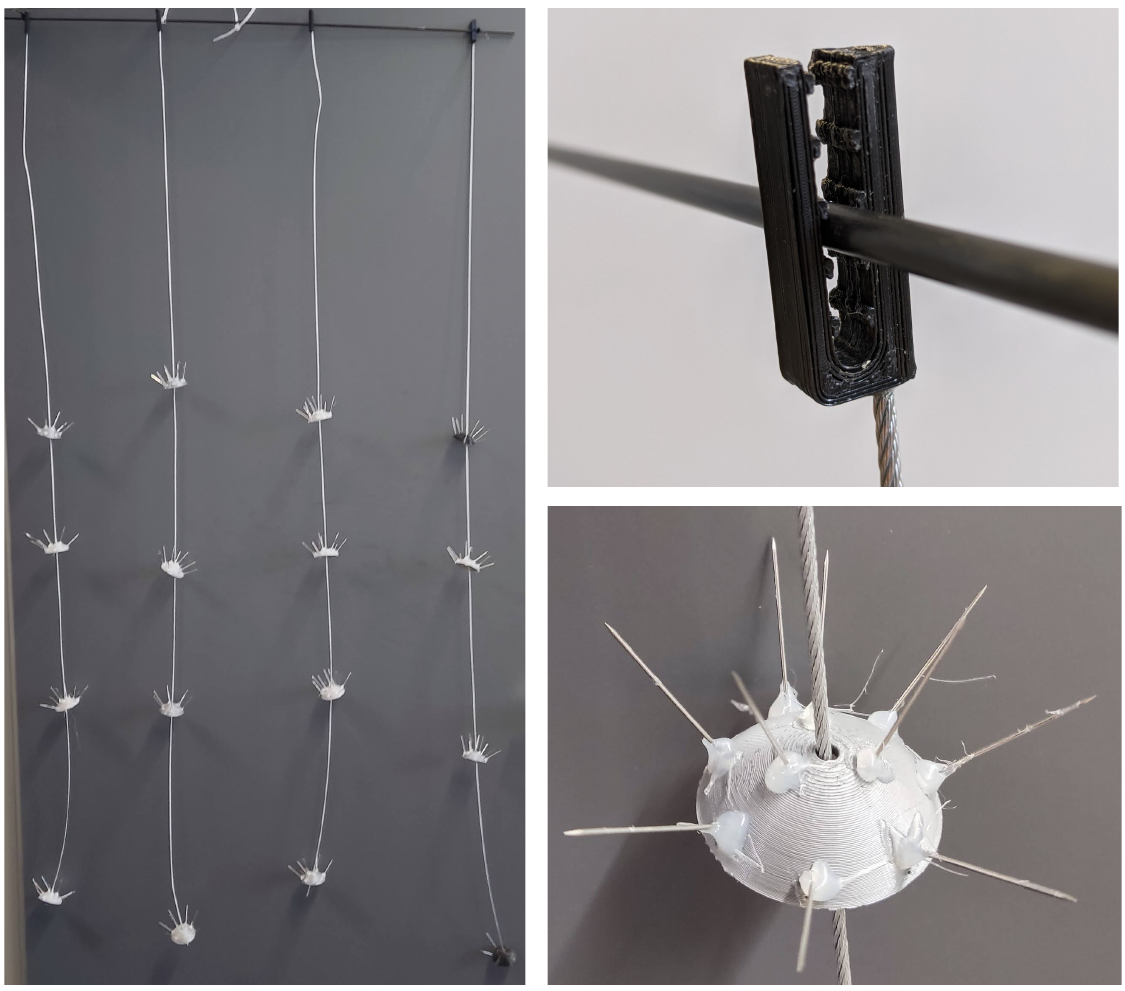}~
  \caption{Detachable spiked tentacles of our UAV ``Jelly''.}
  \label{fig:tentacles}
\end{figure}

Jelly, shown in \reffig{fig:mav}, is based on the DJI Matrice~100 platform. We equipped it with a small but fast Gigabyte GB-BSi7T-6500 onboard PC with an Intel\textsuperscript{\textregistered} Core\textsuperscript{\texttrademark} i7-6500U CPU running at 2.5/3.1\,GHz and 16\,GB RAM. Balloons are perceived by an Intel\textsuperscript{\textregistered} RealSense\textsuperscript{\texttrademark} D415 depth camera with the assistance of a Google Edge TPU USB accelerator. For precise height estimation, Jelly uses a downward-facing LIDAR-Lite v3.

Balloons are punctured with four detachable spiked 1.4\,m long tentacles mounted on a horizontal bar and spaced out at a distance of 30\,cm, as shown in \reffig{fig:tentacles}. When a force of more than 2\,N is applied to a tentacle (\eg by entangling with the poles), it detaches, preventing Jelly from crashing. On each cable, we mounted four needle-spiked hemispheres with a 15\,cm distance. Using flexible popping hardware, Jelly complied with the size restrictions of 1.2\,$\times$\,1.2\,$\times$\,0.5\,m, still offering a forgiving popping system that does \emph{not} require centimeter-level precision.

\begin{figure}
  \centering
  \includegraphics[width=0.50\linewidth]{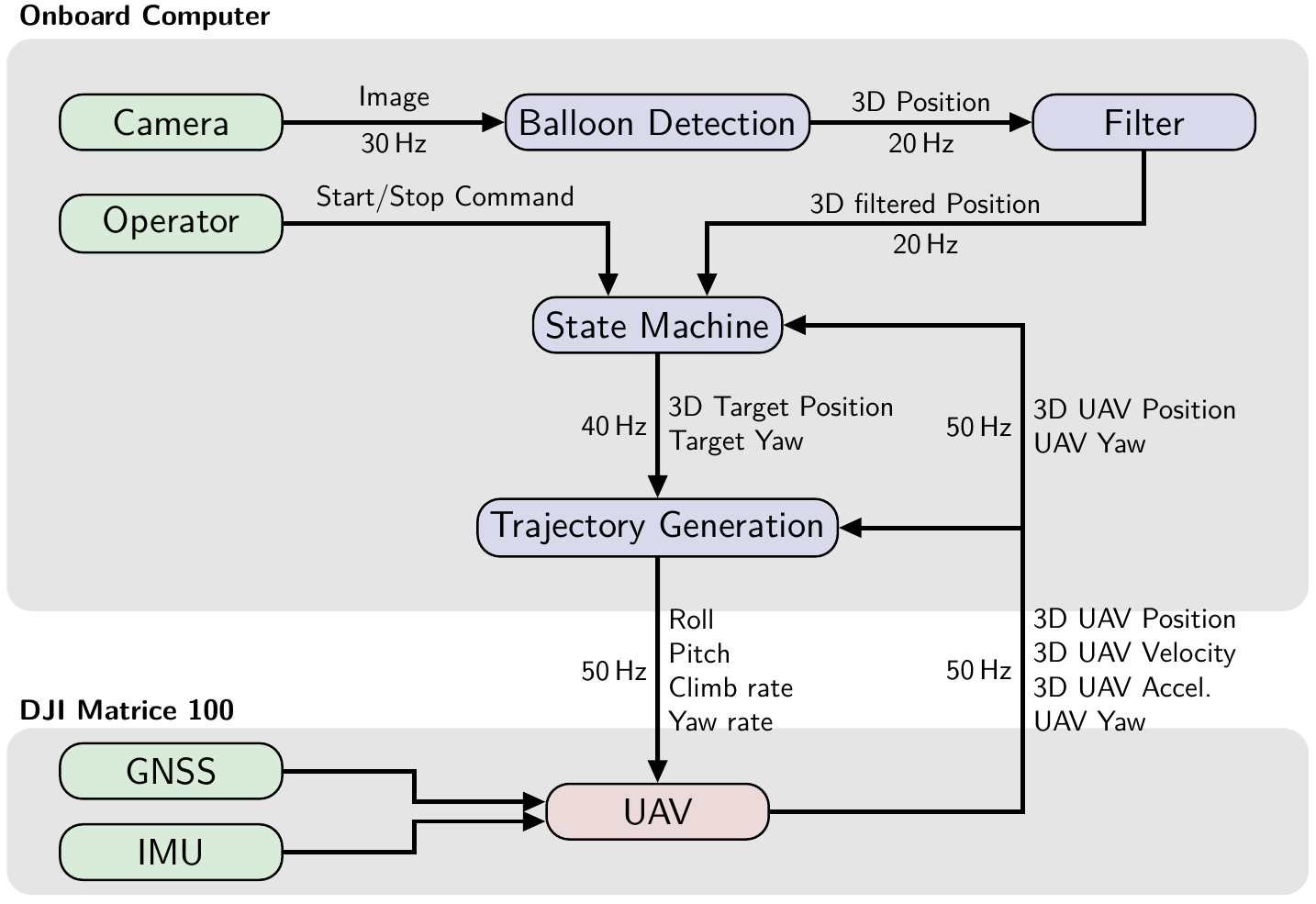}~
  \caption{Structure of our software stack. Green boxes represent external inputs like sensors, blue boxes represent software modules, and the red box indicates the UAV flight control. Position, velocity, acceleration, and yaw are allocentric.}
  \label{fig:structure}
\end{figure}

For allocentric localization and state estimation, we employ the filter onboard the DJI flight control that incorporates GNSS and IMU data. To make all components easily transferable between the test area at our lab and also different arenas on-site, we defined all coordinates (x, y, z, yaw) in a field-centric coordinate system. The center and orientation of the current field were broadcasted by a base station PC to the UAV. In contrast to other teams, we did \emph{not} use advanced satellite-based localization methods like Real-Time Kinematic positioning (RTK-GPS) that need multiple GPS antennas on the UAV. \reffig{fig:structure} gives an overview of the information flow in our system.

\subsection{Mission Control State Machine}
\label{sec:Mission_Control_State_Machine}
\begin{figure}
  \centering
  \includegraphics[width=0.75\linewidth]{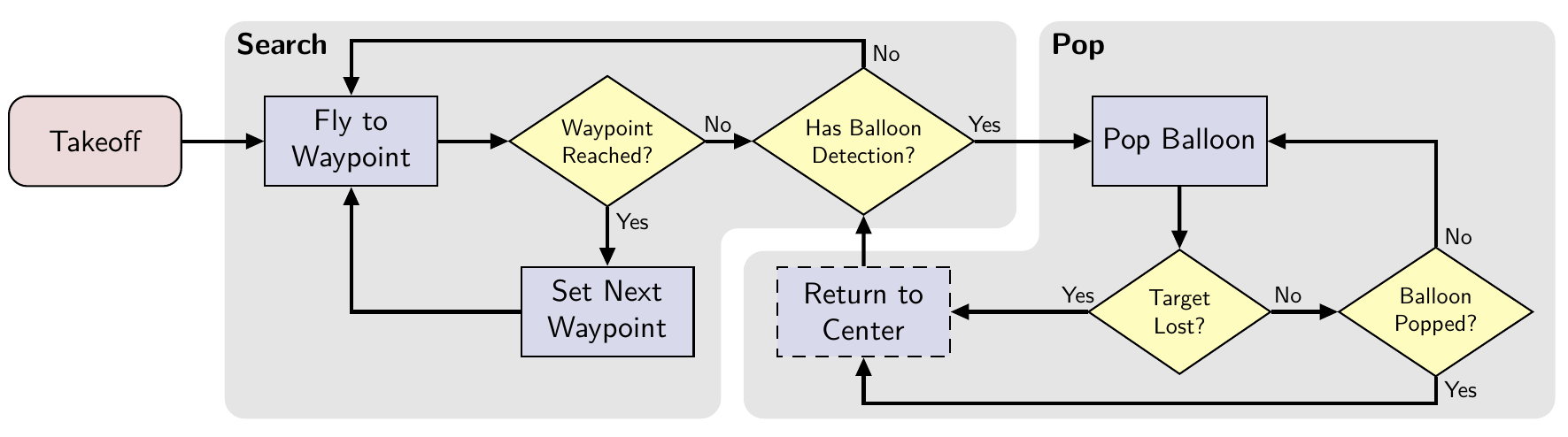}~
  \caption{Flowchart of our state machine for balloon popping.}
  \label{fig:state_machine}
\end{figure}

Jelly's behavior is controlled by a state machine that serves as a generator for waypoints and headings for the subsequent control layers. It also ensures that Jelly does \emph{not} exceed arena limits and stays within a defined altitude corridor so that it always stays above the balloon mounting poles and below the 5\,m minimum altitude of the other subchallenge's UAVs. \reffig{fig:state_machine} shows a flowchart of our state machine, which consists of two alternating parts: Search and Pop. In search mode, Jelly flies a repeating creeping-line pattern along the field's long axis, thereby scanning the entire arena. In Pop mode, Jelly flies a trajectory that drags the tentacles through detected balloons.

Jelly's velocity in search mode is tuned to 5.0\,m/s, and the altitude is 4.0\,m so that balloons can be reliably searched from a safe height. Our balloon detector produces reliable position estimates at ranges over 30\,m. During the Grand Challenge, the search pattern comprised two search lanes, spaced at 10\,m from the arena limits, which proved sufficient. As depicted in \reffig{fig:structure}, all balloon detections are filtered as described in \refsec{sec:Allocentric_Balloon_Filter} before being processed by the state machine. The filter provides a list of verified balloon positions within the arena limits, including those currently out of view. Once the state machine receives at least one detection, it proceeds to approach the closest target. Jelly does \emph{not} consider observation waypoints for the detection but only searches for balloons while flying from one waypoint to the next. Jelly does \emph{not} stop at the waypoints since when it reaches the current waypoint (within a certain radius), the transition is triggered, and the next waypoint is set active.

In Pop mode, we compute a straight-line trajectory, such that the center of the tentacles passes through the balloon center at a non-zero velocity. The tentacles' upward-facing needles are dragged into the balloon surface, effectively puncturing it. As the forward-facing camera cannot perceive the balloon all the way, it is assumed to be popped once Jelly passes over the estimated center of the balloon instance within a 0.5\,m radius. Should the balloon still be intact due to unsuccessful puncturing or missing the intercept point, it will be tackled again later as the search pattern repeats, and the balloon will inevitably be re-detected. If the target is lost during the approach, \eg because the filter discarded a false positive, the attempt is canceled. As an addition, in the Grand Challenge, after each attempt to pop a balloon, Jelly returns to the center of the field to prevent flying into the scaffolding protruding the arena. This method of handling the field's non-convexity is simple, but it introduces additional flying time compared to real obstacle avoidance. On the other hand, it is easy to implement and reliable. After returning to the center, Jelly targets the subsequent closest balloon provided by the filter or resumes with the search pattern if there are no viable balloon hypotheses.

\subsection{Balloon Perception}
\label{sec:Balloon_Perception}
\begin{figure}
    \centering
    \begin{subfigure}[t]{0.4\linewidth}
        \centering
        \includegraphics[width=\linewidth]{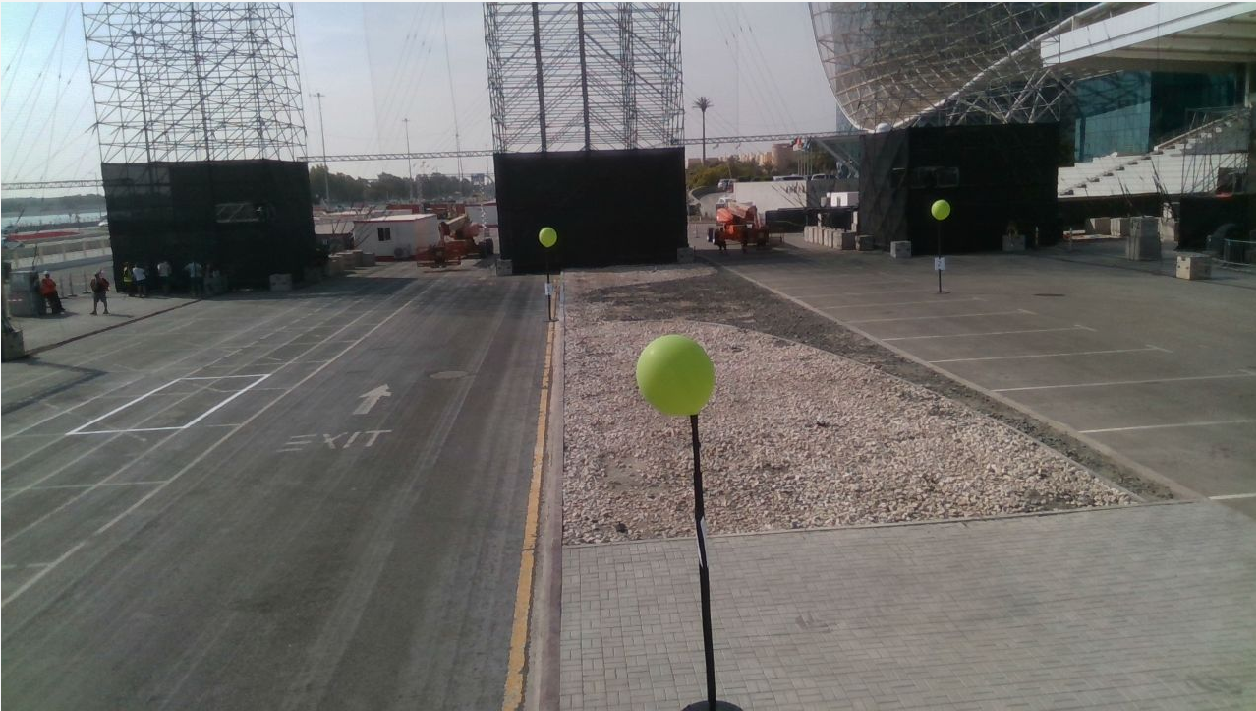}
        \caption{Input image.}
    \end{subfigure}
    \begin{subfigure}[t]{0.4\linewidth}
        \centering
        \fbox{\includegraphics[width=\linewidth]{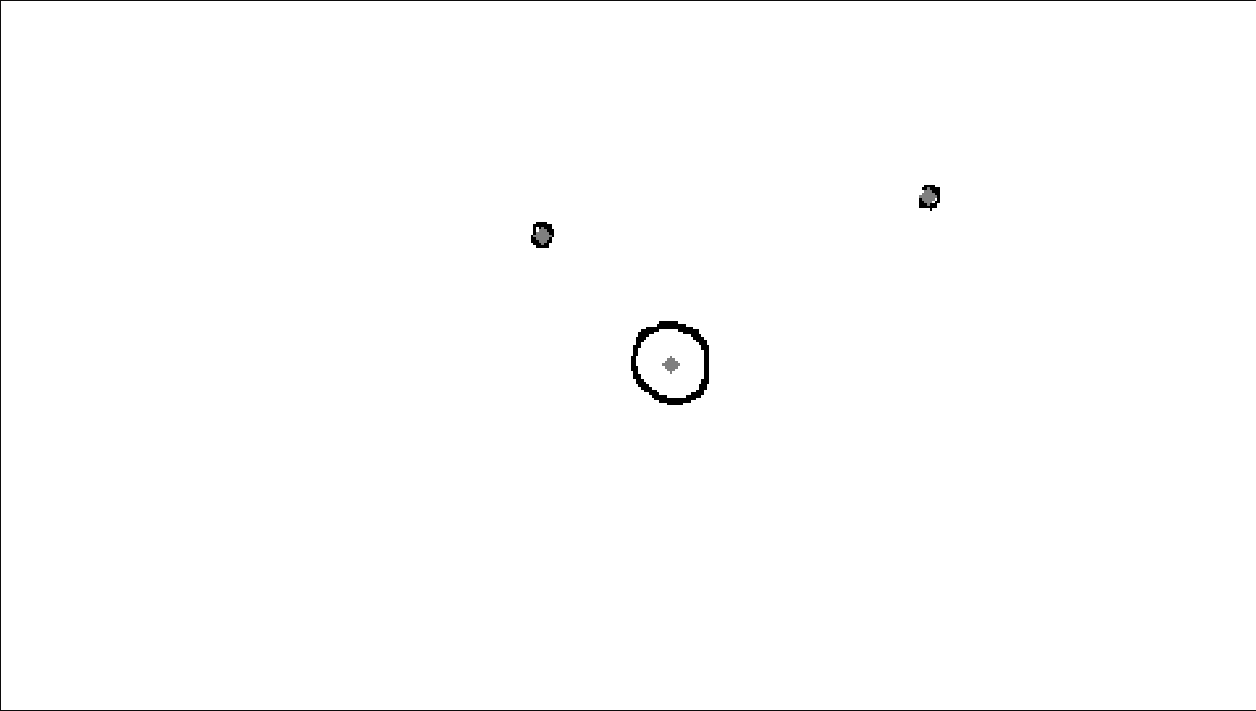}}
        \caption{Balloon detections.}
    \end{subfigure}
    \caption{Balloon perception. The detected balloon outlines are drawn with black lines, and the centers are marked with gray points. All visible balloons are detected correctly, even at large distances.}
    \label{fig:balloon_perception}
\end{figure}

Our approach for detecting balloons in images is based on deep learning methods and split into an inference and a postprocessing step. During the inference step, we employ a neural network for semantic segmentation. Since we aim to detect multiple balloons, we perform a binary segmentation of the raw input image and extract the balloon outlines, as shown in \reffig{fig:balloon_perception}. The balloon detection itself is carried out in the postprocessing step, which provides information like the number of balloons, their confidence values, and their positions in camera coordinates.

\begin{figure}
  \centering
  \includegraphics[trim=00mm 00mm 00mm 00mm,clip,width=0.48\linewidth]{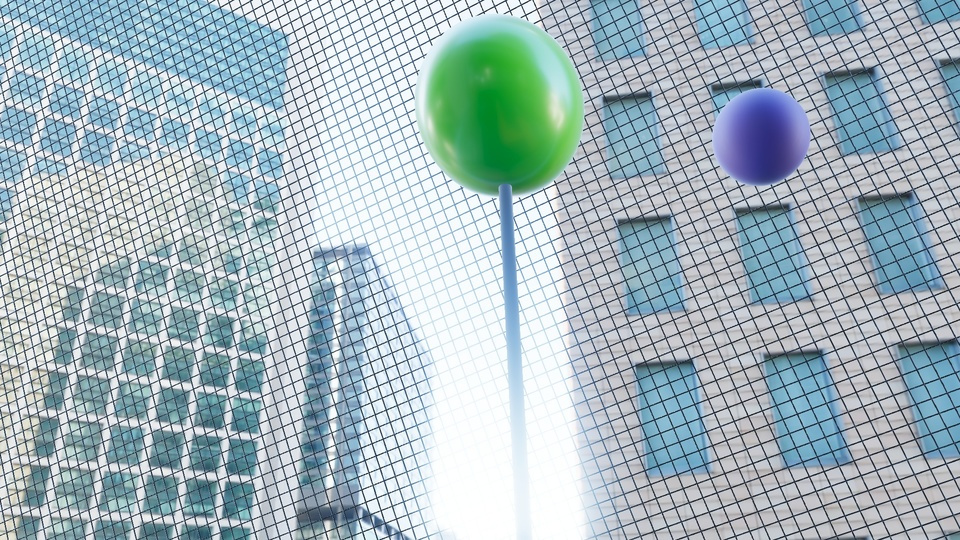}~
  \includegraphics[trim=00mm 00mm 00mm 00mm,clip,width=0.48\linewidth]{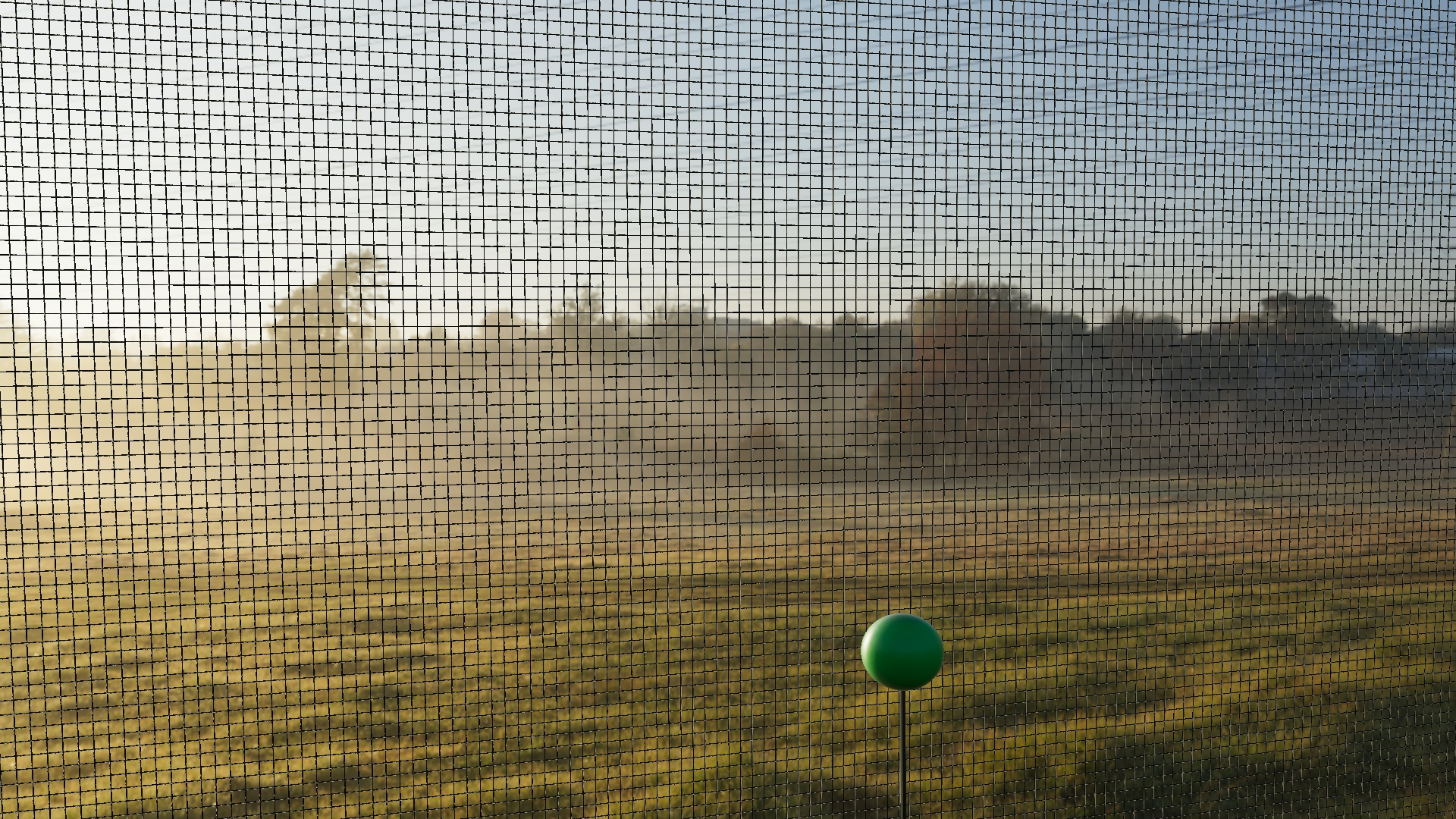}~
  \caption{Synthetic training frames for balloon detection.}
  \label{fig:balloon_synthetic}
\end{figure}

\begin{figure}
  \centering
  \includegraphics[trim=00mm 00mm 00mm 00mm,clip,width=1.0\linewidth]{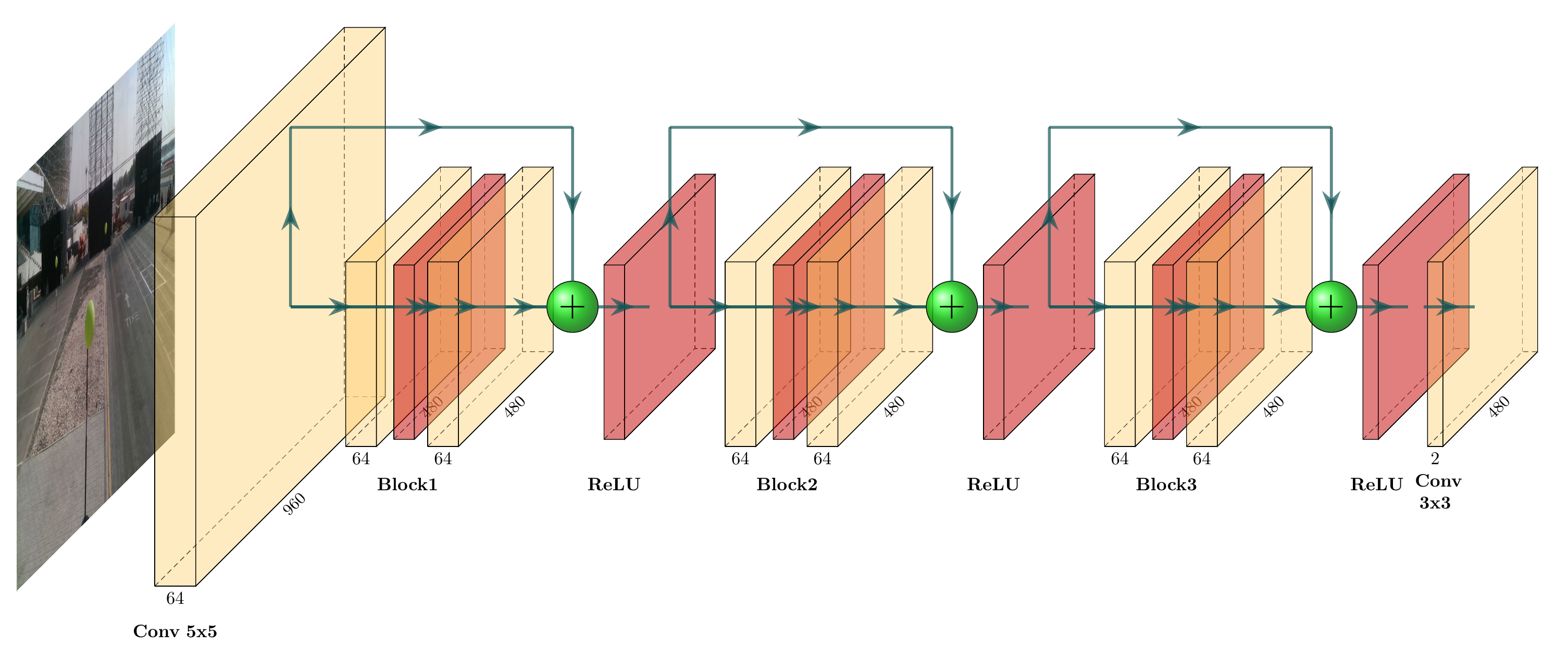}~
  \vspace{-2em}
  \caption{Balloon segmentation CNN network architecture.}
  \label{fig:NN}
\end{figure}

\paragraph{Balloon Outline Segmentation Network}
The neural network structure is based on the first three blocks of ResNet-18 \citep{he_deep_2016}. See \reffig{fig:NN} for a visualization of our network. The network outputs a binary segmentation consisting of background and balloon outline classes.
For training, the network is initialized from ResNet-18 pre-trained with ImageNet. We use a dataset consisting of 10\,k synthetic and 300 real images. The synthetic images were generated by a lightweight physically-based renderer\footnote{\url{https://github.com/RaduAlexandru/easy_pbr}} \citep{rosu2021grapp}. A 60\,cm diameter sphere is randomly placed in different HDR environments, as shown in \reffig{fig:balloon_synthetic}. The images also include spherical shapes that are not green and do not correspond to a balloon to reduce the number of false positives. The real images were recorded with the same Intel\textsuperscript{\textregistered} RealSense\textsuperscript{\texttrademark} D415 camera, which we used during the competition. To enhance the network's robustness even further, we added noise to the synthetic data as described by \citet{inbook}. We trained the network with an image size of 960\,$\times$\,540\,px.

\paragraph{Postprocessing}
The postprocessing step aims to detect balloons in the binary segmentation output, as shown in \reffig{fig:balloon_perception}. The binary segmentation provides contours of the balloons, which can still be noisy or could be false positives on the background. The valid balloon detections are extracted from the segmentation output based on a connected component analysis. In the first step, connected components are extracted, and valid components are filtered by a minimum number of pixels. Then, for each valid component, a circle is fitted based on $N$ points sampled equally distributed from the connected component.
Detections are filtered by a size threshold and a threshold on the residuals of the fitted circle, which give a measure for the quality of the estimated circle.
A 3D position estimate can be calculated based on the detected balloon radius.
The resulting 3D balloon center points are then further processed by an allocentric filter (\cf\refsec{sec:Allocentric_Balloon_Filter}).

The entire pipeline is very time efficient and runs with an average processing time of 45\,ms per frame on the onboard CPU with TPU acceleration. The quantization of the network for processing on the Edge TPU did \emph{not} lead to a decrease in prediction quality.
Finetuning during the competition resulted in a significant decrease in background noise and enhanced the network's balloon outline detection. Consequently, parameters like residual threshold and minimum connected component size in postprocessing were adapted so that we archived stable balloon detections even at large distances of up to 50\,m.

\subsection{Allocentric Balloon Filter}
\label{sec:Allocentric_Balloon_Filter}
For each image frame, the balloon perception (\refsec{sec:Balloon_Perception}) outputs a list of current balloon detections $d^i_1, \dots, d^i_n$, described as egocentric 3D positions in camera coordinates.
These are processed by a filter to reject outliers and to aggregate them into a list of hypotheses $\mathcal H$ of possible balloon positions.
Each hypothesis $\mathcal H_i\in\mathcal H$ consists of
\begin{compactitem}
 \item a history $\mathcal D_i := (d^i_1, \dots, d^i_8)$ of the last eight detections that were assigned to it,
 \item an estimate of the balloon position $P_i:=\frac{1}{|\mathcal D_i|}\sum_{d\in\mathcal D_i} d$, calculated as the running average over the detection history, and
 \item a counter for missed detections.
\end{compactitem}
All hypotheses with at least eight detections are sorted with increasing distance to the current UAV position and forwarded to the state machine.

\begin{figure}[t]
  \centering
  \begin{subfigure}[t]{0.49\linewidth}
    \centering
%     \begin{tikzpicture}[font=\sffamily\footnotesize]
%       \node[inner sep=0] (image1) at ( 0.00, 2.50) {\includegraphics[trim=00mm 00mm 00mm 00mm,clip,width=1.0\linewidth]{grounddistancesmall.png}};
%       \begin{scope}[x={($(image1.south east) - (image1.south west)$)},y={($(image1.north west) - (image1.south west)$)}, shift={(image1.south west)}]
%         \node[black,anchor=east] at (1.0, 0.7) {UAV};
%         \draw[yellow] (0.538,0.445) -- (0.439,0.455);
%         \draw[blue,fill=blue] (0.538,0.445) circle (0.2mm);
%         \draw[black!20!green,fill=black!20!green] (0.439,0.455) circle (0.2mm);
%       \end{scope}
%     \end{tikzpicture}
    \includegraphics[width=1.0\linewidth]{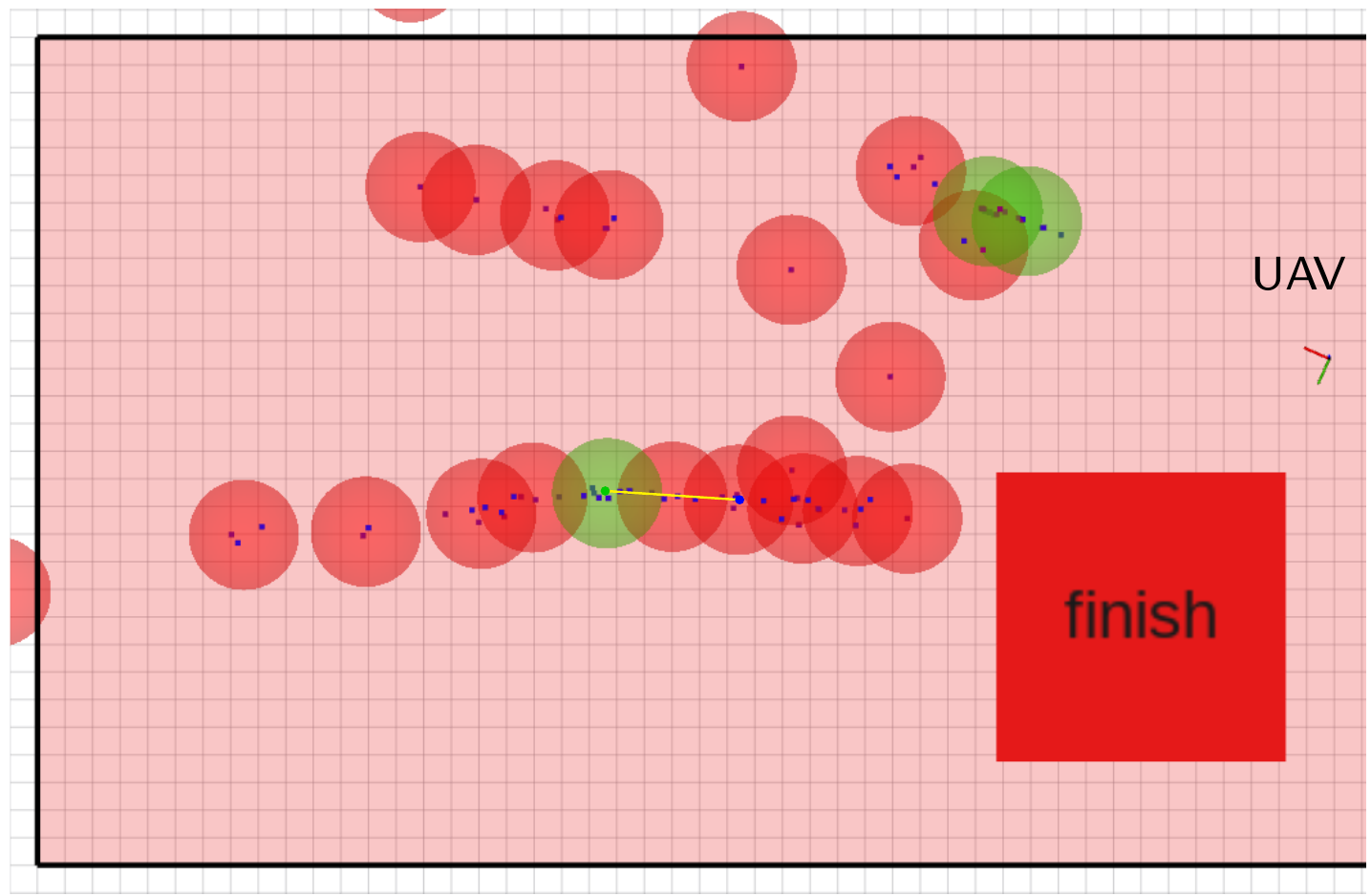}%
    \caption{Euclidean distance on ground plane. Exemplarily, the Euclidean distance of a detection is shown in yellow. It exceeds the threshold and, thus, a new hypothesis is created.}
    \label{fig:distance_metrics_a}
  \end{subfigure}
  \hfill
  \begin{subfigure}[t]{0.49\linewidth}
    \centering
%     \begin{tikzpicture}[font=\sffamily\footnotesize]
%       \node[inner sep=0, right=0.2cm of image1] (image2) {\includegraphics[trim=00mm 00mm 00mm 00mm,clip,width=1.0\linewidth]{raydistancesmall.png}};
%       \begin{scope}[x={($(image2.south east) - (image2.south west)$)},y={($(image2.north west) - (image2.south west)$)}, shift={(image2.south west)}]
%         \node[black,anchor=east] at (1.0, 0.7) {UAV};
%         \draw[yellow] (0.27,0.41) -- (0.276,0.365);
%         \draw[blue,fill=blue] (0.38,0.40) circle (0.2mm); 
%         \draw[blue] (0.0,0.269) -- (0.975,0.605);
%         \draw[black!20!green,fill=black!20!green] (0.27,0.41) circle (0.2mm);
%       \end{scope}
%     \end{tikzpicture}
    \includegraphics[width=1.0\linewidth]{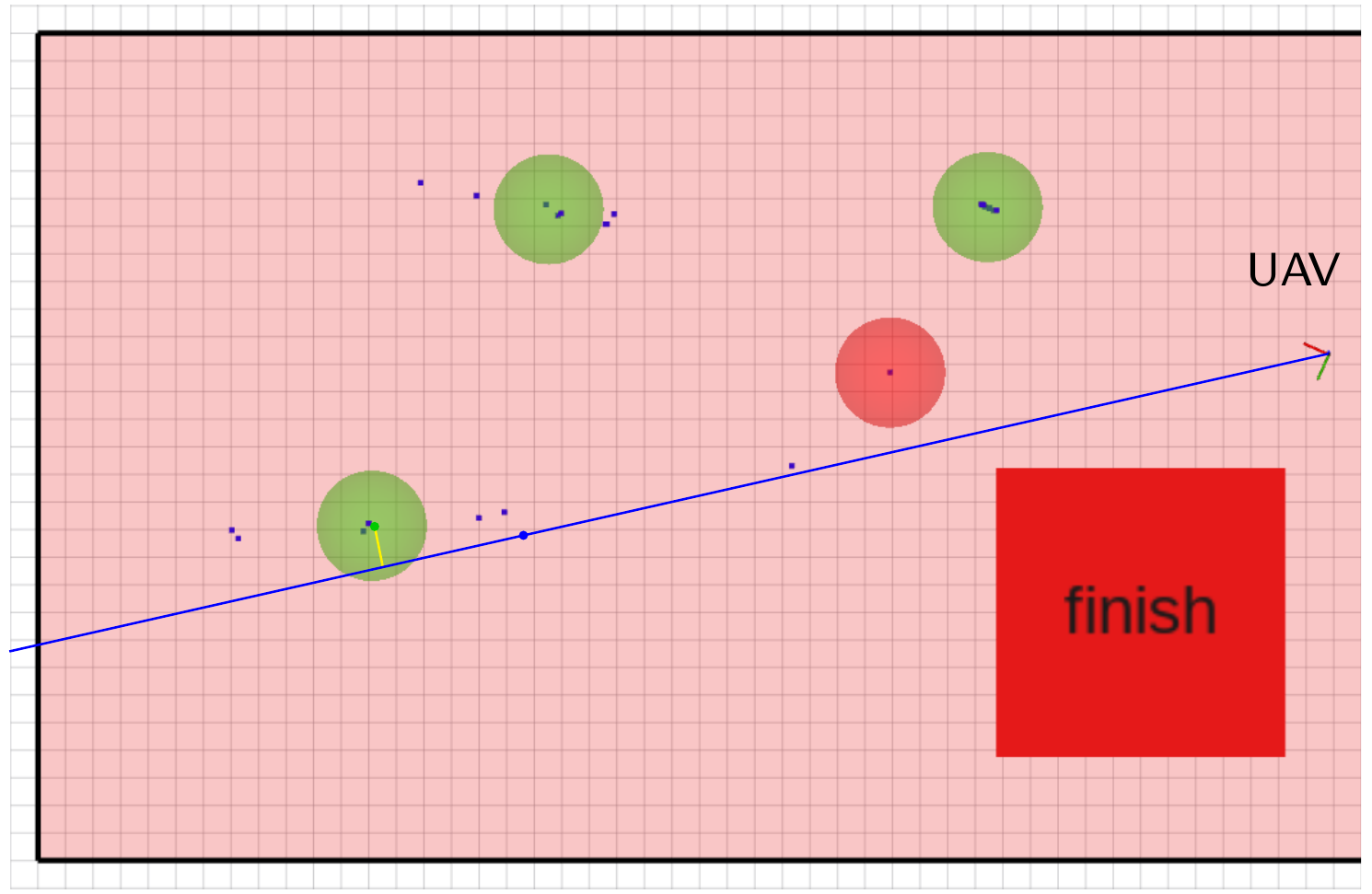}%
    \caption{Distance to detection ray metric. The distance (yellow) to the detection ray (blue) is smaller and does not exceed the threshold. Thus, no new hypothesis is created.}
    \label{fig:distance_metrics_b}
  \end{subfigure}
  \caption{Evaluation of different distance metrics. Detections are shown as blue dots. Hypotheses are shown as spheres, colored green if at least eight detections are assigned to them and red otherwise. The ray metric (b) better reflects the characteristics of our balloon detection pipeline.}
  \label{fig:distance_metrics}
\end{figure}

We transform the egocentric detections of the balloon detector into allocentric field coordinates. Since the height of the balloons was predefined to be at 2.5\,m, all detections outside a height corridor from 1.5 to 5.0\,m are discarded. For each remaining detection $d$, we determine the closest hypothesis $\mathcal H_{i^*}$ by minimizing the distance between the detection and all position estimates, \ie choosing $i^*=\arg\min_i\{\textit{dist}(d,P_i)\}$. If the distance is smaller than a threshold of 2.0\,m, we assign $d$ to $\mathcal H_{i^*}$. Otherwise, we create a new hypothesis. Finally, hypotheses are merged when their estimated balloon positions come closer than 2.0\,m.

The choice of the distance measure $\textit{dist}(\boldsymbol{\cdot})$ is important to reduce the influence of detection noise and thus to achieve accurate assignments. Since the center height of all balloons is fixed to $2.5\,\textrm{m} + \frac{0.6\,\textrm{m}}{2} = 2.8\,\textrm{m}$, $\textit{dist}(\boldsymbol{\cdot})$ can be chosen as the Euclidean distance on the ground plane to eliminate noisy height measurements. However, due to the noisy depth estimation of the egocentric detections, this may result in multiple different hypotheses for the same balloon (\reffig{fig:distance_metrics_a}). Instead, we cast a ray $\tau$ in the detection direction and define $\textit{dist}(\boldsymbol{\cdot})$ as the distance between the estimated balloon position and $\tau$. The latter distance metric better reflects our balloon detector's characteristics since it has a high accuracy in the image plane but is relatively inaccurate in the depth dimension. This change results in more accurate hypotheses assignments, as shown in \reffig{fig:distance_metrics_b}.

Once Jelly reaches a position above an estimated balloon, we assume the balloon to be popped and remove the corresponding hypothesis. If popping was \emph{not} successful, the balloon is detected again later, and thus a new hypothesis for this balloon will be added (\cf\refsec{sec:Mission_Control_State_Machine}).

\subsection{Laser Height Filter}
\label{sec:Laser_Height_Filter}

\begin{figure}
  \centering
  %\resizebox{0.9\linewidth}{!}{\input{height_plot.pgf}}
  \includegraphics[width=0.9\linewidth]{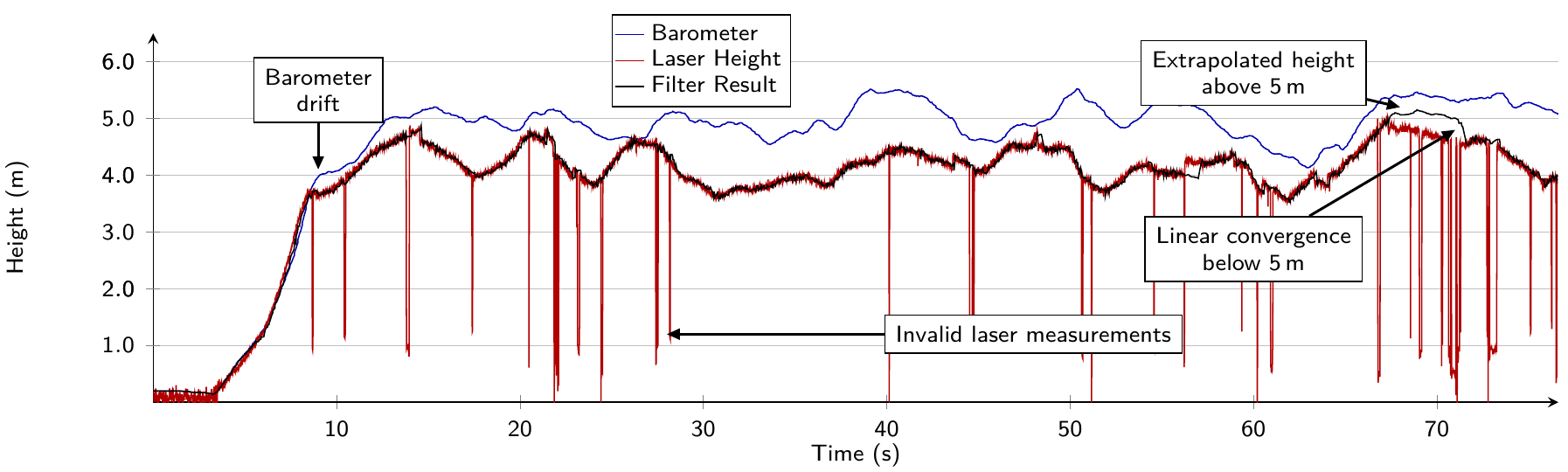}~
  \caption{Laser height estimation during Run 2. The estimated height (black) is either based on barometer data (blue) or laser measurements (red).}
  \label{fig:height_eval}
\end{figure}

Precise height estimation can make the critical difference between popping a balloon, missing a balloon, or hitting a pole.
We therefore use measurements of a downward-facing LIDAR-Lite v3 as the primary height source.
The laser height filter determines whether the laser measurements are valid (\ie within expected boundaries) and thus can be used as height estimation.
However, when the laser measurements are assumed invalid, we extrapolate the latest height estimate using the change in the fused GNSS and barometric height.
As soon as the laser measurement is assumed valid again, we immediately correct our height estimate to the laser height or---if the extrapolated height drifted too much---linearly interpolate, allowing a maximum slope of 1.5\,m/s.
\reffig{fig:height_eval} depicts height measurements and the filtered height estimates during Run~2. During takeoff, height estimation is based on barometer data since laser measurements are unreliable for too close distances. Once Jelly reaches an estimated height of 1.0\,m, height estimation is based on filtered laser data. Above 5.0\,m, the laser becomes unreliable in the bright outdoor conditions, and the height estimate is extrapolated using the change in barometric height measurements.

\subsection{Evaluation}
We operated Jelly for balloon popping in three competition runs during MBZIRC 2020. A video showcasing our Grand Challenge Run can be found on our website\footnote{\url{https://www.ais.uni-bonn.de/videos/fr_2021_mbzirc}}.

% \pgfplotstableread{
% detected v0 v1 v2 v3
% 0   738  3   0   0
% 1   35   443 2   0
% 2   121  69  153 0
% 3   13  48  35 2
% }\balloondetectionhistogram
% 
% \pgfplotstableread{
% radius detected
% 0 111
% 10 476
% 20 163
% 30 84
% 40 30
% 50 22
% 60 16
% 70 8
% 80 11
% 90 13
% 100 6
% 110 2
% 120 2
% 130 1
% 140 2
% 150 2
% }\balloonradiushistogram
% 
% \pgfplotstableread{
% dist detected
% 0 9
% 1 20
% 2 23
% 3 31
% 4 30
% 5 32
% 6 36
% 7 38
% 8 32
% 9 26
% 10 23
% 11 31
% 12 26
% 13 21
% 14 18
% 15 34
% 16 26
% 17 31
% 18 34
% 19 32
% 20 41
% 21 45
% 22 38
% 23 53
% 24 38
% 25 39
% 26 7
% 27 8
% 28 8
% 29 12
% 30 4
% 31 5
% 32 9
% 33 7
% 34 7
% 35 2
% 36 4
% 37 5
% 38 4
% 39 4
% 40 3
% 41 0
% 42 0
% 43 0
% 44 1
% }\balloondisthistogram

\begin{figure}[t]
  \centering
    \begin{subfigure}[T]{0.29\linewidth}
%         \begin{tikzpicture}[font=\footnotesize]
%             \begin{axis}[ybar=1pt, bar width=0.15cm, enlarge x limits=0.15, legend pos=north east, legend cell align={left}, xlabel=\# Visible Balloons, height=5cm, width=1.1\linewidth, xtick distance=1,tick pos=left]
%             \addplot+ table [x=detected,y=v0] {\balloondetectionhistogram};
%             \addlegendentry{0 Detections}
%             \addplot+ table [x=detected,y=v1] {\balloondetectionhistogram};
%             \addlegendentry{1 Detection\ }
%             \addplot+ table [x=detected,y=v2] {\balloondetectionhistogram};
%             \addlegendentry{2 Detections}
%             \addplot+ table [x=detected,y=v3] {\balloondetectionhistogram};
%             \addlegendentry{3 Detections}
%             \end{axis}
%         \end{tikzpicture}
        \includegraphics[width=1.0\linewidth]{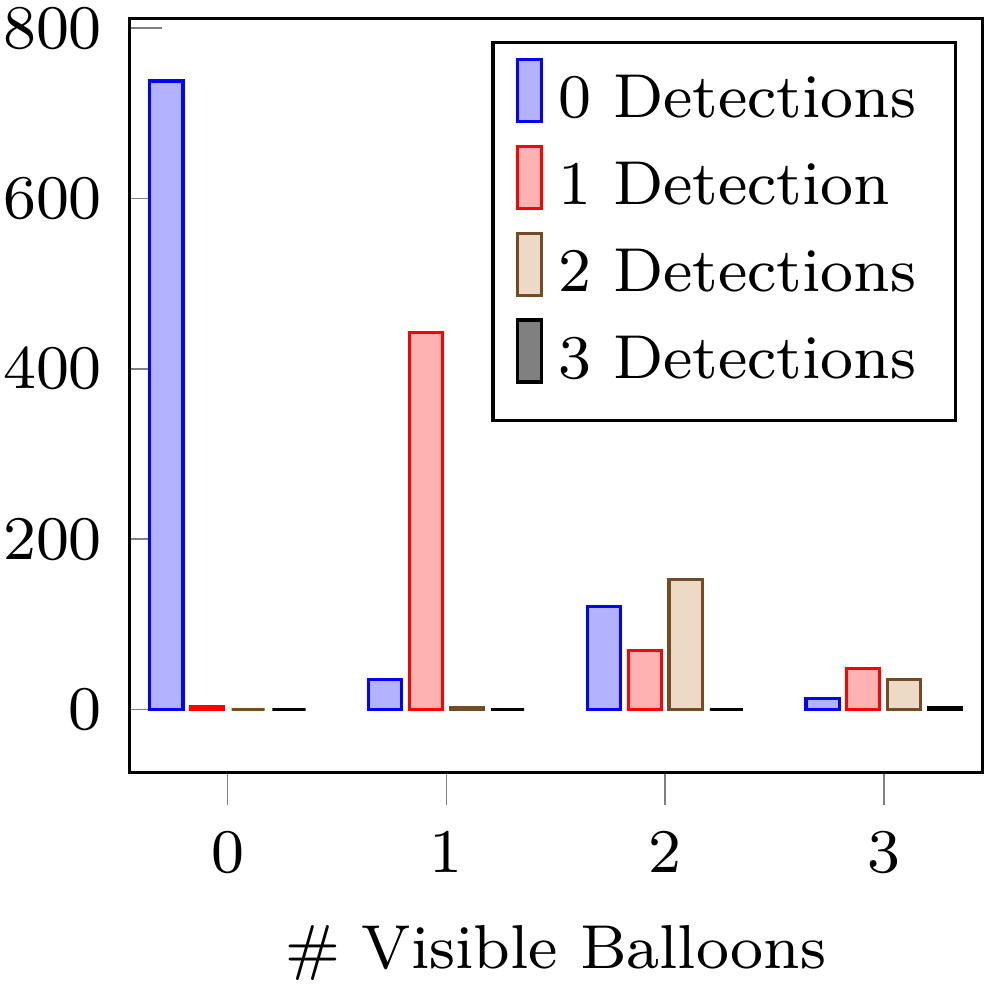}
    \caption{Number of detections compared to number of visible balloons per frame.}
    \label{fig:balloon_detection_histogram_multiple}
    \end{subfigure}
    \hfill
    \begin{subfigure}[T]{0.415\linewidth}
%         \begin{tikzpicture}[font=\footnotesize]
%             \begin{axis}[ybar, bar width=0.13cm, enlarge x limits=0.02, legend pos=north west, xlabel=Balloon Distance \text{[m]}, height=5cm, width=1.1\linewidth, xtick distance=5,tick pos=left, ytick distance=10]
%             \addplot+ table [x=dist,y=detected] {\balloondisthistogram};
%             \end{axis}
%         \end{tikzpicture}
        \includegraphics[width=1.0\linewidth]{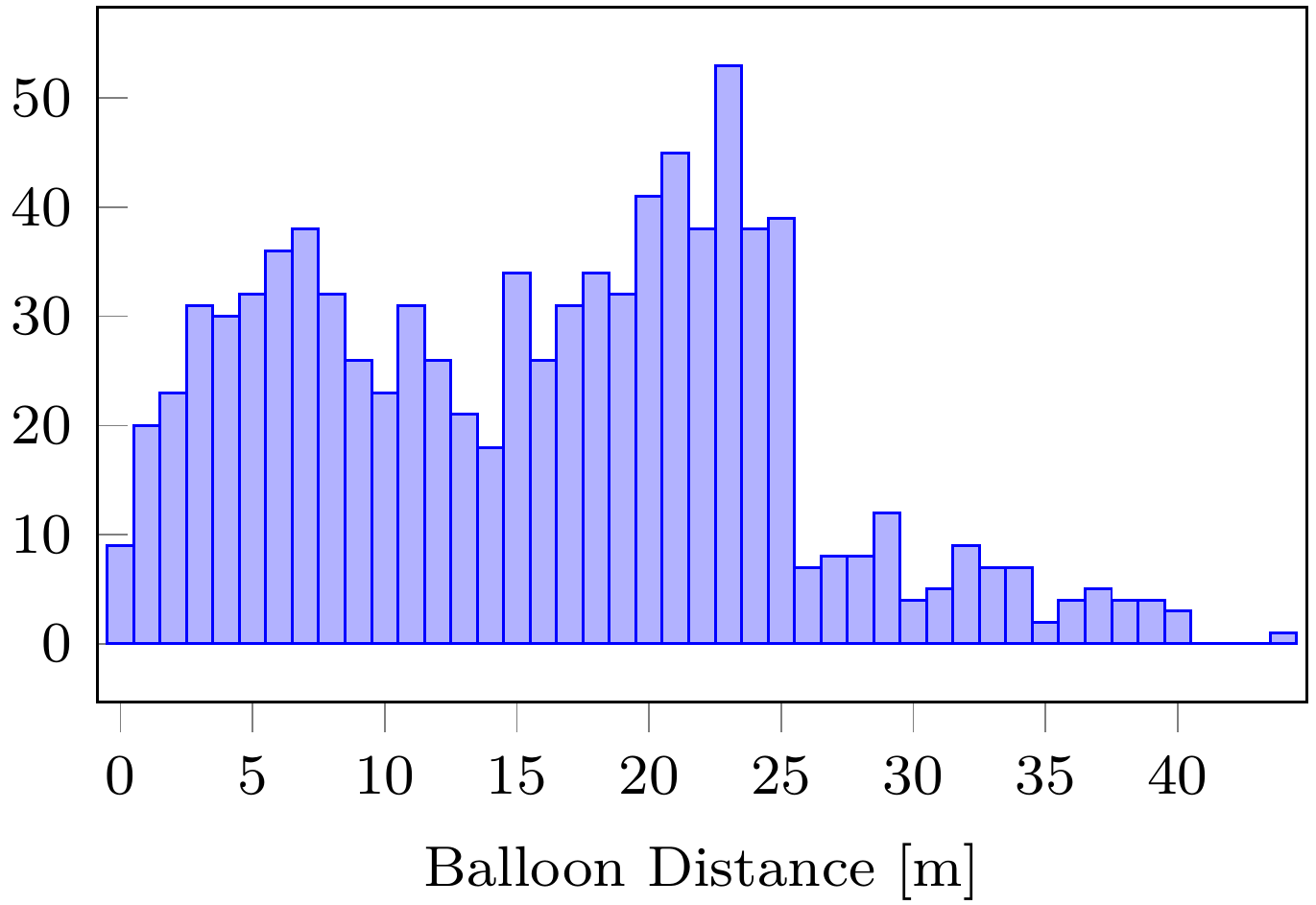}
    \caption{Number of detections compared to distance to closest ground truth balloon position.}
    \label{fig:balloon_detection_histogram_dist}
    \end{subfigure}
    \hfill
    \begin{subfigure}[T]{0.25\linewidth}
%         \begin{tikzpicture}[font=\footnotesize]
%             \begin{axis}[ybar, bar width=0.15cm, enlarge x limits=0.05, legend pos=north west, xlabel=Detected Balloon Radius \text{[Pixel]}, height=5cm, width=1.1\linewidth, xtick distance=40,tick pos=left, ytick distance=100]
%             \addplot+ table [x=radius,y=detected] {\balloonradiushistogram};
%             \end{axis}
%         \end{tikzpicture}
        \includegraphics[width=1.0\linewidth]{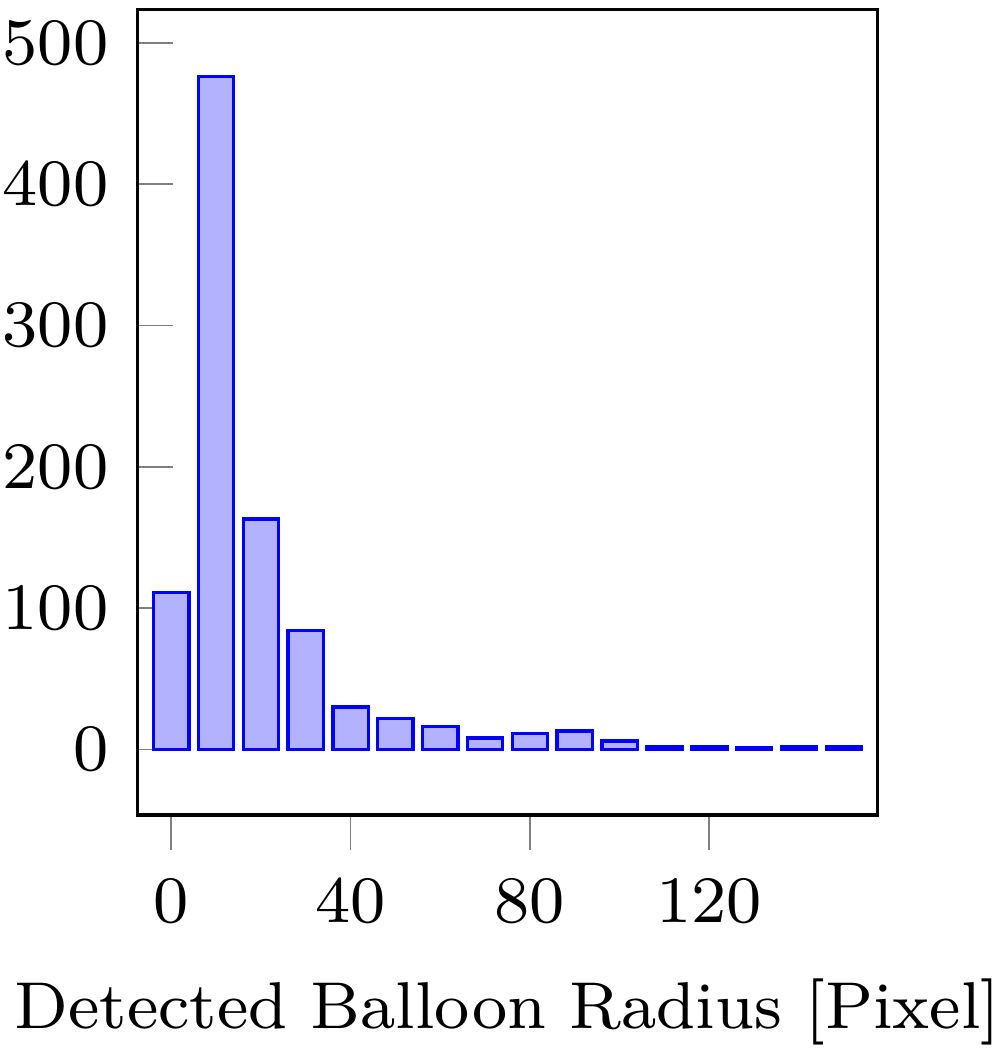}
    \caption{Number of detections compared to pixel radius of detection.}
    \label{fig:balloon_detection_histogram_radius}
    \end{subfigure}
    \caption{Histograms for the number of balloon detections.}
    \label{fig:balloon_detection_histogram}
\end{figure}

To evaluate the performance of our balloon perception (\cf\refsec{sec:Balloon_Perception}), we manually check the balloon detections for the Grand Challenge Run, during which a total of 1662 image frames have been processed by our vision pipeline.
In total, 1460 balloons occur in these images, of which 949, \ie 64\,\%, are correctly detected, while there are only five false detections.
\reffig{fig:balloon_detection_histogram_multiple} shows histograms of the number of balloon detections per frame.
Frames where no balloon is visible are correctly classified in 99.6\,\% of all cases.
If a single balloon is visible, we achieve an accuracy of 92.3\,\%.
However, frames with more than one balloon often contain more distant balloons, which are more difficult to detect.
Thus, detections are missed more frequently in these cases.

To analyze for which distances our vision pipeline works reliably, we plot the number of detections against the distance to the corresponding balloon (see \reffig{fig:balloon_detection_histogram_dist}).
As ground truth, we use the last filtered position estimate immediately before the balloon is punctured.
We reliably detect balloons up to a distance of 24\,m, but can even detect balloons at a distance of up to 44.5\,m.
The corresponding pixel sizes are shown in \reffig{fig:balloon_detection_histogram_radius}.

\begin{table}
\caption{Assignments of all balloon detections to the different balloon hypothesis during Run~3.}
\label{tab:balloon_detection}
\centering
\begin{threeparttable}
\begin{tabular}{l|ccccc|c}
  \toprule
                & Balloon~1 & Balloon~2 & Balloon~3 & Balloon~4 & Balloon~5  & No Assignment \\
  \midrule
  \#Assignments & 58        & 229       & 117       & 252       & 248        & 45            \\
  \bottomrule
\end{tabular}
\end{threeparttable}
\end{table}

The number of detections at distances of 25--40\,m is still sufficient to generate valid hypothesis with our allocentric detection filter (\cf\refsec{sec:Allocentric_Balloon_Filter}).
Thus, balloon hypotheses were added to the world model shortly after takeoff during all runs. During the second run, all five balloons were known to the filter only 15\,s after takeoff. In the other runs, only a fraction of the balloons was inside the field of view directly after takeoff. The known target hypotheses are approached right away; Jelly then flies on a search pattern for a short time only until it detects the remaining targets.
During the Grand Challenge Run, 95.3\,\% of all detections could be successfully assigned to one of the five balloons.
The corresponding numbers are reported in \reftab{tab:balloon_detection}.

\begin{figure}
    \centering\hfill
    \begin{subfigure}[t]{.48\linewidth}
        \centering
%         \begin{tikzpicture}[font=\footnotesize\sffamily]
%             \node[inner sep=0] (image1) at (0, 0) {\includegraphics[width=\linewidth]{flightpathrun2completewithballoonsandtowers.png}};
%             \begin{scope}[x={($(image1.south east) - (image1.south west)$)},y={($(image1.north west) - (image1.south west)$)}, shift={(image1.south west)}]
%              \node at (0.418, 0.483) {1};
%              \node at (0.121, 0.445) {2};
%              \node at (0.788, 0.791) {3};
%              \node at (0.125, 0.788) {4};
%              \node at (0.768, 0.424) {5};
%              \node[anchor=north east] at (0.9, 0.24) {\tiny UAV};
%              \node[anchor=west,inner sep=0] at (0.949,0.13) {\fontsize{4pt}{4pt}\selectfont start};
%              \node[anchor=west,inner sep=0] at (0.949,0.87) {\fontsize{4pt}{4pt}\selectfont end};
%             \end{scope}
%         \end{tikzpicture}
        \includegraphics[width=1.0\linewidth]{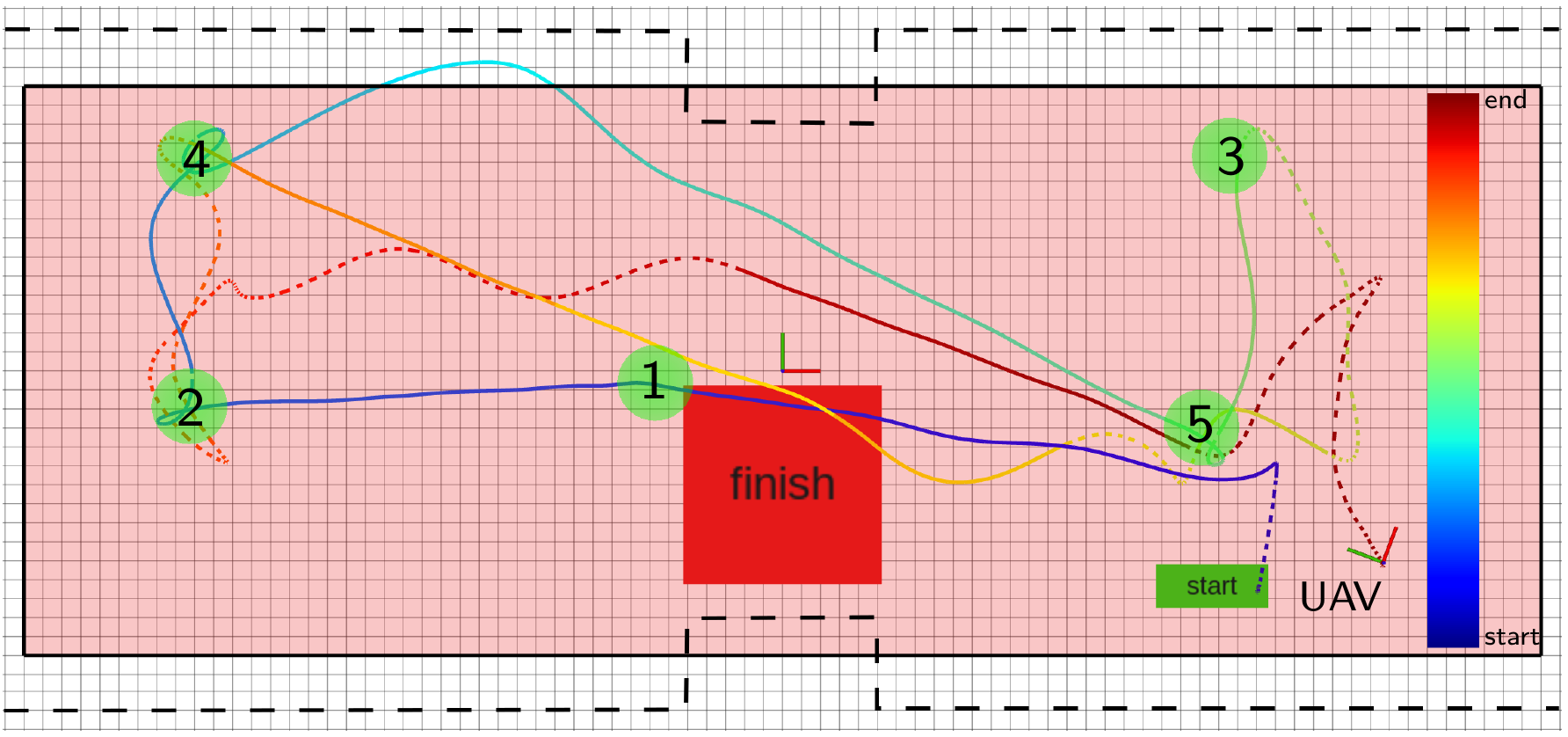}
        \caption{Run~2: The UAV chooses the direct path between consecutive balloons. In some cases, it turns directly above the balloons, which prevents the piercing tentacles from working correctly. Balloons 4 and 5 need to be passed two resp. three times.}
        \label{fig:flight_path_a}
    \end{subfigure}
    \hfill
    \begin{subfigure}[t]{.48\linewidth}
        \centering
%         \begin{tikzpicture}[font=\footnotesize\sffamily]
%             \node[inner sep=0] (image1) at (0, 0) {\includegraphics[width=\linewidth]{flightpathrun3withballoonsandtowers.png}};
%             \begin{scope}[x={($(image1.south east) - (image1.south west)$)},y={($(image1.north west) - (image1.south west)$)}, shift={(image1.south west)}]
%              \node at (0.855, 0.424) {1};
%              \node at (0.885, 0.74) {2};
%              \node at (0.413, 0.76) {3};
%              \node at (0.21, 0.74) {4};
%              \node at (0.206, 0.448) {5};
%              \node[anchor=north east] at (0.4, 0.48) {\tiny UAV};
%              \node[anchor=west,inner sep=0] at (0.949,0.13) {\fontsize{4pt}{4pt}\selectfont start};
%              \node[anchor=west,inner sep=0] at (0.949,0.87) {\fontsize{4pt}{4pt}\selectfont end};
%             \end{scope}
%         \end{tikzpicture}
        \includegraphics[width=1.0\linewidth]{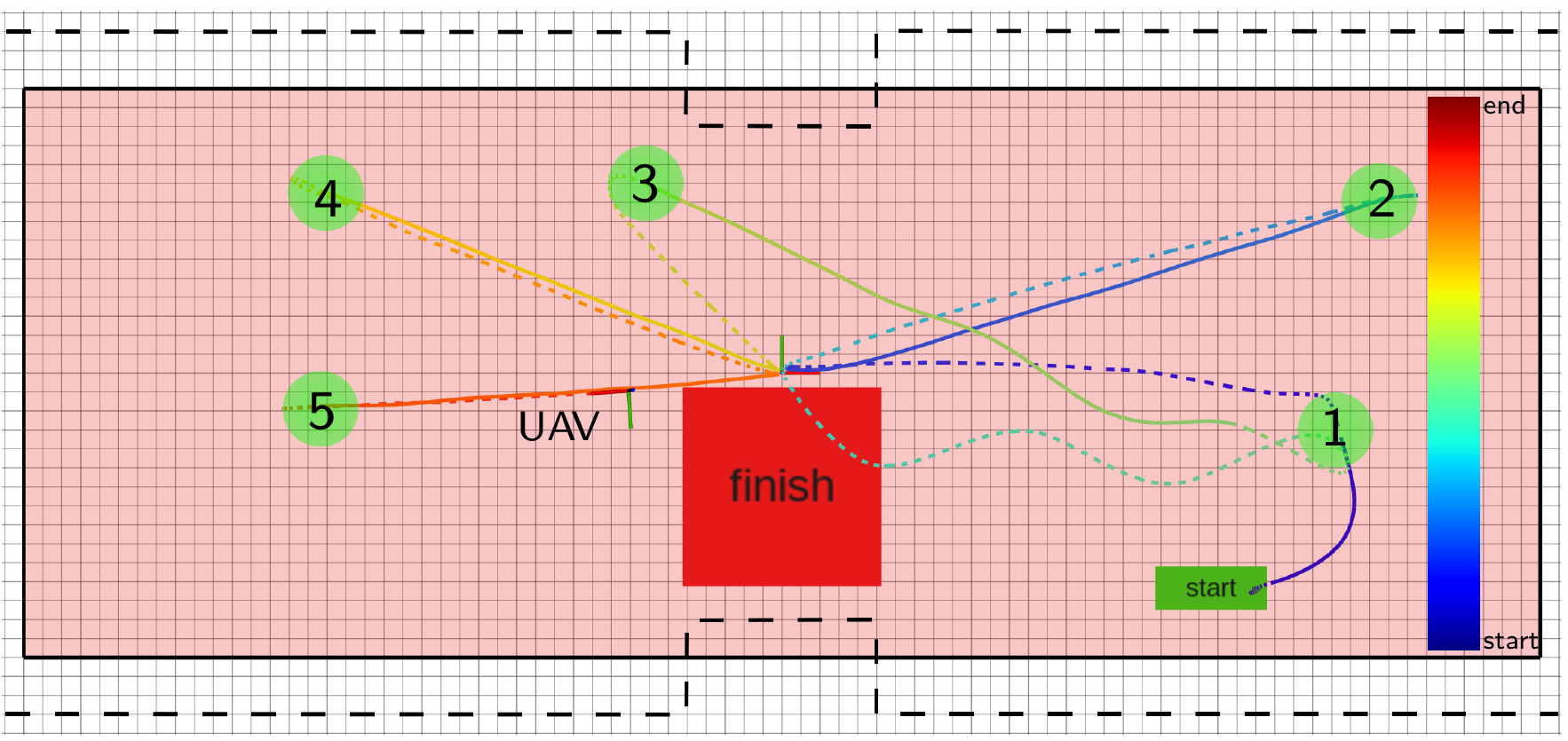}
        \caption{Run~3: The UAV passes through the arena center after each balloon. It moves straight through the balloons, which leads to them being pierced reliably at the first attempt. Paths close to the boundaries are avoided by this strategy.}
        \label{fig:flight_path_b}
    \end{subfigure}\hfill\strut
    \caption{Comparison of flight paths between Runs 2 and 3 (colored by time). Solid line: Pop mode, dashed line: Search mode. The allowed flying area is shaded in red; the physical arena boundaries are marked with a dashed black line. Balloon hypotheses are displayed as green circles.}
    \label{fig:flight_path}
\end{figure}

During the three different competition runs, we successfully punctured 15 balloons using only 18 tries. 
The times at which the respective balloons were punctured and the corresponding number of attempts are given in~\reftab{tab:timings}.
In the first run, two balloons were popped after 30\,s. Then a reset occurred for 8\,min, as Jelly had gotten stuck in the net at the arena borders due to an error in the GNSS-based geofencing. The challenge was completed after 9\,min~28\,s, but only 1\,min~28\,s flight time.
In the second run, the first two balloons were punctured immediately. Then, however, two balloons were missed---the puncturing did not work due to a suboptimal flight pattern (see~\reffig{fig:flight_path_a}). By repeating the search pattern and re-approaching the missed targets, in this run, all balloons were punctured after a total duration of 1\,min~40\,s.
In the final run during the Grand Challenge, all balloons were punctured in the first attempt. The time between two consecutive balloons was very similar (12--15\,s). Between Balloon 2 and 3, Jelly flew a search pattern to discover the remaining ones, explaining the longer time interval. The challenge was completed after a total time of 1\,min~21\,s.

During the first and second run, Jelly always chose the direct path between two consecutive balloons. In the ideal case, this results in the shortest duration between two consecutive balloons (\eg 6\,s between the first and second balloon in Run~2). However, this can lead to Jelly flying dangerously close to the arena borders. It could even have led to a crash during Run~2 if the controller had chosen to pierce Balloons~3 and 4 in direct sequence. Our simple GNSS-based geofencing system, which restricts the allowed flying area to a single rectangle, could \emph{not} correctly model the arena's non-convex shape (see~\reffig{fig:flight_path_a}).

\begin{table}
\caption{Time and number of tries needed to puncture the individual balloons.}
\label{tab:timings}
\centering
\begin{threeparttable}
\begin{tabular}{l|cc|cc|cc|cc|cc}
  \toprule
                & \multicolumn{2}{c|}{Balloon~1} & \multicolumn{2}{c|}{Balloon~2} & \multicolumn{2}{c|}{Balloon~3} & \multicolumn{2}{c|}{Balloon~4} & \multicolumn{2}{c}{Balloon~5}\\
                & Time             & Tries & Time        & Tries & Time                        & Tries & Time                                    & Tries & Time                      & Tries\\
  \midrule
  Run~1$^*$     & 23\,s & 1 & 30\,s & 2 & 9\,min~4\,s & 1 & 9\,min~11\,s            & 1 & 9\,min~28\,s & 1 \\
  Run~2         & 17\,s & 1 & 23\,s & 1 &       51\,s & 1 & 1\,min~15\,s            & 2 & 1\,min~40\,s & 2 \\
  Run~3         & 11\,s & 1 & 27\,s & 1 &       56\,s & 1 & 1\,min~\hphantom{0}8\,s & 1 & 1\,min~21\,s & 1\\
  \bottomrule
\end{tabular}
\quad($^*$) 8\,min reset time between 2nd and 3rd balloon.
\end{threeparttable}
\end{table}

An additional waypoint was added in the middle of the arena after each balloon for the final run. This results in a star-shaped flight pattern (see~\reffig{fig:flight_path_b}) and the balloons being passed in a straight line without turning above them. Consequently, each balloon was pierced in the first attempt. The time between two consecutive balloons is slightly higher than it was before in the ideal case but almost constant for each target, as no misses occur.
Moreover, the star-shaped flight pattern avoids trajectories close to the arena borders and leads to safe flight paths despite the non-convex arena outline without any additional obstacle avoidance system (\cf\refsec{sec:Mission_Control_State_Machine}).

\begin{figure}
  \centering
%   \begin{tikzpicture}[
%      b/.style={inner sep=0},
%     ]
%     \definecolor{red}{rgb}{0.7,0.0,0.0}
%     \node[b] (image1) {\includegraphics[frame,trim=3mm 28mm 125mm 25mm,clip,width=0.4\linewidth]{1.png}};
%     \node[b,right=0.1cm of image1] (image2) {\includegraphics[frame,trim=3mm 28mm 125mm 25mm,clip,width=0.4\linewidth]{2.png}};
%     \node[b,below=0.1cm of image1] (image3) {\includegraphics[frame,trim=3mm 28mm 125mm 25mm,clip,width=0.4\linewidth]{3.png}};
%     \node[b,right=0.1cm of image3] (image4) {\includegraphics[frame,trim=3mm 28mm 125mm 25mm,clip,width=0.4\linewidth]{4.png}};
% 
%     \begin{scope}[x={($(image1.south east) - (image1.south west)$)},y={($(image1.north west) - (image1.south west)$)}, shift={(image1.south west)}]
%         \draw[line width=0.5mm, red] (0.75,0.53) circle (0.5cm and 0.5cm);
%     \end{scope}
%     \begin{scope}[x={($(image2.south east) - (image2.south west)$)},y={($(image2.north west) - (image2.south west)$)}, shift={(image2.south west)}]
%         \draw[line width=0.5mm, red] (0.55,0.5) circle (0.5cm and 0.5cm);
%     \end{scope}
%     \begin{scope}[x={($(image3.south east) - (image3.south west)$)},y={($(image3.north west) - (image3.south west)$)}, shift={(image3.south west)}]
%         \draw[line width=0.5mm, red] (0.39,0.59) circle (0.5cm and 0.5cm);
%     \end{scope}
%     \begin{scope}[x={($(image4.south east) - (image4.south west)$)},y={($(image4.north west) - (image4.south west)$)}, shift={(image4.south west)}]
%         \draw[line width=0.5mm, red] (0.55,0.5) circle (0.5cm and 0.5cm);
%     \end{scope}
%   \end{tikzpicture}
  \includegraphics[width=0.8\linewidth]{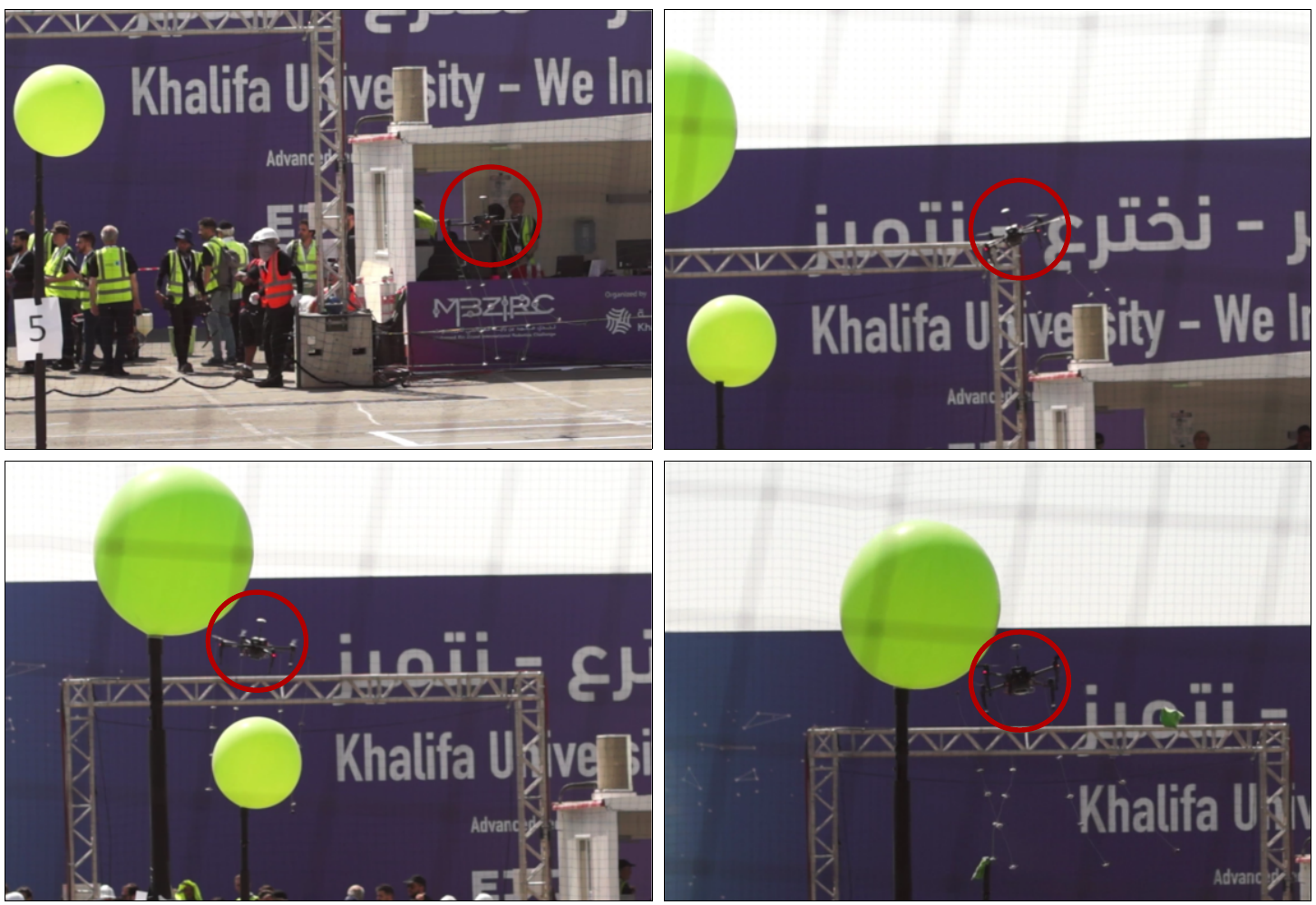}
  \caption{Image sequence of popping the first balloon in the Grand Challenge. Jelly (marked with the red circle) 1) takes off to 4.0\,m. 2) After detecting the first balloon, the robot targets a position 2.0\,m behind and 0.7\,m above balloon's center. 3) It further accelerates to pass through the balloon with significant velocity. 4) It successfully pops the balloon. The entire shown process only takes 4.9\,s.}
  \label{fig:popping_sequence}
\end{figure}

See \reffig{fig:popping_sequence} for an exemplary balloon popping sequence. Overall, we placed 5th in Challenge~1, including the Ball Interception Challenge, and 2nd in the Grand Challenge, including five other sub-challenges.

\section{Moving Target Interception}
\label{sec:Moving_Target_Interception}
The second task of the MBZIRC 2020 Challenge~1 required a robot to intercept a moving target UAV in an outdoor arena of size 90\,$\times$\,40\,m. A yellow ball with approximately 13\,cm diameter was attached below the target UAV, which moved with up to 10\,m/s on a 3D figure-eight trajectory in the arena. The teams had to track the target, detach the ball without damaging the target UAV, and finally deliver it to a drop-off zone.

The task mainly promoted the development of advanced (visual) perception methods, fast and precise UAV control, and aerial manipulation capabilities that are beneficial for real-world applications like \eg drone defense.

Our plan was to command our UAV to wait on a corner point of the figure-eight with the camera directed towards the field center and search for the ball. After the target UAV passes from behind and the ball is detected, our UAV pitches to the maximum allowed magnitude of 35° to gain speed on the figure-eight's diagonal to intercept the ball.

\subsection{Hardware}
\label{sec:chaser_hw}
\begin{figure}
  \centering
  \includegraphics[width=1.0\linewidth]{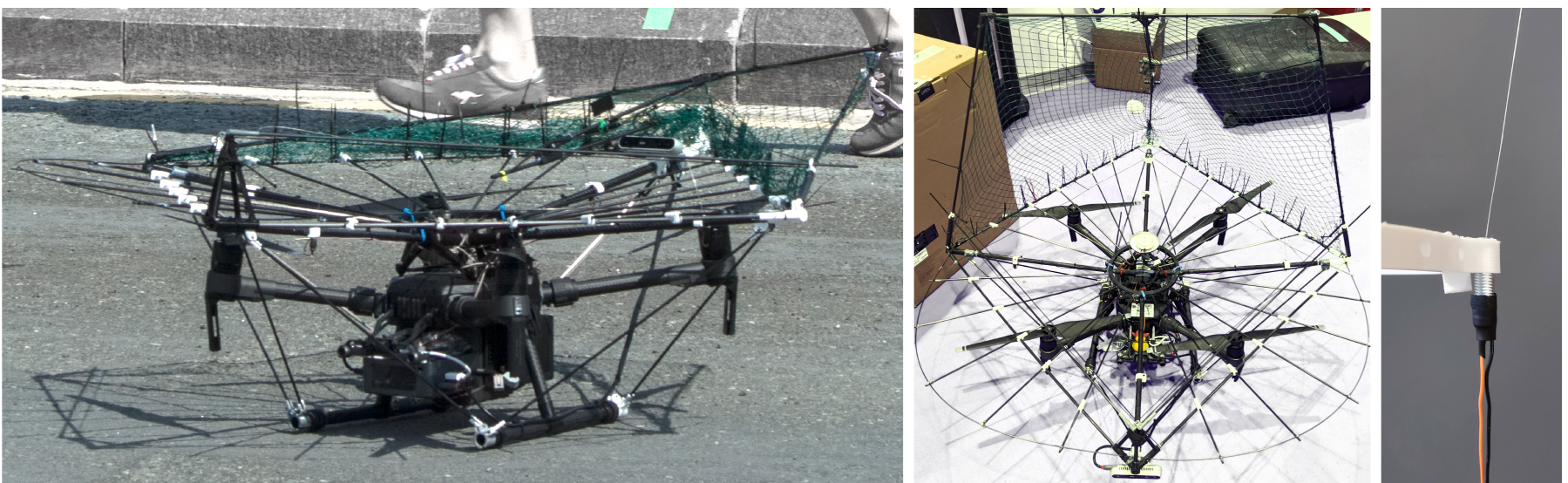}~
  \caption{Design of our UAV ``Chaser'' equipped with a foldable net to capture the target, a fast onboard computer, an Intel\textsuperscript{\textregistered} RealSense\textsuperscript{\texttrademark} D415 camera and a Google Edge TPU.}
  \label{fig:chaser}
\end{figure}

For this task, we designed our integrated UAV ``Chaser'' shown in \reffig{fig:Chaser}, which is based on the DJI Matrice~210~v2 platform. It is equipped with a compact, but fast onboard PC (Intel\textsuperscript{\textregistered} NUC), an Intel\textsuperscript{\textregistered} RealSense\textsuperscript{\texttrademark} D415 depth camera for visual perception, and a Google Edge TPU USB accelerator for visual inference. The target ball is detached and caught by a foldable net, which, in a folded state, respects the required maximum takeoff size and, after takeoff, automatically extends by a spring release to offer a large volume for catching the target ball.
The overall mechanism is shown in \reffig{fig:chaser} and attaches on top of our UAV. Once the ball is caught by the net, it will drop onto the funnel-like contraption with radial struts and roll towards the center where multiple omnidirectional switches were installed. The switch consists of spring steel wire within a metal tube. Upon contact, the wire pushes against the tube and closes the circuit and an Arduino Nano micro controller will signal the onboard computer. 
To ensure safe interaction with the ball's attachment cable, the struts connect outwards to a round prop-guard steering the cable away from the propeller. Additional connections between the contraption and the landing gear improves stability and prevents oscillation of the flexible construction. 
The outer net struts can rotate backwards and together with the hinge in the middle of the central strut; the net becomes foldable. A spring at the central strut tensions and unfolds these struts. We prevent unfolding with a small chord from the strut to below the landing gear. During takeoff, the chord releases automatically and the spring unfolds and tensions the net. Zip ties attach the net to its frame while zip tie pairs at the bottom frame pointed upwards prevent the net from being dragged into the rotors before unfolding. 
Initially, we mounted the D415 at the rear-end on the funnel below the net, but moved it to the front due to USB interfering with the GNSS and the contraption being partially visible in the camera image. Since onboard GNSS remained unusable, we added in the middle a separate DJI N3 flight controller with protruding antenna and extra shielding below.

\subsection{Target Detection}
\label{sec:Target_Detection}
\begin{figure}
  \centering
  \includegraphics[trim=70mm 20mm 50mm 30mm,clip,width=0.32\textwidth]{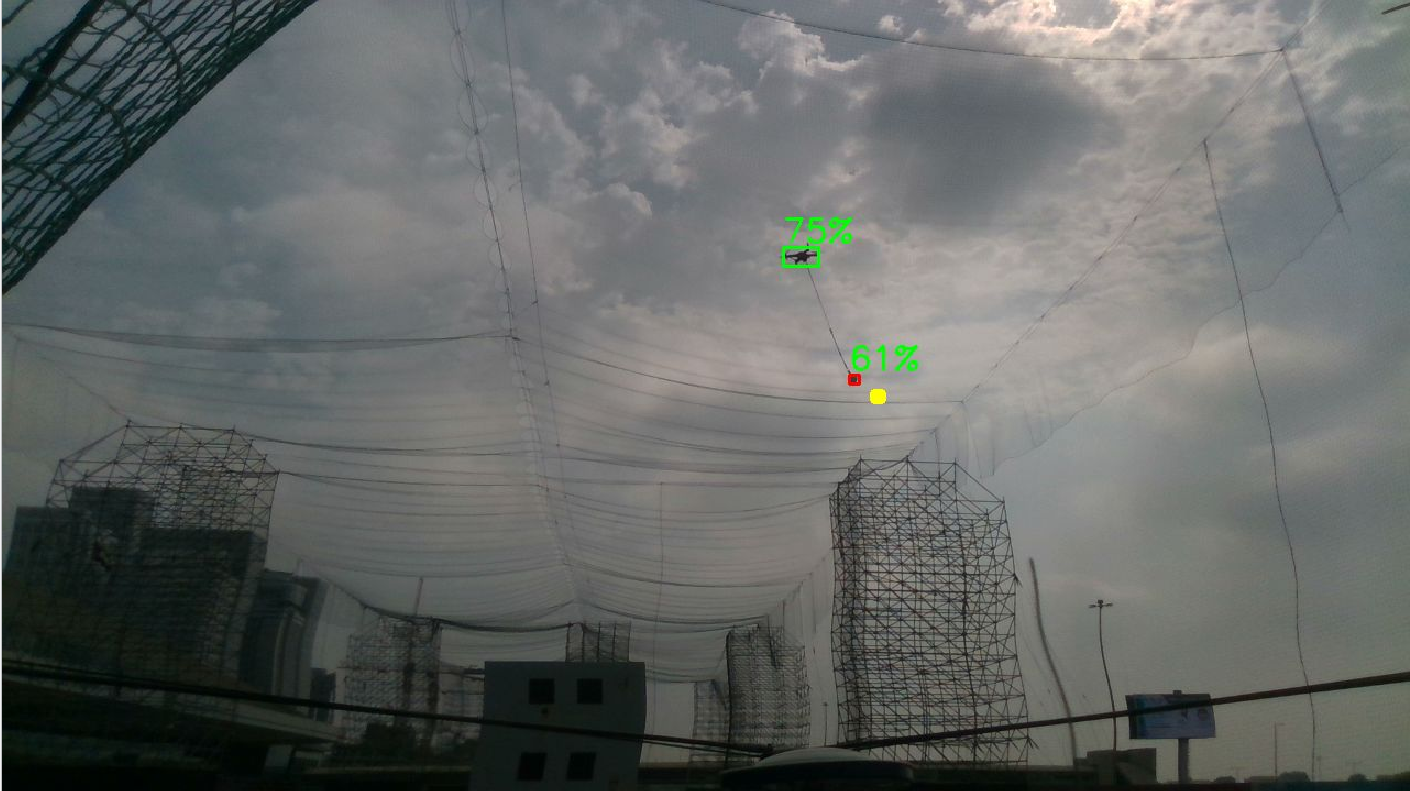}\,
  \includegraphics[trim=70mm 20mm 50mm 30mm,clip,width=0.32\textwidth]{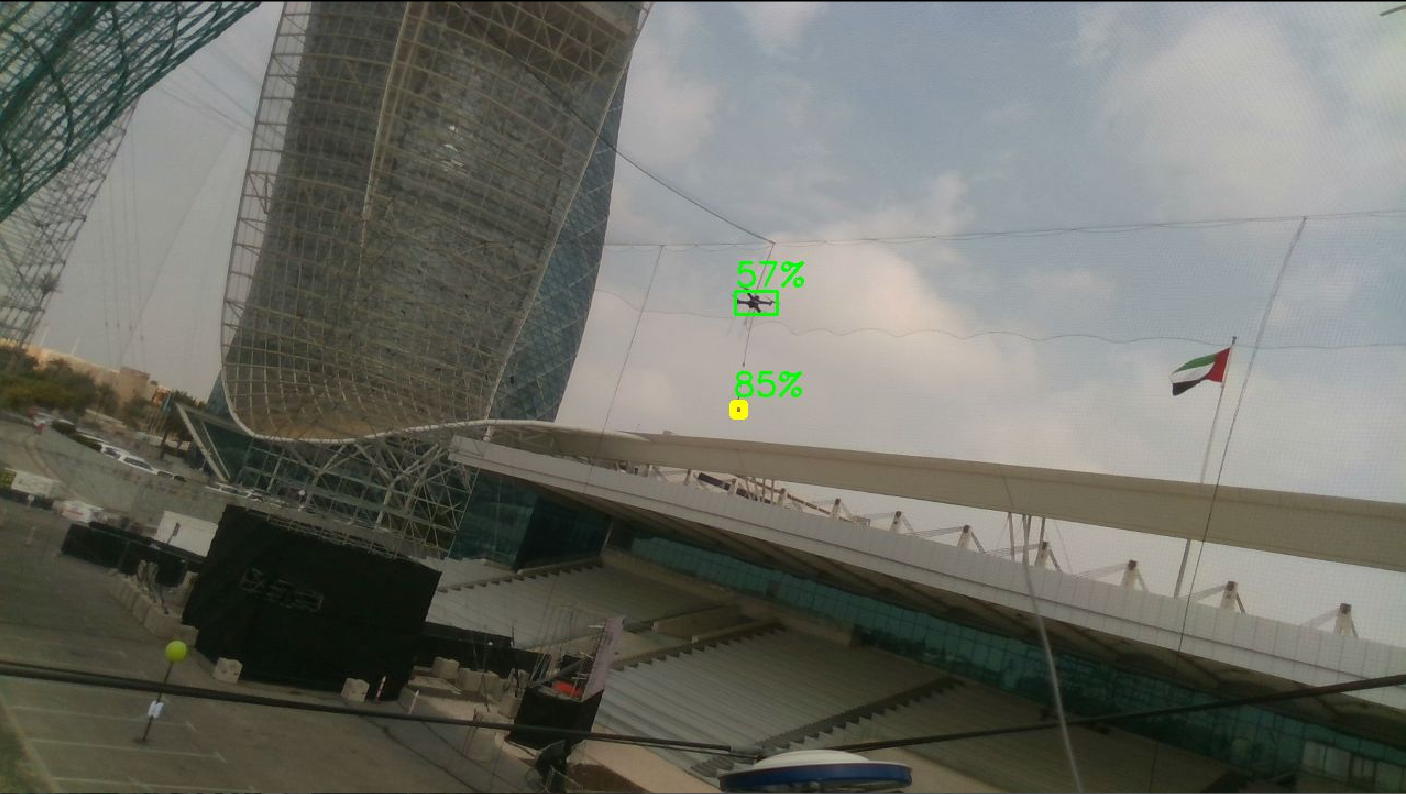}\,
  \includegraphics[trim=70mm 20mm 50mm 30mm,clip,width=0.32\textwidth]{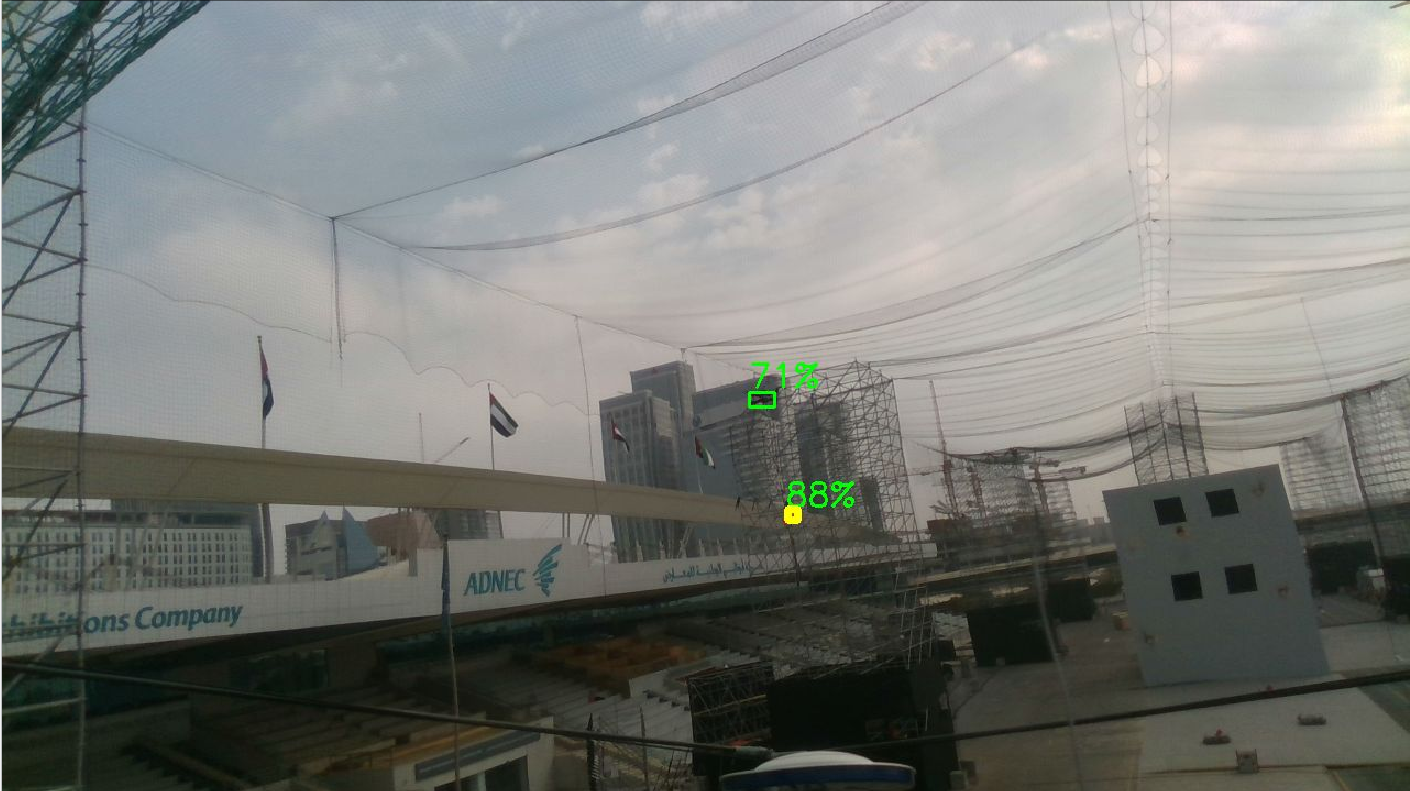}\,
  \caption{Detections of the target UAV (green bounding box) and ball (red bounding box) in the RGB image, with detection probabilities. The yellow bounding box shows the detection of the tracked ball from the previous frame, used to determine the cropped region of the current frame passed to the detector.}
  \label{fig:copter_ball_det}
\end{figure}

We employ a deep learning-based vision pipeline for the perception of the moving target. RGB images of the RealSense\textsuperscript{\texttrademark} camera are processed by a detector based on the lightweight SSD architecture \citep{liu2016ssd} using an efficient MobileNetV2 \citep{mobilenetv2_2018} backbone and provides bounding box detections for the yellow ball and the drone (see \reffig{fig:copter_ball_det}). The network input resolution is set to 848\,$\times$\,480\,px, processing cropped regions of the camera images of 1280\,$\times$\,720\,px resolution. The region of interest to pass to the detector is determined based on the currently tracked detection from the previous frame. When no hypothesis is known, the entire image is analyzed in four separate tiles.

\begin{figure}
  \centering
  \includegraphics[width=0.75\linewidth]{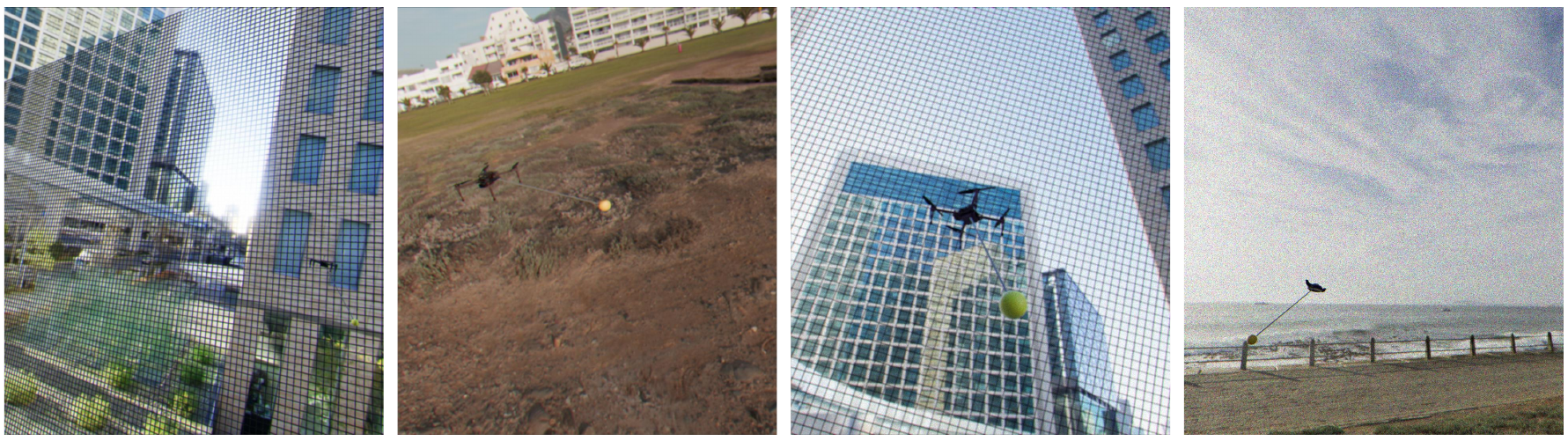}~
  \caption{Synthetic training images for target detection of drone and yellow ball.}
  \label{fig:targets}
\end{figure}

The detector was trained before the competition on synthetic images of the target UAV and the ball, and fine-tuned with a few background images from the real arena on site. To prevent the network from learning to only detect dark patches in the sky, like cranes, we added negative training examples of objects like tea pots, submarines and lamps with an attached yellow ball. In total, we used approximately 20\,k synthetic and 1000 real images. Some examples are depicted in \reffig{fig:targets}.

The vision model runs in 8-bit quantized mode on the Edge TPU USB accelerator, achieving the full camera frame rate of 30\,Hz, thus sufficient for fast and reactive flight control based on the detections. The distance towards the detected UAV and ball is estimated by fusing the depth reading from the RealSense\textsuperscript{\texttrademark} RGB-D camera with the distance estimated from the known diameter of the target ball and the camera parameters.

We want to stress that although we use our detection pipelines for detecting objects relevant to the MBZIRC 2020, our practical method and the theoretical contribution it is based on are \emph{not} limited to these demonstrations. In contrast to classic computer vision approaches that are often custom-tailored for detecting specific patterns like the landing pattern during MBZIRC 2017~\citep{beul2019team}, our pipeline makes no assumption about the specific object it detects. Here again, scientific contributions transfer to real-life applications that result in added value.

\subsection{Allocentric Detection Filter}
\label{sec:Allocentric_Detection_Filter}
The vision model outputs lists of copter and ball detections for each image frame.
As already discussed for the Balloon Hunt Challenge (\cf \refsec{sec:Allocentric_Balloon_Filter}), the depth estimation of the visual perception model can be noisy.
However, precise allocentric position information and additional velocity estimates are necessary to successfully intercept the target.
Thus, we process our detections with an allocentric filter before forwarding them to the Mission Control module.
Our filter module consists of two steps: First, the most probable ball and copter detections are selected from the detection list based on the previous estimates. Afterwards, an Extended Kalman Filter (EKF) is applied to the position information of the detections to generate velocity estimates.

When selecting the most probable ball and copter detections, we first remove outliers by only considering copter detections if they are within a distance of 5\,m around the previous estimate.
From those, we choose the one with highest confidence score as provided by the detector.
Afterwards, we process the ball detections in a similar way but utilize the fact that the ball has to be close to the target copter.
Thus, if a valid copter detection for the current frame was found, we consider all ball detections within a distance of 5\,m around the current copter detection.
Otherwise, we search for a ball detection close to the previous estimated ball position.
If multiple valid detections exist, we select the one with highest confidence score.

Differentiating the yellow ball from the green balloons of the first Challenge~1 task (\cf \refsec{sec:Balloon_Hunt}) is very challenging for our vision pipeline.
Thus, we additionally search for all ball detections that might actually correspond to balloons.
We estimate the distance between camera and each ball detection using the size of the bounding box and the known balloon diameter.
If the projection into allocentric 3D coordinates results in a feasible balloon position, \ie if the height is within a corridor from 1.5 to 5.0\,m, we probably misclassified a balloon as ball, and thus discard the detection.

The selected detections are processed by two independent instances of an EKF, one for ball and copter detections, respectively.
Here, we utilize the implementation from the robot\_localization library~(\cite{moore2014ekf}).
The EKF output is forwarded to the Mission Control module.
Additionally, the current estimates are used by our vision pipeline to crop a region of interest from the camera images, as described in \refsec{sec:Target_Detection}.

\begin{figure}
  \centering
  \includegraphics[width=0.9\textwidth]{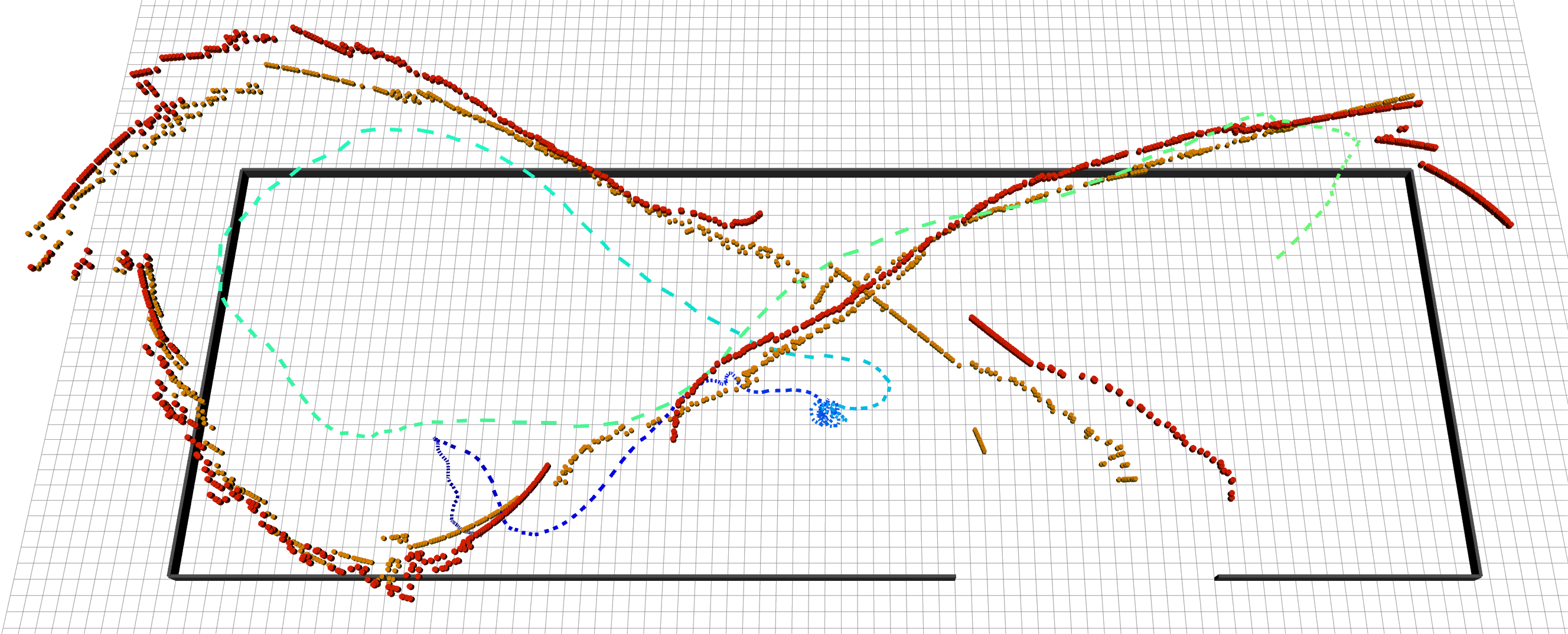}
  \caption{Tracked trajectory of target UAV (red) and ball (yellow), flight path of our UAV (dashed line), GNSS-based geofencing (gray). After takeoff, our UAV ascends to the search height and flies towards a waiting point (dark-blue dashed line). After the first detection of the target UAV and ball (bottom right), our UAV turns and follows the targets. The chase is abandoned at the top-right corner, as our UAV needs to brake before the turn. The three-dimensional figure-eight shape of the target's trajectory is clearly discernible.}
  \label{fig:copter_ball_traj}
\end{figure}

\subsection{Evaluation}

\begin{table}
\small
\caption{Evaluation of the vision pipeline for the moving target interception task.}
\label{tab:evaluation_copter_ball_detection}
\vspace{-1.0em}
\begin{center}
\setlength{\tabcolsep}{1.9mm}
\begin{tabular}{c|cccc|ccc}
  \toprule
         & \#Frames & \#True    & \#False   & \#No      & \#Frames   & \#No      & \#False   \\
         & Visible  & Detection & Detection & Detection & Invisible  & Detection & Detection \\
  \midrule 
  Copter & 566      & 456       & 15        & 95        & 738        & 707       & 31        \\
  Ball   & 557      & 450       & 15        & 92        & 747        & 718       & 29        \\
  \bottomrule
\end{tabular}
\end{center}
\end{table}

\begin{table}
\small
\caption{Evaluation of the filter accuracy for distinguishing ball and balloon detections.}
\label{tab:evaluation_balloon_misdetection_filter}
\vspace{-1.0em}
\begin{center}
\setlength{\tabcolsep}{1.9mm}
\begin{tabular}{l|cc}
  \toprule
   True Label  & \# Classified as ``Balloon'' & \# Classified as ``No Balloon''  \\
  \midrule 
  ``Balloon''      & 43                       & 8              \\
  ``No Balloon''   & 8                        & 490           \\
  \bottomrule
\end{tabular}
\end{center}
\end{table}

To evaluate the performance of our perception pipeline, we manually check the detections for a portion of an autonomous flight during one challenge run.
The analyzed sequence has a duration of 43\,s and 1304 frames. 
It starts with the moment when Chaser reaches the initial search pose and ends when Chaser aborts the first chase by breaking on the end of the figure eight's diagonal.
If multiple detections for the same frame exist, we only consider the one with highest confidence score.
The performance is similar for copter and ball detections (see \reftab{tab:evaluation_copter_ball_detection}).
The perception pipeline works robustly and detects the moving target in about 80\,\% of all cases where it is visible.
Only in about 3.5\,\% of all analyzed frames does the most confident detection not correspond to the target copter or ball.
Some example images, where the moving target is detected at a considerable distance even in front of cluttered backgrounds, are shown in \reffig{fig:copter_ball_det}.

Distinguishing the yellow target ball from the green balloons of the other Challenge~1 task was challenging under the harsh sunlight.
During the analyzed sequence, our vision pipeline reports 594 ball detections, of which 51 actually correspond to balloons.
However, our allocentric filter (\refsec{sec:Allocentric_Detection_Filter}) is able to detect those wrong classifications in most of the cases, as shown in \reftab{tab:evaluation_balloon_misdetection_filter}.
For both, precision and recall the filter achieves scores of 84\,\%.
Note, that in contrast to \reftab{tab:evaluation_copter_ball_detection}, we count all ball detections per frame and not only the most confident one.

Based on the filtered detections, the target's figure-eight trajectory could be tracked in 3D space, and flight trajectories were generated to follow the target UAV and accelerate below it to detach the target ball (see \reffig{fig:copter_ball_traj}). 
Unfortunately, in all trials, the acceleration was insufficient to gain enough speed until Chaser had to break to prevent flying into the net on the end of the diagonal. In multiple tries, Chaser reliably achieved velocities exceeding the 10\,m/s of the target UAV and also came close to it but never managed to intercept the ball before breaking. With 6.2\,kg and 1.1\,$\times$\,1.1\,$\times$\,1.0\,m size and due to the wind resistance of the large netting construction on top, Chaser was simply not agile enough for the proposed task.

3\,min before the end of our run, we decided to abort the autonomous attempts and salvage points by manually approaching the target. For this, we prepared a bare DJI Matrice~100 in advance and removed all unnecessary hardware to reduce weight and increase agility. Although the UAV featured no catching hardware, the intention was to at least detach the ball from the target UAV, which would yield at least partial points.

With the stripped-down UAV, our trained safety pilot was able to manually detach the ball with only 8\,s flight time, securing a 5th place in Challenge~1 together with the balloon-hunting task.

\section{Wall Construction}
\label{sec:lofty}
In Challenge~2, participants were required to build walls out of supplied bricks, both with a UGV and a team of up to three UAVs. The task setting was particularly interesting because it required complete autonomy, robustness under real-world outdoor conditions with harsh sunlight and wind, and independence from any external reference system besides the globally available GPS.

In the following, we will describe our UAV ``Lofty''. For more detailed information about Lofty, we refer to \citet{lenz2020ssrr}.

Since the initial rules specified a shared wall where UAVs and UGV could collaborate, we concentrated our efforts on the UGV design. In a late rule revision, UAV and UGV walls were separated, making it clear to us that UAV points had to be scored in order to win. Thus, our UAV design focused on a minimal solution that could achieve almost full points: We decided to ignore the orange bricks of 1.8\,m length, intended to be carried by two UAVs. We designed our system to only support the red (0.3\,m), green (0.6\,m), and blue (1.2\,m) bricks.

\subsection{Hardware Design}
\label{sec:lofty:hw}

Because of the weight of the larger bricks and their size, we decided to use a large UAV, the DJI Matrice~600, for this task. The DJI Matrice~600 offers sufficient payload and battery life (roughly 20\,min in our configuration).

\begin{figure}
  \sbox\twosubbox{%
    \resizebox{\dimexpr.95\textwidth-1em}{!}{%
      \includegraphics[height=3cm,clip,trim=000 020 000 100]{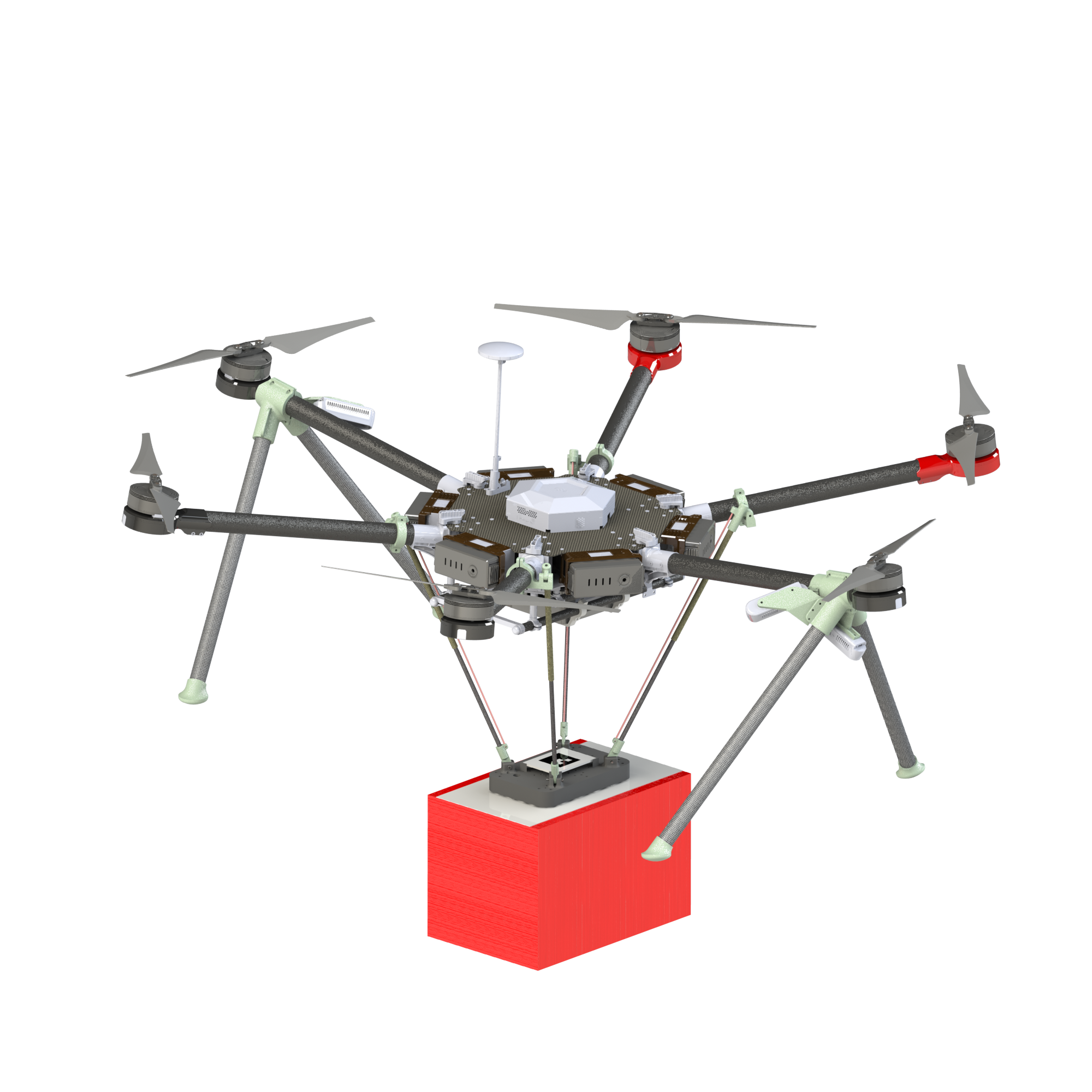}%
      \includegraphics[height=3cm,clip,trim=050 070 020 020]{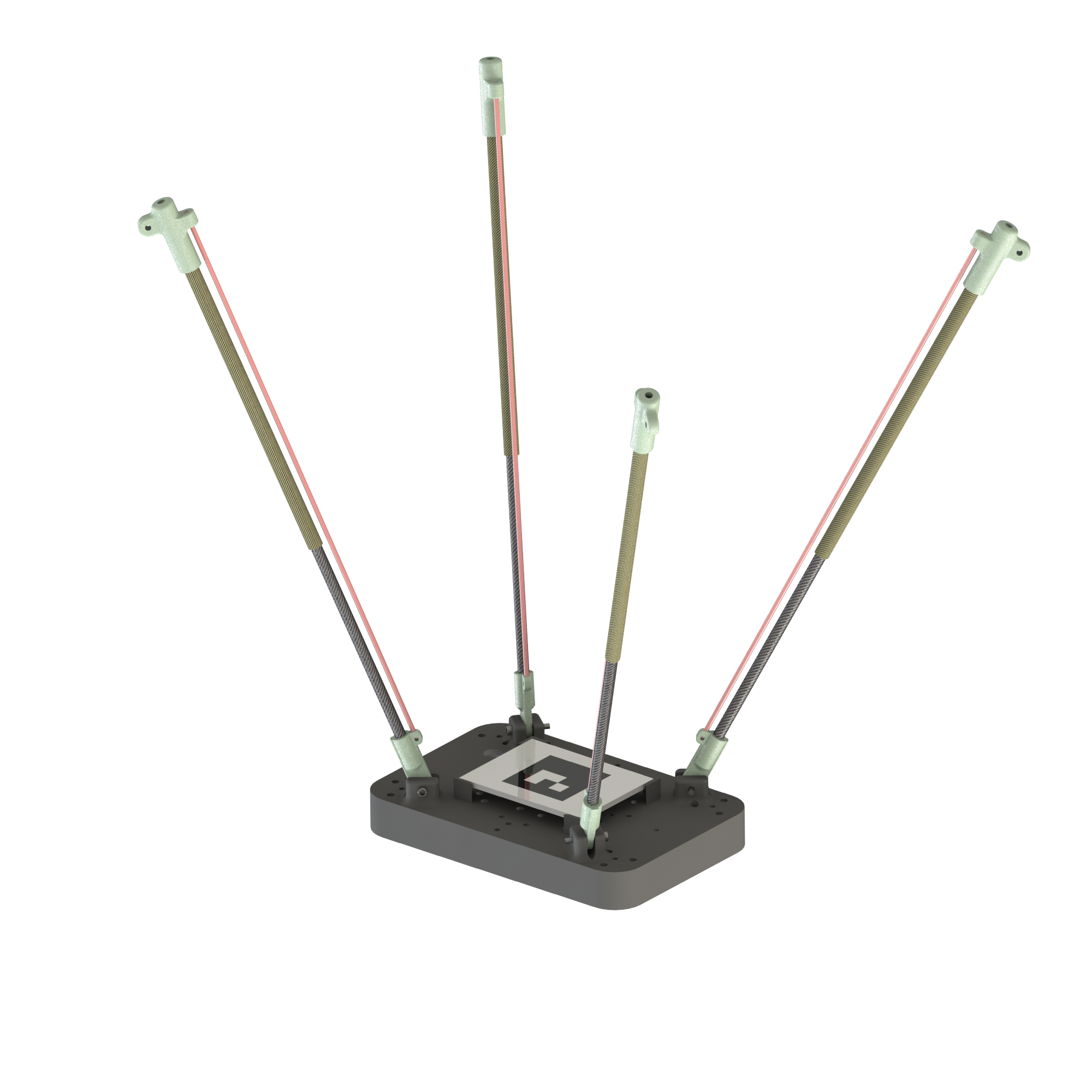}%
    }%
  }
  \setlength{\twosubht}{\ht\twosubbox}
  \centering
  \subcaptionbox{Full assembly}{%
    \includegraphics[height=\twosubht,clip,trim=000 020 000 100]{cadlofty2.png}%
  }
  \subcaptionbox{Magnetic gripper with passive compliance}{%
    \includegraphics[height=\twosubht,clip,trim=050 070 020 020]{cadgripper.png}%
  }
  \caption{Wall building UAV hardware design. The four telescopic rods are shown in the fully extended configuration.}
  \label{fig:lofty_cad}
\end{figure}

A key component for aerial manipulation is the robotic gripper. UAVs pose unique constraints when compared with ground-based manipulation. The gripper has to be lightweight in order to fit inside the payload constraints. Furthermore, a certain flexibility and mechanical compliance is desired for two reasons: First, this allows a grasp to succeed even if the approach was not fully precise. Secondly, a rigid connection between the UAV and the ground can be perilous since UAVs usually tightly and dynamically control their attitude to hold position. One can easily imagine situations where the UAV has to drastically change attitude in response to wind gusts, and of course, hindrance by the gripper system should be limited. However, during the placement phase of the pick-and-place operation, we require exact control of the target object. Here---at least while the target object is still in the air---we want a rigid attachment to the UAV. To resolve these seemingly contradicting goals, we designed the gripper system to be rigid only while load is applied, \ie the brick is hanging below Lofty.

Our gripper design (see \reffig{fig:lofty_cad}) consists of four carbon fiber telescopic rods, which hold the magnetic gripper plate below Lofty. When the rods are fully extended, the gripper
plate is in a fixed pose and can only move upwards. The more the gripper plate is pressed upwards (\eg due to contact with a brick), the more it can move sideways and rotate due to the gained movement range in each rod. The gripper is equipped with a switch to detect successful grasping.

Since the standard foldable landing legs on the DJI Matrice~600 would interfere with the gripper, we replaced them with fixed landing legs (see \reffig{fig:lofty_cad}).

\subsection{Brick Perception}

\begin{figure*}
 \centering
%  \begin{tikzpicture}[
%     font=\footnotesize\sffamily\footnotesize,
%     img/.style={draw=black, rounded corners, align=center, fill=yellow!20}
%   ]
%   \node[img] (input) {a) RGB Input\\
%   \parbox{.1\linewidth}{%
% 	\includegraphics[width=\linewidth]{cam1.png}
% 	\includegraphics[width=\linewidth]{cam2.png}
%   }\hspace{.11cm}%
%   \parbox{.1\linewidth}{%
%     \includegraphics[width=\linewidth]{input.png}
%   }
%   };
%   \node[img,right=0.4cm of input] (segmentation) {b) Patch Segmentation\\\includegraphics[width=0.21\linewidth,frame]{segmentation.png}};
%   \node[img,right=0.4cm of segmentation] (vis) {c) Patch Extraction\\\includegraphics[width=0.21\linewidth]{vis.png}};
%   \node[img,right=0.4cm of vis] (rviz) {d) Allocentric Tracking\\\includegraphics[width=0.21\linewidth,frame]{rvizcrop.png}};
%   \draw[-latex,very thick] (input) -- (segmentation);
%   \draw[-latex,very thick] (segmentation) -- (vis);
%   \draw[-latex,very thick] (vis) -- (rviz);
%  \end{tikzpicture}
 \includegraphics[width=1.0\linewidth]{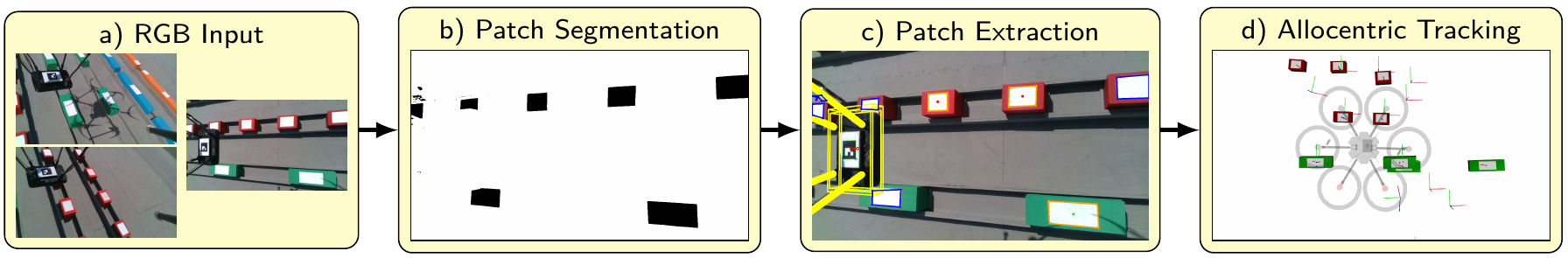}
 \caption{Camera-based UAV brick perception pipeline.
   a)~Input frames from all cameras.
   b)~White patch segmentation.
   c)~Patch corner extraction \& pose estimation.
     Patch contours in orange (verified) and blue (wrong shape).
     Brick type is indicated by a colored center point.
     The gripper is overlaid in yellow.
   d)~Tracking of detections from all three cameras in GPS frame. Detections
     are shown as bricks, while tracked hypotheses are shown as coordinate axes.
 }
 \label{fig:lofty_perception}
\end{figure*}

The competition task involves two perception challenges: finding and precisely localizing the bricks and localizing with respect to the target wall. Similar to the gripper system, the UAV places unique constraints on the perception system. Since the gripper is mounted directly beneath Lofty, any brick observation close to the gripper must be conducted from the side. The necessary off-center mounting of the sensors severely limits the sensor weight. We chose the Intel\textsuperscript{\textregistered} RealSense\textsuperscript{\texttrademark} D435 RGB-D camera as a primary sensor for its low weight and its capability to work in sunlight. To achieve good coverage of the terrain below Lofty and to be able to observe large parts of the wall during the placement process, we mounted three D435 sensors (see \reffig{fig:lofty_cad} and \ref{fig:lofty_perception}).

The gripper is visible in all camera images and would lead to confusion with bricks. For this reason, we mounted an ArUco marker~\citep{romero2018speeded} on it. The marker pose can be efficiently estimated in each of the three cameras and is low-pass filtered to obtain a robust estimate of the gripper pose below Lofty. We exclude pixels in the immediate vicinity of the detected gripper from further processing.

Since the white patches on the bricks are quite distinctive (see \reffig{fig:lofty_perception}), we use them to detect the bricks and estimate their pose. In a first step, we convert the input image (resolution 1280\,$\times$720\,px) to the HSV color space. To detect high-saturation pixels (the colored bricks) in the neighborhood, we downsample the input image to half resolution and run a box filter with kernel size 290\,$\times$\,290\,px to obtain a local saturation average $\bar{S}$ and local value average $\bar{V}$. A pixel $p$ is classified as a \textit{patch} if $S(p) < \bar{S}(p) - \lambda_S  \wedge  V(p) > \bar{V}(p) + \lambda_V$, or, in other words, the saturation is less than the local average and the value (brightness) is larger than the local average by user-specified thresholds. This simple segmentation method is modeled after the ones used for detecting chessboard patterns and leads to highly robust performance (see \reffig{fig:lofty_perception}).

Contours with exactly four corners (after contour simplification) are processed further: We check that each corner has a \textit{patch} pixel on the inside and a high-saturation pixel on the outside at a specified distance of $d = 4\,\mathrm{px}$. The high-saturation pixels on the outside are independently classified into the four possible colors. If all agree, the brick is detected. Finally, a PnP solver is used to determine the 6D pose of the brick from the recovered 2D-3D correspondences. Here, we assume that the longer side in the 2D image corresponds to the longer brick side in 3D---an assumption, which is only violated at extreme viewing angles. To fuse the detections from all three cameras and track bricks over time, we apply a basic Multi-Hypothesis Tracking (MHT) method with one Kalman filter
per hypothesis.

\subsection{Brick Placement}
\label{sec:Brick_Placement}

\begin{figure}[t]
  \centering
  \begin{subfigure}[t]{0.45\linewidth}

%   \begin{tikzpicture}
%   \node[inner sep=0] (image1) at ( 0.00, 1.30) {\includegraphics[trim=00mm 00mm 00mm 00mm,clip,width=0.55\linewidth]{scene2cam2.png}};
%   \node[inner sep=0] (image2) at ( 0.00,-1.30) {\includegraphics[trim=00mm 00mm 00mm 00mm,clip,width=0.55\linewidth,angle=180]{scene2cam1.png}};
%   \node[inner sep=0] (image3) at ( 3.78, 0.00) {\includegraphics[trim=00mm 00mm 00mm 00mm,clip,width=0.35\linewidth]{scene2briorot.png}};
%   \end{tikzpicture}
    \includegraphics[width=1.0\linewidth]{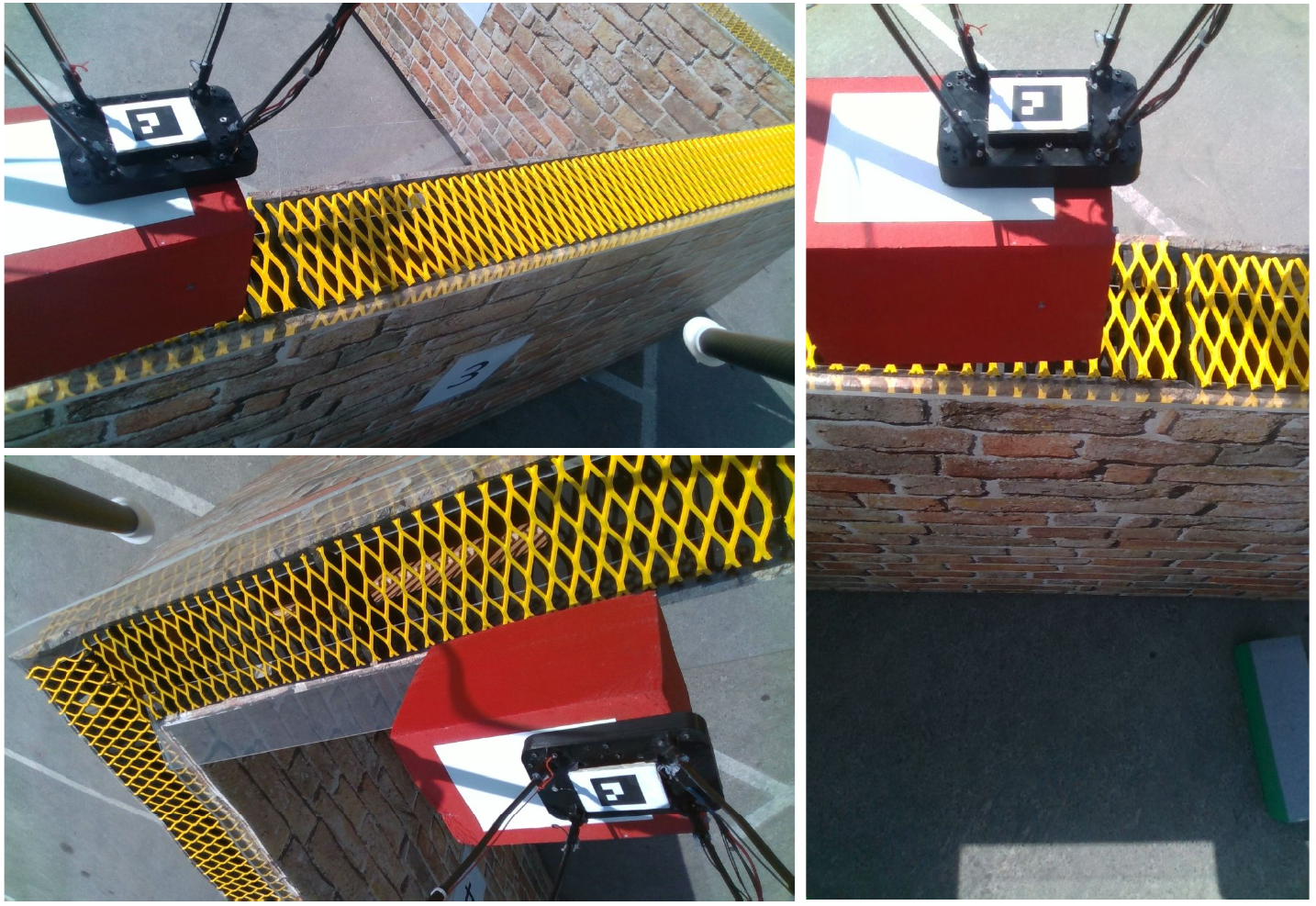}
    \caption{RGB images from all three cameras.}
  \end{subfigure}
  \begin{subfigure}[t]{0.4\linewidth}
    \centering
    \includegraphics[angle=90,width=1.0\linewidth,clip,frame,trim=220 80 270 80]{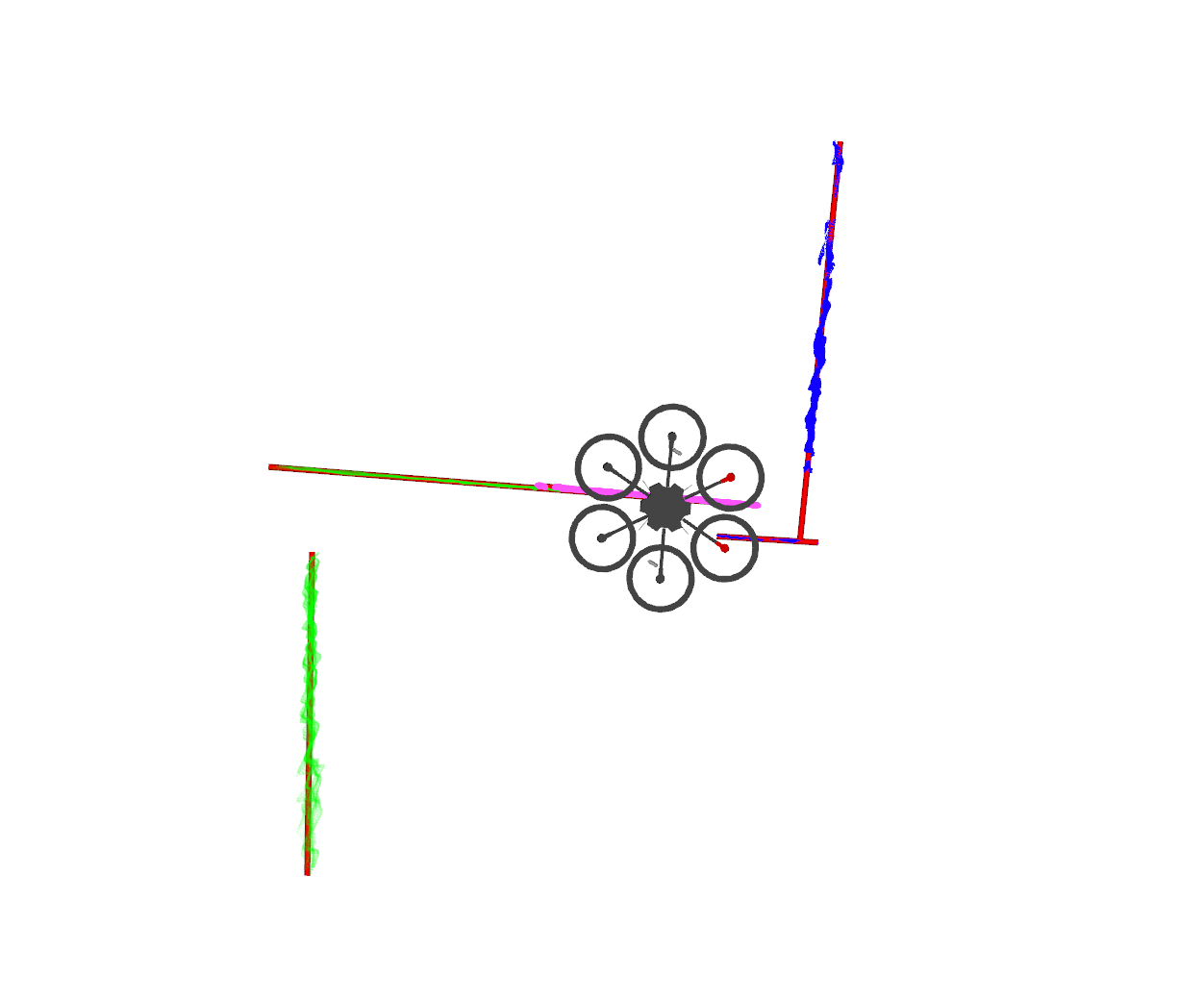}
    \caption{Top-down view of detected wall points per camera (green, blue, purple) and detected wall segments (red lines).}
  \end{subfigure}
  \caption{Wall localization.}
  \label{fig:wall_perception}
\end{figure}

Since not much was known about the target wall for brick placement prior to the competition apart from its shape, we decided to use only the wall geometry for detection and localization. Since there is no larger structure in the vicinity of the wall, we felt this would be sufficient. Of course, the bright yellow mesh grids on top of the wall (see \cref{fig:wall_perception}) would have been a good feature as well, but these were not known to the participants beforehand.

Our wall perception module estimates the height above ground from the depth image of the downward-facing camera. 3D points measured from each camera are then filtered so that only points above 1.0\,m and below 1.7\,m remain.
We then project the data to 2D, where we extract lines using RANSAC. Each line, if fit correctly, corresponds to a side view of one wall segment (see \reffig{fig:wall_perception}).
The system is initialized with a user-specified initial wall pose relative to the starting pose, which serves as the search pose.
The wall pose is updated any time two parallel line segments of valid length with 4\,m distance are found.
Under the assumption that the wall did not rotate 180° relative to the initialization, this is unambiguous.
Carrying a brick, Lofty targets a specific pose on one of the wall segments to place the brick as indicated by the wall plan
given by the challenge organizers.
During close approach, the detected segment closest to the expected segment pose is identified, and the goal position is projected onto this segment.

\subsection{High-level Control}

The high-level control module is implemented in an FSM framework. It is supplied with the target wall pattern as defined by the organizers of the competition. The basic cycle of events is designed as follows:
\begin{compactenum}
 \item Fly to the last known pile pose and fly a search pattern until the next brick requested by the pattern is found.
 \item Grasp the brick and lift it.
 \item Fly to the target position (relative to the last known wall pose) and look for a wall segment.
 \item Approach the projected position on the wall segment and place the brick.
\end{compactenum}

Similar to our MBZIRC 2017 approach~\citep{beul2019team}, we utilize a ``cone of descent'' during grasping and placement, in which Lofty is allowed to descend towards the target pose. If it drifts outside of the cone, it has to stay at that height until the disturbance is countered. The cone angle is 10° with a hysteresis of 3° to prevent oscillations. The cone was shifted such that at the target height, it had a radius of 9\,cm, which we determined as the maximum allowable deviation that would still allow successful magnetic grasping.

\subsection{Evaluation}
\label{sec:Experiments}
During the MBZIRC 2020, Lofty performed in six arena runs: three rehearsal runs, two Challenge~2 runs, and the final Grand Challenge run. A video showcasing the evaluation can be found on our website\footnote{\url{https://www.ais.uni-bonn.de/videos/fr_2021_mbzirc}}.

\begin{figure}[t]
  \centering
  \begin{subfigure}[T]{0.30\linewidth}
%     \begin{tikzpicture}[font=\footnotesize]
%       \begin{axis}[ybar, bar width=0.2cm, legend pos=north west, xlabel=Day, width=1.1\linewidth, height=3.7cm, xtick distance=1,tick pos=left]
%         \addplot+ table [x=day,y=success] {\dailysuccess};
%         \addlegendentry{Success}
%         \addplot+ table [x=day,y=total] {\dailysuccess};
%         \addlegendentry{Attempts}
%       \end{axis}
%     \end{tikzpicture}
    \vspace{3mm}
    \includegraphics[width=1.0\linewidth]{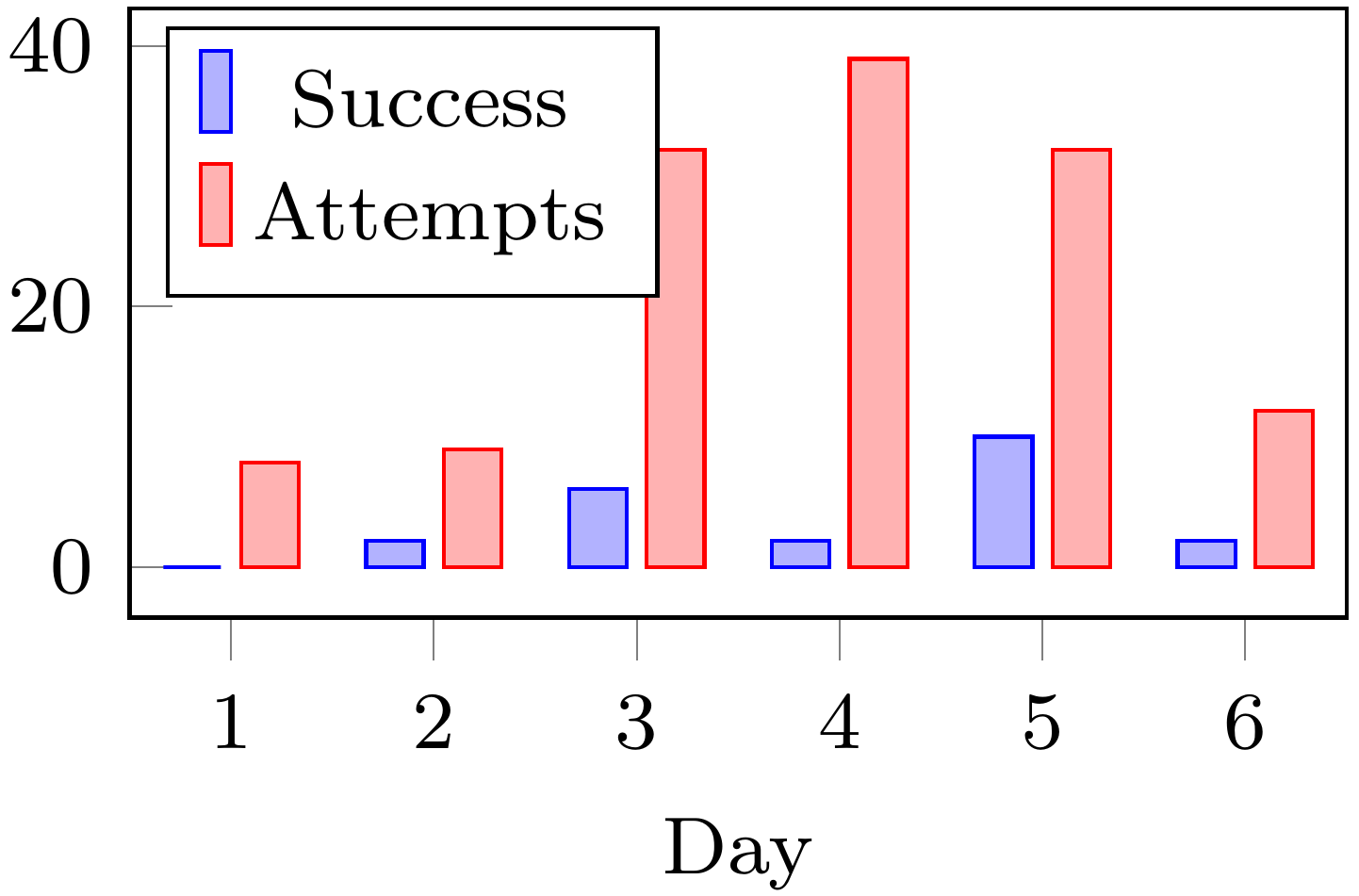}
    \caption{Pick success rate over the duration of the competition.}
    \label{fig:picking_1}
  \end{subfigure}
  \hfill
  \begin{subfigure}[T]{0.30\linewidth}
%     \begin{tikzpicture}[font=\footnotesize]
%       \begin{axis}[ybar, width=1.1\linewidth, height=3.7cm, xlabel={Failed pick duration [s]},tick pos=left]
%         \addplot +[hist={bins=12,data min=0.0,data max=30.0}] table [y index=0] {data/ch2/lofty_pick_fail_times.txt};
%         \draw[green!50!black,very thick,dashed] (axis cs:26, \pgfkeysvalueof{/pgfplots/ymin})--(axis cs:26, \pgfkeysvalueof{/pgfplots/ymax});
%       \end{axis}
%     \end{tikzpicture}
    \vspace{7mm}
    \includegraphics[width=1.0\linewidth]{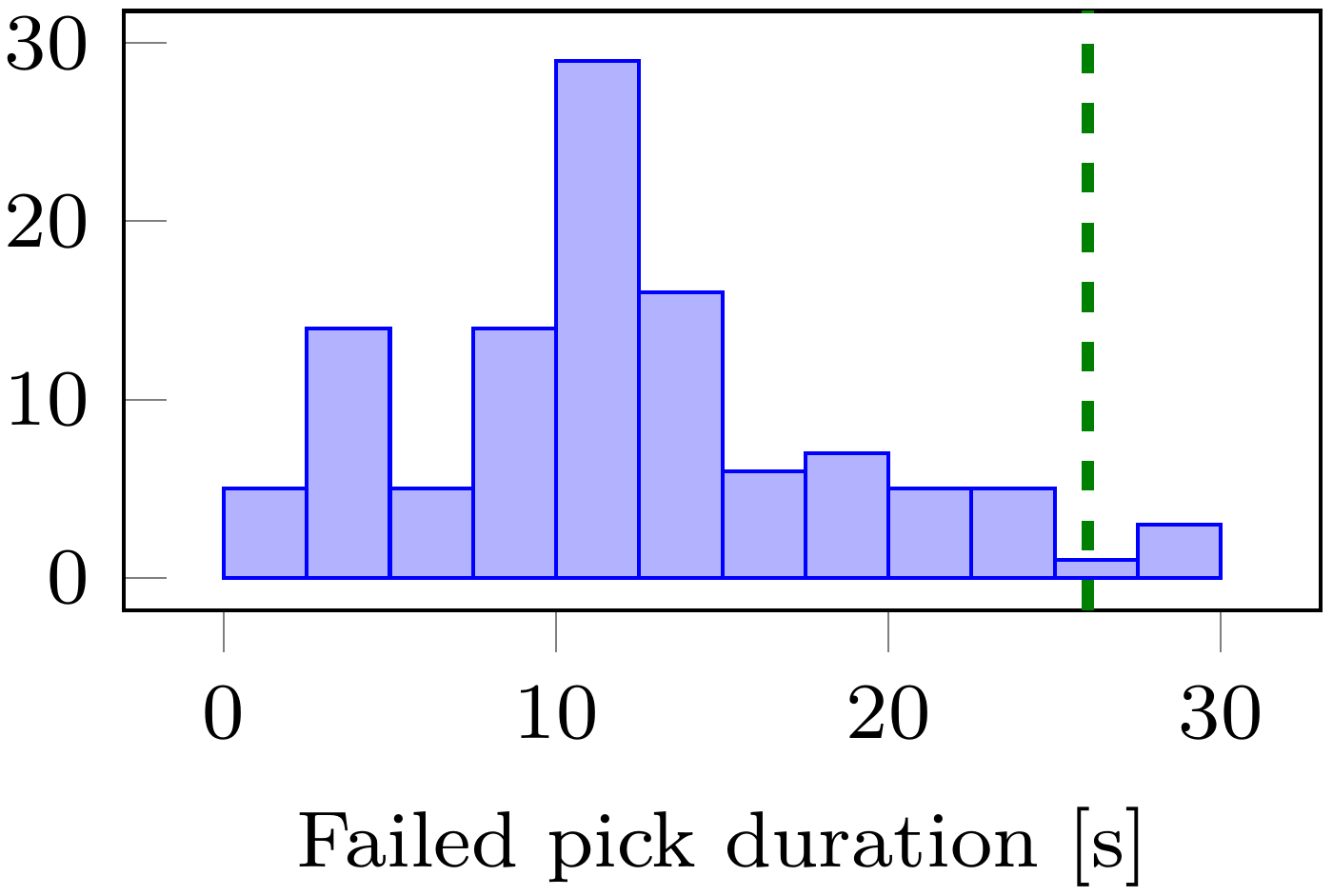}
    \caption{Histogram of failed pick durations. The average successful pick is shown in green.}
    \label{fig:picking_2}
  \end{subfigure}
  \hfill
  \begin{subfigure}[T]{0.36\linewidth}
%    \begin{tikzpicture}[font=\footnotesize]
%     \begin{axis}[ybar stacked, width=0.7\linewidth, height=3.7cm,bar width=0.2cm, legend pos=outer north east, xlabel=Day, xtick distance=1, tick pos=left, legend style={font=\scriptsize}, reverse legend, xmin=3.1, xmax=6.9, xtick align=outside]
%       \addplot+ table [x=day,y=lostapproach] {\placeresult}; \addlegendentry{Lost brick};
%       \addplot+ table [x=day,y=tracking] {\placeresult};     \addlegendentry{Lost wall tracking};
%       \addplot+ table [x=day,y=beside] {\placeresult};       \addlegendentry{Besides the wall};
%       \addplot+ table [x=day,y=blown] {\placeresult};        \addlegendentry{Blown off};
%       \addplot+[fill=green] table [x=day,y=success] {\placeresult};      \addlegendentry{Success};
%     \end{axis}
%    \end{tikzpicture}
   \includegraphics[width=1.0\linewidth]{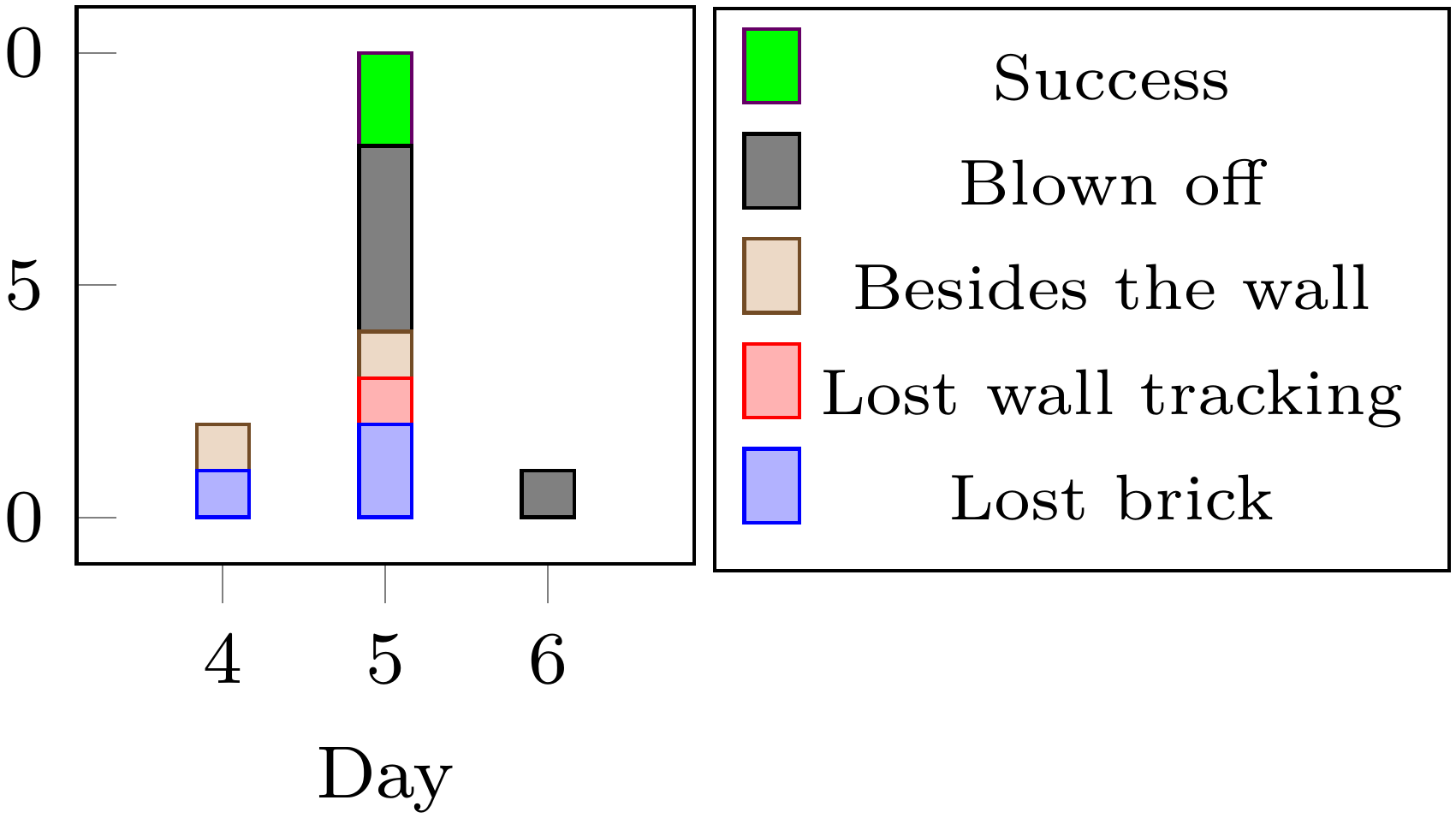}
   \caption{Placement results.}
   \label{fig:placing}
  \end{subfigure}
  \caption{Challenge 2: Pick \& place robustness.}
  \label{fig:pick_success_rate}
\end{figure}

We used the rehearsal runs to get used to the conditions in Abu Dhabi and continuously improved our pick success rate (see \reffig{fig:pick_success_rate}). During our first Challenge~2 run, we only picked one red and one green brick due to difficulties with our magnetic gripper. Both bricks were dropped close to but not \emph{on} the wall due to wall tracking problems. The wall tracking module had not been tested until this point due to short development time and lack of suitable testing opportunities at the competition.

After improving our gripper overnight, we managed to pick four red bricks and one green brick and placed two red bricks successfully during our second Challenge~2 run. The other bricks were sadly dropped right next to the wall due to another wall tracking problem. This run was scored as 1.33 points, which secured a second place in Challenge~2, next only to the Prague-UPenn-NYU team.

In the Grand Challenge, Lofty managed to pick a red brick but placed it a bit too high, and it fell off the wall. After a long pause to allow our Challenge~1 UAV to operate, it started again and picked up a green brick. Sadly, it falsely detected a W-shaped wall behind the Arena netting. Due to a rushed setup sequence, both wall search pose and the geofencing was not set up correctly and did not prevent the false detection nor Lofty from flying into the net. After a short, unsuccessful rescue attempt during a reset, we had to leave it there for the rest of the Grand Challenge.

Overall, Lofty executed 132 pick attempts in Abu Dhabi, of which 22 were successful, which gives a success rate of 16.7\,\%. Since a failed attempt took 12\,s on average, this limited the number of attempts we had for placing bricks on the wall. The number of pick attempts increased throughout the competition (see \reffig{fig:pick_success_rate}) as the rest of the system became more robust. There are two peaks in the duration histogram for failed picks: One at roughly three seconds which corresponds to tracking failures during the initial approach, and a larger one around 10\,s, which corresponds to misaligned picks or magnet
failures.

While we cannot provide a detailed accuracy analysis of the brick pose estimator due to missing ground truth, we can draw some
conclusions regarding the perception module. In 57\,\% of the pick attempts, the gripper made contact with a brick. This establishes
a lower bound on the detection performance. Out of these, 29\,\% resulted in a successful pick, which indicates sufficient precision
for magnetic gripping. Of course, other failure causes such as insufficient magnet power or wind gusts will also have reduced this
fraction.

We also show placement results in \reffig{fig:pick_success_rate}. In our best run, our second Challenge 2 run, Lofty attempted
ten placements, out of which 2 succeeded. The most prevalent failure reason was slightly off-center placement, which resulted in the placed brick being blown off the wall by Lofty's rotor wash.

% \pgfplotstableread{
% day success total percentage
% 1   0       8     0.0
% 2   2       9     22.2
% 3   6       32    18.8
% 4   2       39    5.13
% 5   10      32    31.3
% 6   2       12    16.7
% }\dailysuccess
% 
% \pgfplotstableread{
% day lostapproach tracking beside blown success
% 4   1            0        1      0     0
% 5   2            1        1      4     2
% 6   0            0        0      1     0
% }\placeresult

\section{Fire-Fighting}
\label{sec:ch3}

Challenge~3 of the MBZIRC 2020 targeted an urban fire-fighting scenario for autonomous robots to foster and advance the state of the art in perception, navigation, and mobile manipulation. Participants had to detect, approach, and extinguish multiple simulated fires around and inside a building with up to 3 UAVs and one UGV. Each fire provided a 15\,cm circular opening with a heated plate recessed 10\,cm on the inside. Holes on the outside facade were surrounded by a propane fire ring, while indoor fires had a moving silk flame behind the heating element. For these tasks, we developed our UAV ``Splasher'', described in the following.

\subsection{Hardware Design}
\label{sec:ch3:hw}
\begin{figure}
  \centering
  %\hfill\adjustbox{valign=M}{\input{uav.pgf}}\hfill
  %\adjustbox{valign=M}{\input{system.pgf}}\hfill\strut
  \includegraphics[width=1.0\linewidth]{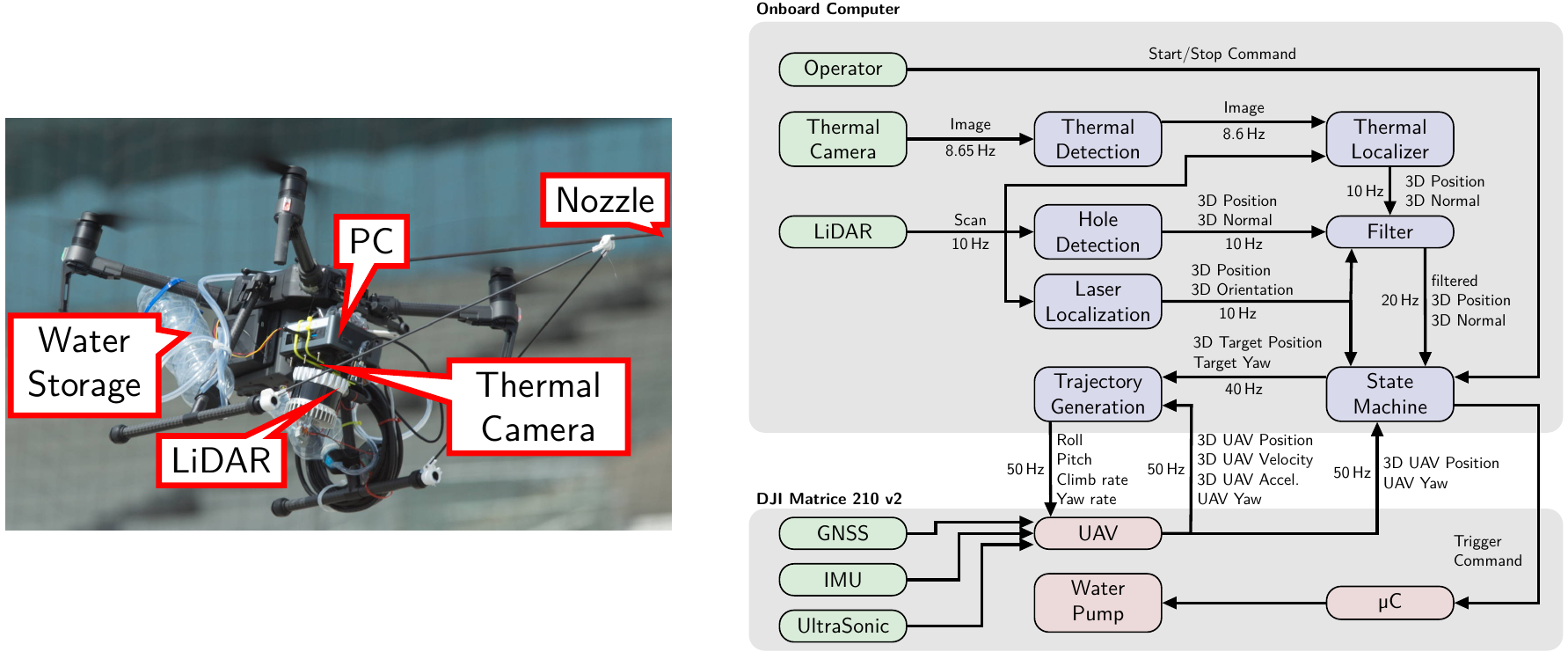}~
  \caption{Hard- and software design of our fire-fighting UAV ``Splasher''.}
  \label{fig:splasher}
\end{figure}

\reffig{fig:splasher} shows the setup of Splasher. It is based on a DJI Matrice~210~v2 and is equipped with an Intel\textsuperscript{\textregistered} Bean Canyon NUC8i7BEH with a Core\textsuperscript{\texttrademark} i7-8559U processor and 32\,GB RAM. We combine an Ouster OS1-64 LiDAR and a FLIR Lepton 3.5 thermal camera with 160\,$\times$\,120\,px resolution for perception and localization of the fires. The DJI Matrice~210~v2 provides GNSS-based ego-motion estimates. We deactivated the obstacle avoidance of the DJI Matrice~210~v2 in the forward direction to get close enough to fires for extinguishing.

The water supply is stored in two downward-facing 1.5\,l PET-bottles attached between the rear frame arms and the landing gear. The screw caps are connected via a flexible hose to a windscreen washer pump. We mounted a carbon tube on top of Splasher as an extended nozzle. It is supported by two additional carbon struts from the landing gear. The high location compensates for the height difference in the water jet parabola, allows to perceive the fire with the sensors below, and prevents the water from flowing out on its own during flight maneuvers. We chose a 10° downturn for the nozzle, LiDAR, and thermal camera to fly above the fire while maintaining observability.

\subsection{Laser Localization}
\label{sec:LaserLoc}
GNSS-based localization is subject to local position drift and is unreliable close to structures. The usage of RTK-/D-GPS was allowed but penalized. Hence, we localize Splasher w.r.t. the building using LiDAR. In a test run, we collected data to generate a Multi-Resolution Surfel Map (MRSMap) of the arena with the approach from \citet{DavidCTSLAM}. During localization, we register the current LiDAR point cloud against the map. The DJI Matrice~210~v2 provides an estimate of its ego-motion from onboard sensors, which we use to update the initial pose. After registration, we update an offset transformation that compensates for the local position drift of the GPS. The translation between consecutive offset updates is bounded to 30\,cm to prevent potential jumps from incorrect registration.

\subsection{Hole \& Thermal Detection}
\label{sec:uav_thermal_detection}
\begin{figure}
  \centering
  \includegraphics[height=4cm]{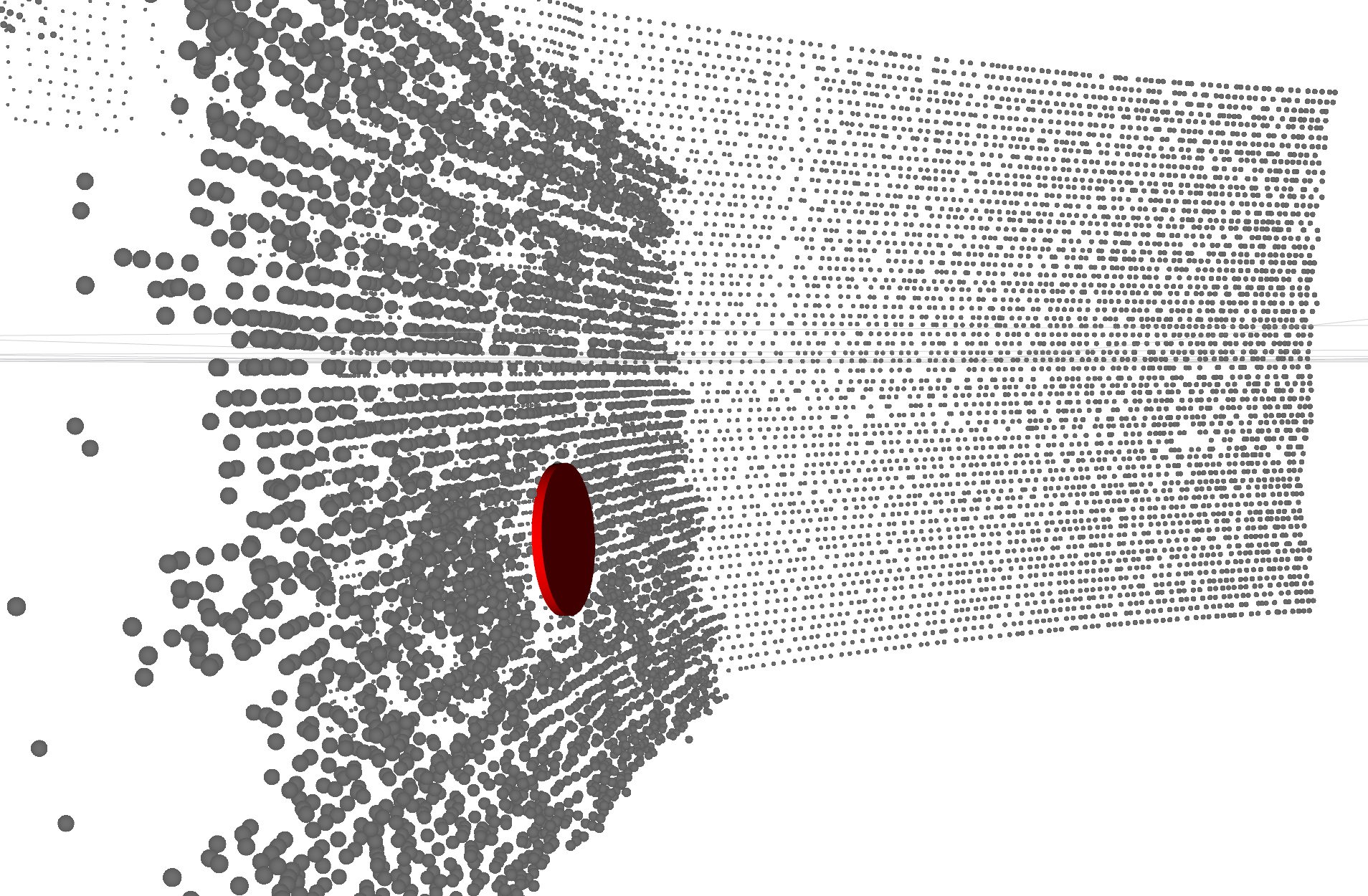}
  \includegraphics[height=4cm]{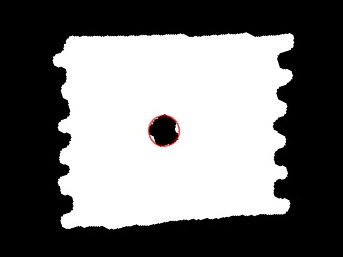}
  \caption{Detected hole (red disk) inside a point cloud and points projected into the virtual camera after morphological transformation and the detected circle.}
  \label{fig:hole_detection}
\end{figure}

The water jet exiting the nozzle is very narrow, so precise aiming is required in order to successfully extinguish a fire with a limited amount of water. We apply a hole detection algorithm on the LiDAR point clouds to use for relative navigation (see \reffig{fig:hole_detection}). First, planes are extracted using RANSAC. For each plane, we project the contained points into a virtual camera perpendicular to it. Morphological transformations are applied to the resulting image in order to close gaps between projected points. Shape detection is used on the image to find potential holes. After re-projecting the circles into 3D, we discard overly large or small holes, as the scenario specifies a constant 150\,mm diameter for all target holes.

Our thermal detector applies lower (23000) and upper bounds (65535) to threshold\footnote{The intensity thresholds are unitless.} the intensity. We find contours in the thresholded image. For each contour, we compute the area, bounding box, center of intensity, as well as min, max, and mean intensity for further processing. We only retain contours of a specific size.
After that, we project the LiDAR point cloud into the thermal image and filter points outside of the detected bounding box. Finally, we calculate the centroid and normal direction of the remaining points.

The thermal and the hole detector both output lists of detections $d_i=(p_i,n_i)$, each consisting of a position $p_i$ and a normal $n_i$. A filtering module processes these detections to reject outliers and combines both detection types into an estimate of the position and orientation of the currently tracked fire. To do this, we keep track of a history $\mathcal H = ( (p_1, n_1), \dots, (p_{10}, n_{10}) )$ of ten recent valid detections and estimate the target position as running averages over the detection history.

Mind that the detection history may contain both thermal and hole detections. Thermal detections are necessary to determine which of the possible targets is currently on fire. However, we found that hole detections give a more precise estimation of the target position and especially of its orientation. Hence, we use thermal detections to initialize the target estimate and subsequently update it using hole detections if available. Since the detection history contains both types of detections, they are \emph{implicitly} fused by the filter. If one of the detection systems fails and, thus, outputs no detections, the history is only filled with the other type of detection. During regular operation, however, the filtered fire position comprises both detections.

Although we keep track of the most recent thermal detection, we only add it to the detection history $\mathcal H$ if there has not been a valid hole detection within the last second.
Thus, we ensure that we can still estimate target positions if the hole detector fails, but otherwise only use the more precise information from hole detections.
In the case of multiple holes close to the target position, the estimate might drift away from the target if we only use hole detections.
Furthermore, we have to recover if initial heat detections are wrong or if the fire was extinguished in the meantime.
To address these issues, we only add hole detections to the detection history $\mathcal H$ if there has been a heat detection within the last second and the detected hole position lies within a radius of 1.0\,m around the latest heat detection.

\begin{figure}
  \centering
  \includegraphics[height=4cm]{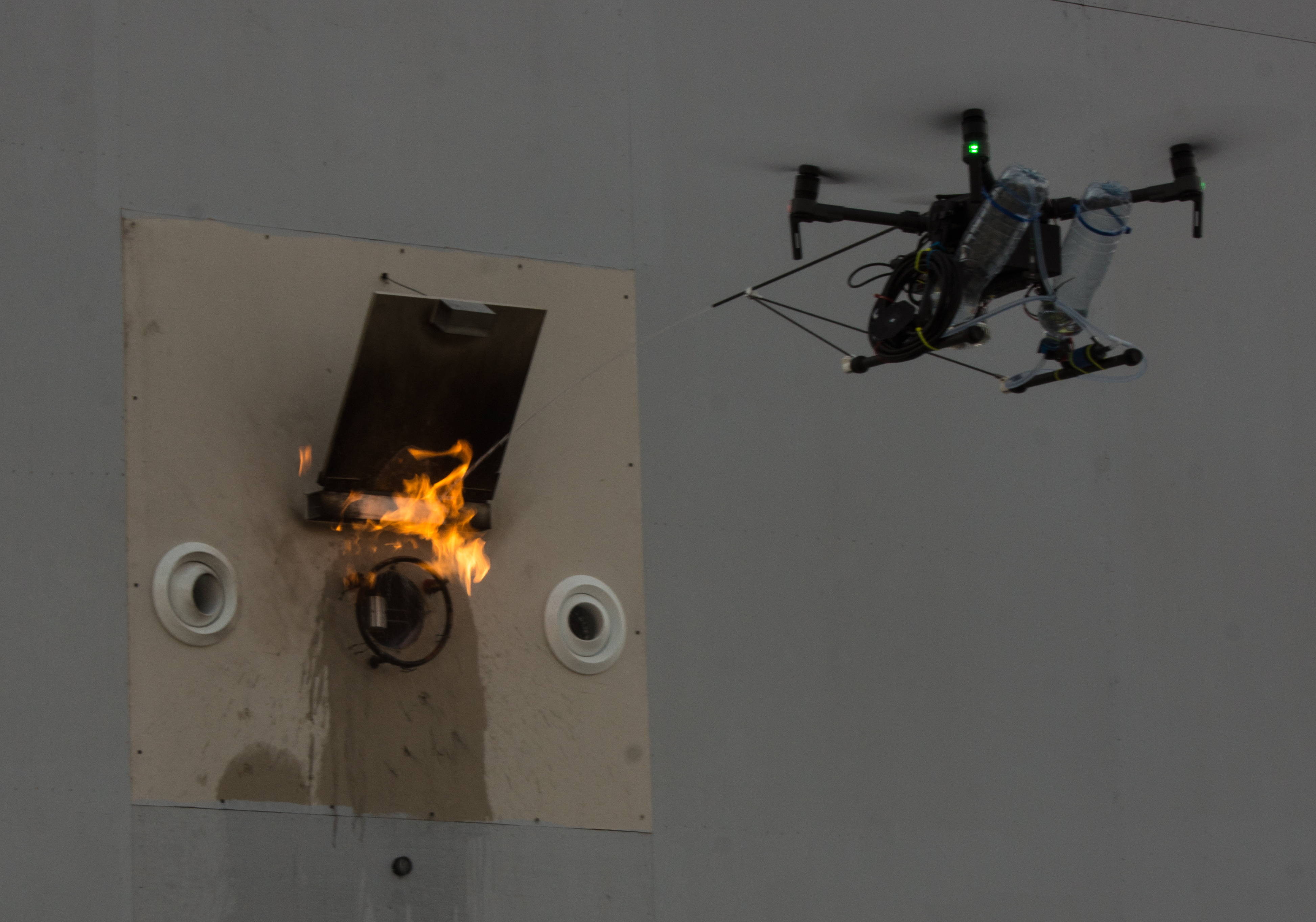}
  \caption{Splasher extinguishing a fire during the MBZIRC 2020.}
  \label{fig:copter_with_fire}
\end{figure}

One might think that the open fire (\reffig{fig:copter_with_fire}) would negatively affect LiDAR performance due to smoke and heat. To the contrary, a thorough investigation of the LiDAR scans after the competition trials yielded no significant impact
on LiDAR performance.

\subsection{Fire-fighting control}
\label{sec:Fire_fighting_control}
\begin{figure}
    \centering
    \includegraphics[width=0.95\linewidth]{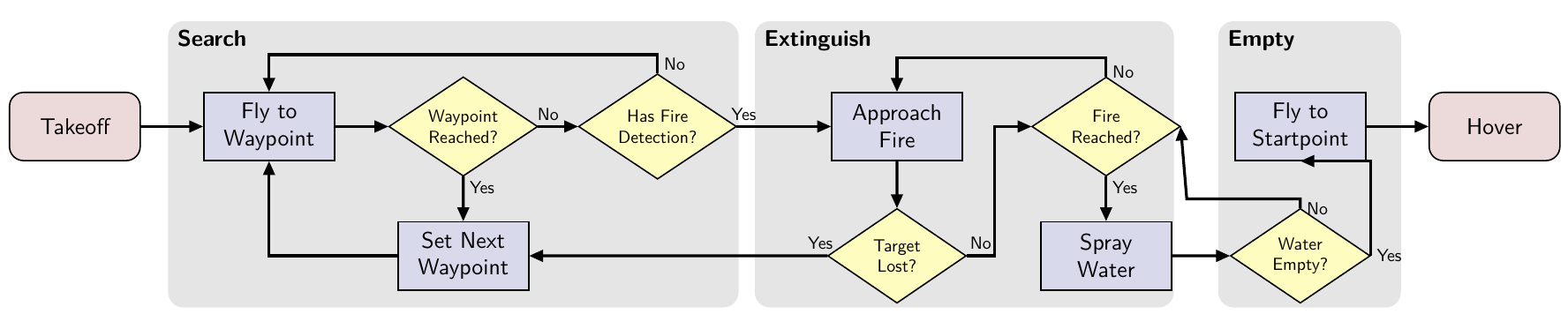}~
    \caption{Flowchart of Splasher's state machine.}
    \label{fig:uav_state_machine}
\end{figure}

The high-level control of Splasher is performed by a Finite State Machine (FSM). The FSM uses inputs from components described above to produce navigation waypoints to locate and approach fires, as well as to control the water spraying during extinguishing. It also ensures that Splasher stays within the arena limits and altitude corridor. The diagram of the FSM is shown in \reffig{fig:uav_state_machine}.

\textbf{\textit{Search} state:} Splasher flies around the building in order to detect and localize a fire on all sidewalls. The route around the building is manually defined as a linked list of waypoints specified in the building frame, obtained from laser localization.
While moving between waypoints, the detection and filter pipeline collects data. Some waypoints are marked as \textit{Observation} waypoints. Upon arrival at an \textit{Observation} waypoint, located in front of the known hole locations, Splasher hovers for a predefined duration of 2\,s, looking for fires. Splasher switches into the \textit{Extinguish} state if a detection was observed at least five times in a row and it is within the allowed height and angle boundaries.

\textbf{\textit{Extinguish} state:} The purpose of this state is to arrive at the extinguishing position without losing the detected fire on the way. We do so by steering the robot
relative to the detected fire (\ie visual servoing to the target position). To do so, we first transform the egocentric fire detection into an allocentric frame. Based on the allocentric fire detection, we derive an extinguishing waypoint that lies relative to the detection (horizontal offset: 2.1\,m, vertical offset: 0.35\,m), which Splasher targets.
Once Splasher reaches the goal pose, the pump is started. It is stopped as soon as Splasher deviates from the current goal pose more than a predefined threshold for either position or heading. Splasher keeps track of the fire and updates the goal pose accordingly. If Splasher did \emph{not} lose the target during extinguishing, the pump is stopped after all water has been sprayed and we enter the \textit{Refilling} state.

\textbf{\textit{Refilling} state:} We do not \emph{directly} measure water content but measure the time the pump is active. Once the water reserve is depleted, Splasher flies back to the starting position and hovers there awaiting manual landing and refueling of the water storage.

\subsection{Evaluation}
In the first scored challenge run, software issues and wrong predefined search poses prevented Splasher from detecting any fire. On the second challenge day, we experienced incorrect height estimates. We attribute this to the ultrasonic sensor measuring the building wall rather than the ground, thus estimating the height too low. In hindsight, we believe that a height sensor with a smaller opening angle (like employed in Challenge~1) could have solved this problem, but we were not aware of the problem's severity during the competition.

\begin{figure}[t]
  \centering
  \begin{subfigure}[t]{0.56\linewidth}
    \centering
%     \begin{tikzpicture}[boxstyle/.style={orange!80!black,fill=orange!80!black,fill opacity=0.5,text opacity=1,text=black,draw,ultra thick,align=center}]
%       \node[anchor=south east, inner sep = 0] (left_image) at (0,0) {\includegraphics[height=6cm,clip,trim=700 0 100 0]{uavday2bottoms.png}};
%     \end{tikzpicture}
    \includegraphics[height=6cm]{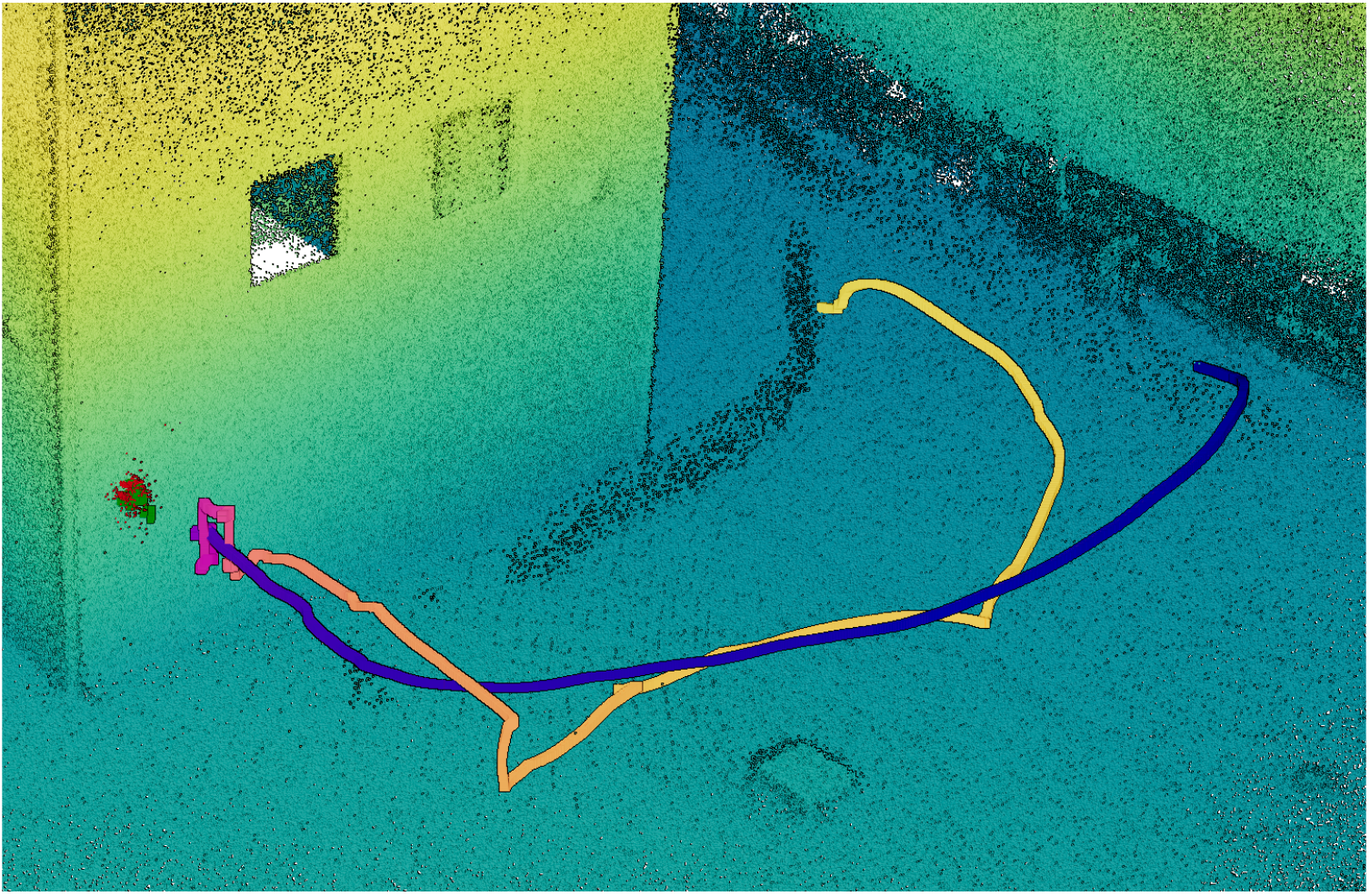}
    \caption{Fire detection and UAV localization during manual fire extinguishing of the windy ground-level facade fire during the second challenge day. The trajectory is color-coded by time, yellow to blue.}
  \end{subfigure}
  \hfill
  \begin{subfigure}[t]{0.43\linewidth}
    \centering
%     \begin{tikzpicture}[font=\sffamily,boxstyle/.style={red,fill=white,text=black,draw,ultra thick,align=center}]\clip (-7cm,0) rectangle (0, 6cm);
%       \node[anchor=south east, inner sep = 0] (left_image) at (0,0) {\includegraphics[height=6cm]{uavgchcbs.png}};
%       \begin{scope}[shift=(left_image.south west),x={(left_image.south east)},y={(left_image.north west)}]
%         \node[boxstyle,text width=1.9cm] at (0.43,0.11) {Laser\\Localization};
%         \node[boxstyle,text width=1.9cm] at (0.875,0.6) {GNSS-based\\Localization};
%         \node[red,draw,line width=2pt,from={ 0.35,0.2 to 0.5,0.4}]{};
%         \node[red,draw,line width=2pt,from={0.8,0.35 to 0.95,0.5}]{};
%     \end{scope}
%     \end{tikzpicture}
  \includegraphics[height=6cm]{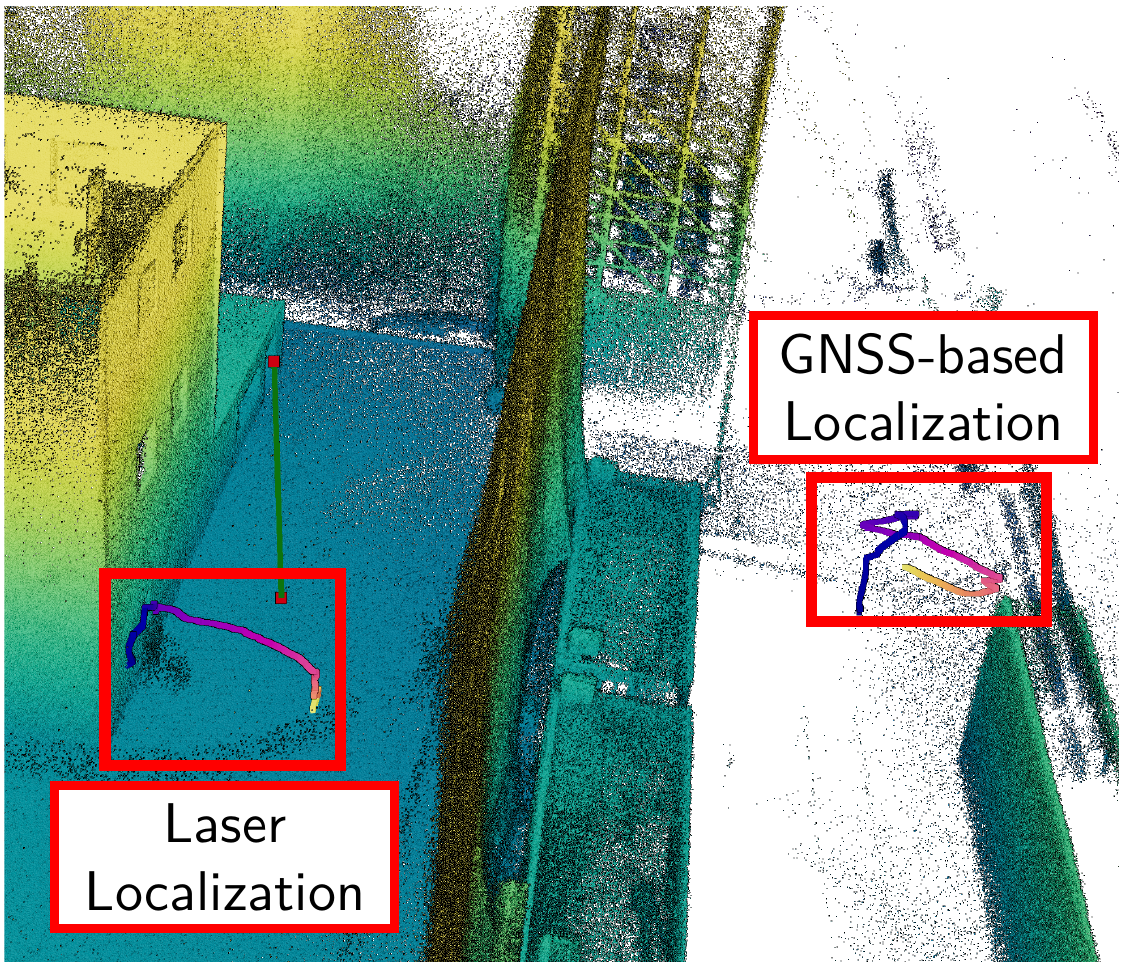}
  \caption{Incorrect GNSS-based UAV localization during the Grand Challenge. The laser localization shows the trajectory before crashing into the building.}
  \end{subfigure}
  \caption{UAV localization.}
  \label{fig:manual_fire}
\end{figure}

We noticed that during all trials Splasher flew too high to detect the fire. We then switched to manual mode and were able to fill the container of the windy ground-level facade fire with 322\,ml of water near the end of the second challenge run. The manually flown trajectory and localized detections of the fire are shown in~\reffig{fig:manual_fire}. After the Grand Challenge, we found out that our LiDAR-based localization was disabled the whole time.

As described in \refsec{sec:Runtime_Control_and_Supervision}, we employ tools that should catch errors like disabled components. Unfortunately, we did \emph{not} include the laser localization in our supervision system since we never expected such a core component to be subject to human error.

Shortly after the Grand Challenge started, a series of unfortunate circumstances led to Splasher crashing into the building, as shown in \reffig{fig:manual_fire}. The laser localization was disabled due to human error, and position estimates relied solely on the GNSS-based ego-motion estimates. To compensate for position drift, we added a static offset to the GPS poses, which was calculated as the difference between the predefined start position and the GPS pose when the challenge started. However, the GPS signal strongly drifted while computing the offset. Thus, a large offset was added to the GPS pose, resulting in an initial localization error of approximately 20\,m. Moreover, it continued to drift afterwards. The canyon-like starting position between the building and the operator booth, as well as interference with our UGV, might be responsible for the drift.

\begin{figure}[t]
  \centering
  \begin{subfigure}[t]{0.36\linewidth}
  \centering
%     \begin{tikzpicture}[boxstyle/.style={orange!80!black,fill=orange!80!black,fill opacity=0.5,text opacity=1,text=black,draw,ultra thick,align=center}]
%       \node[anchor=south east, inner sep = 0] (World_model) at (0,0) {\includegraphics[height=3.5cm]{uavpoppelsdorfs.png}};
%       \begin{scope}[shift=(World_model.south west),x={(World_model.south east)},y={(World_model.north west)}]
%         \node[fill=orange,circle,inner sep=0pt,minimum size = 12pt]at(0.035,0.21){1};
%         \node[fill=orange,circle,inner sep=0pt,minimum size = 12pt]at(0.42,0.9){2};
%         \node[fill=orange,circle,inner sep=0pt,minimum size = 12pt]at(0.43,0.65){3};
%         \node[fill=orange,circle,inner sep=0pt,minimum size = 12pt]at(0.495,0.495){4};
%         \node[fill=orange,circle,inner sep=0pt,minimum size = 12pt]at(0.65,0.55){5};
%         \node[orange,draw,line width=2pt,from={ 0.57,0.37 to 0.92,0.83}]{};
%     \end{scope}
%     \end{tikzpicture}
    \includegraphics[height=3.5cm]{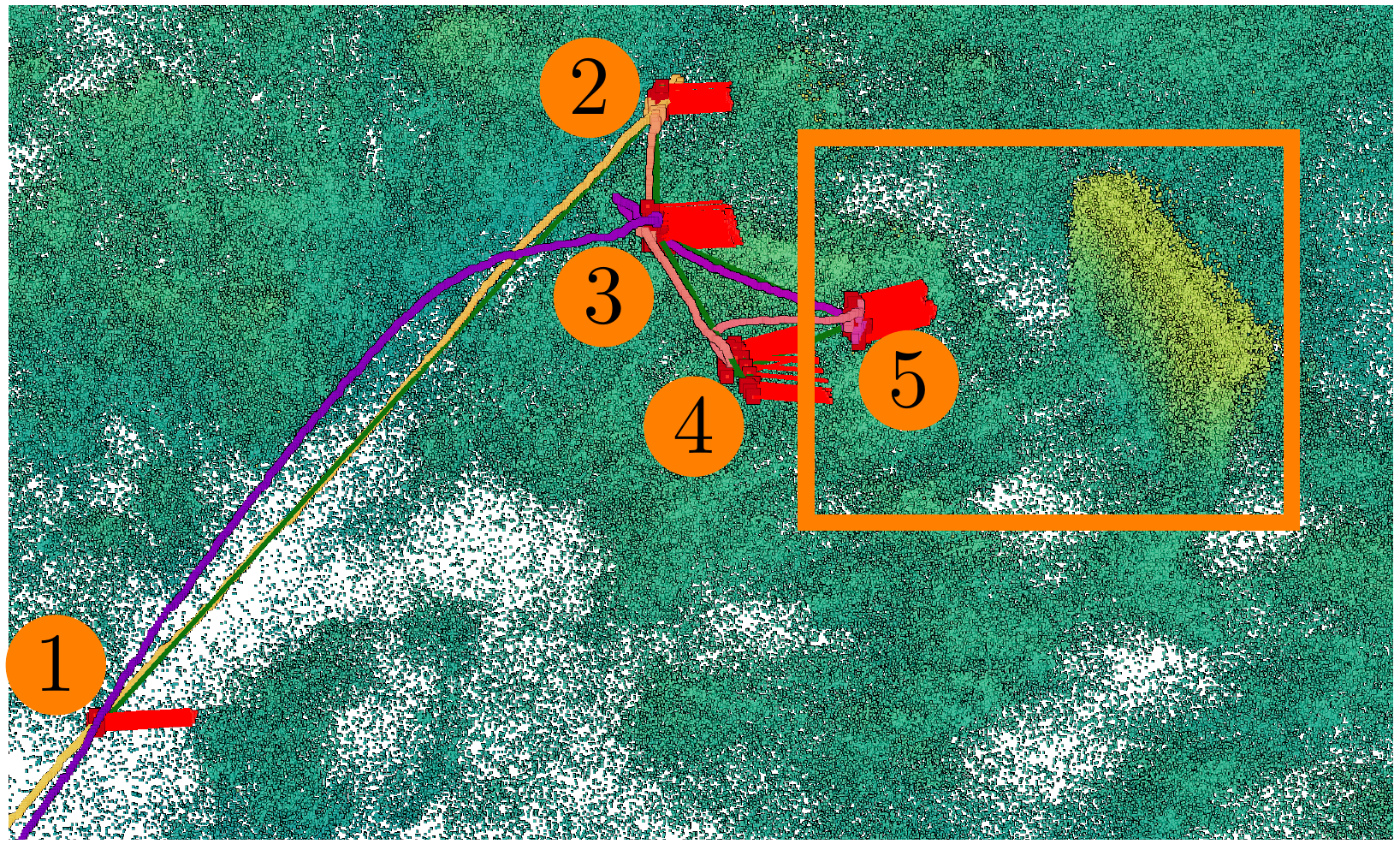}
    \caption{World model of the scene.}
    \label{fig:World_model}
  \end{subfigure}
  \begin{subfigure}[t]{0.29\linewidth}
  \centering
    \includegraphics[height=3.5cm]{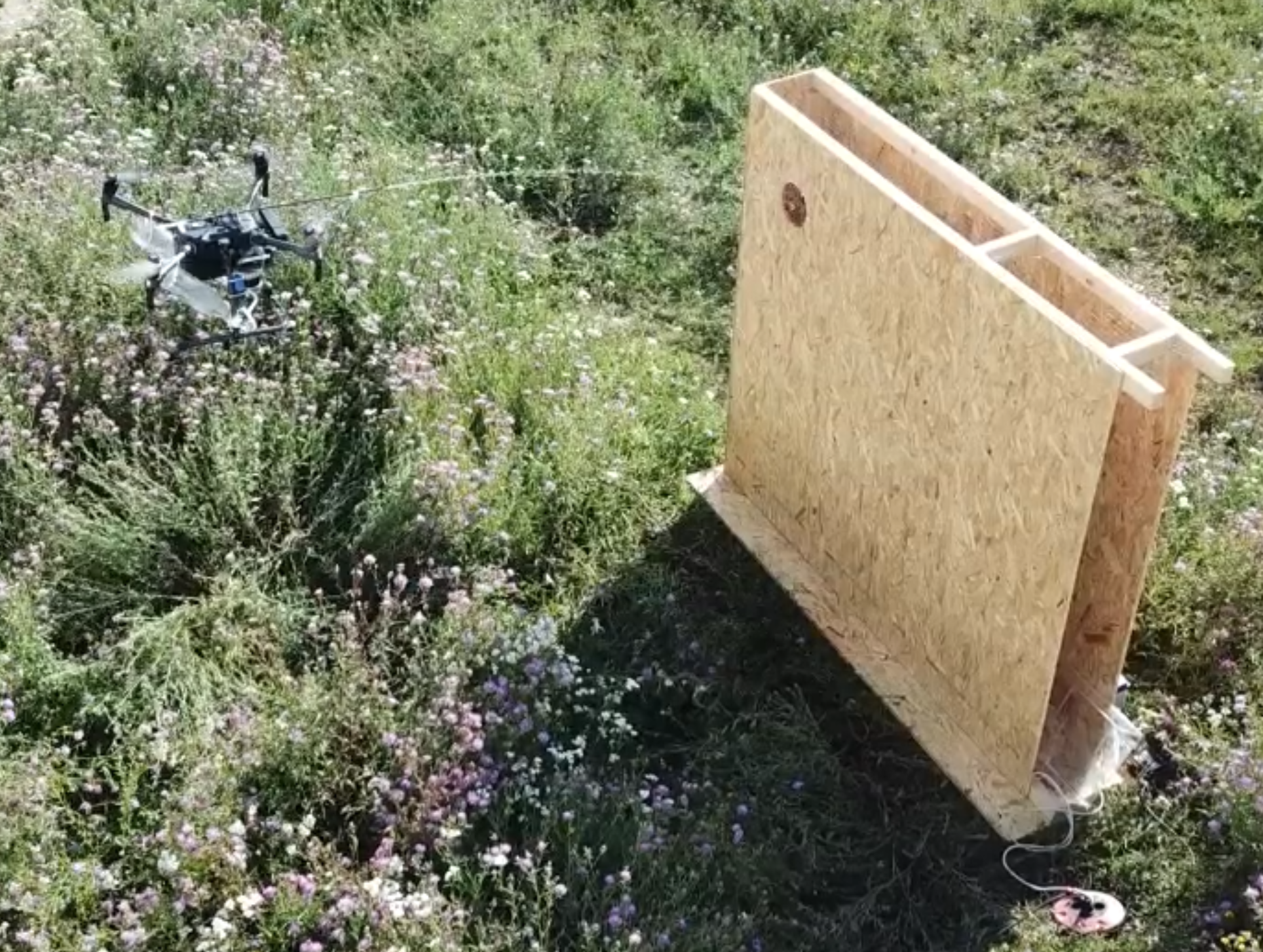}
    \caption{Bird's-eye view.}
    \label{fig:Birds_eye}
  \end{subfigure}
  \begin{subfigure}[t]{0.17\linewidth}
  \centering
    \includegraphics[height=3.5cm]{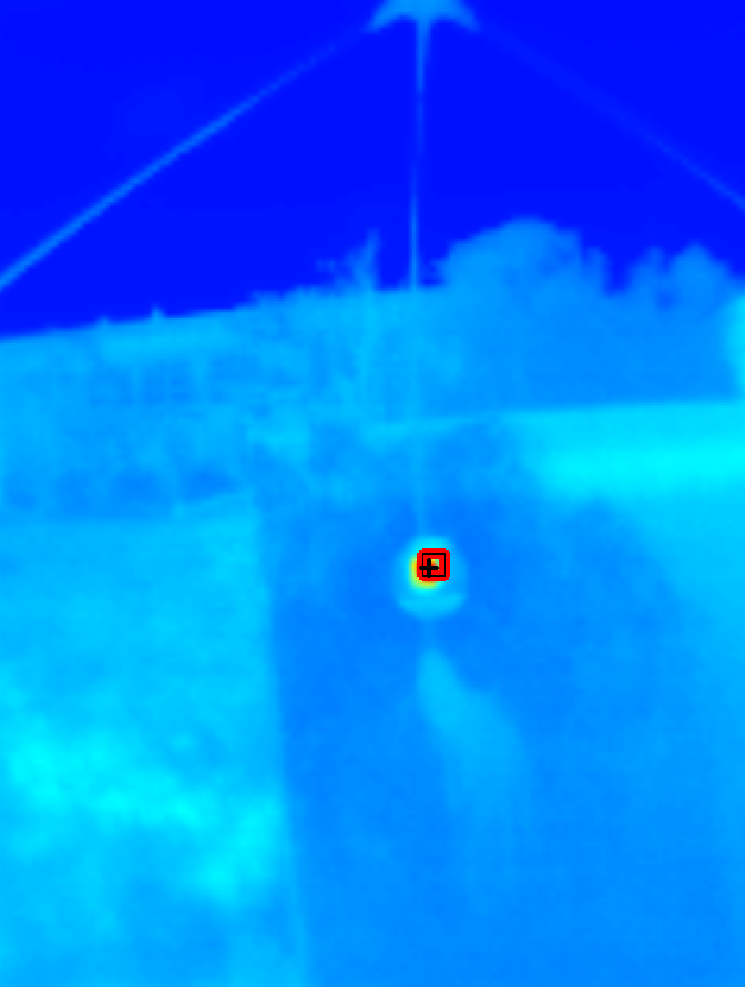}
    \caption{Onboard thermal camera.}
    \label{fig:Onboard_thermal}
  \end{subfigure}
  \begin{subfigure}[t]{0.14\linewidth}
    \centering
    \includegraphics[height=3.5cm]{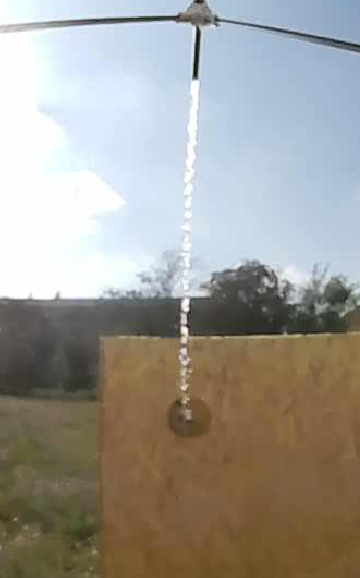}
    \caption{Onboard RGB-camera.}
    \label{fig:Onboard_RGB}
  \end{subfigure}
  \caption{Autonomous fire extinguishing on mockup target with heating element. Red line segments indicate the UAV's heading at various numbered waypoints (orange). The waypoints are targeted in ascending order on the colored trajectory. 1) The UAV starts 2,3) It searches for the heat source 4) It found the heat source 5) It approaches the wall and extinguishes the fire.}
  \label{fig:uav_poppelsdorf}
\end{figure}

After repair, we performed further lab tests to showcase Splasher's abilities on a mockup target (\reffig{fig:uav_poppelsdorf}). After lift-off, Splasher flies towards multiple predefined search waypoints (\reffig{fig:World_model}, Poses~1--3). The heat source is first detected at Waypoint~3. Now Splasher begins to navigate only relative to the detection and turns towards it before flying closer (Pose~4) and extinguishing the fire (\reffig{fig:Birds_eye}--\reffig{fig:Onboard_RGB}). A video showcasing the evaluation can be found on our website\footnote{\url{https://www.ais.uni-bonn.de/videos/fr_2021_mbzirc}}.
We repeated the experiment on the mockup seven times.
In all cases, the target was first detected on Pose~3. It took on average 1.01\,s from first detection to a stable filter estimate and already 4.96--6.60\,s later the target was reached and the pump started spraying water for on average 14.57\,s. The mean GPS drift from first detection until spraying stops was 0.68\,m while our UAV remained in a stable position relative to the heat source. This highlights the necessity of relative navigation for such a high-precision task.

\begin{figure}
\begin{subfigure}[t]{0.47\linewidth}
% \begin{tikzpicture}
% \begin{axis}[font=\footnotesize,width=\linewidth, height=5cm,
%   name={theaxis},
%   xlabel={error \text{[m]}},
%   ymin=0,
%   xmin=0,xmax=0.3,
%   xtick distance=0.06, xtick align=outside,
%   x tick label style={/pgf/number format/.cd, fixed,precision=3}
% ]
% \addplot [color=blue,fill,fill opacity = 0.2, hist={bins=11,data min=0,data max=0.3}] table {\lidarProjDistErrD};
% \end{axis}
% \end{tikzpicture}
\includegraphics[width=1.0\linewidth]{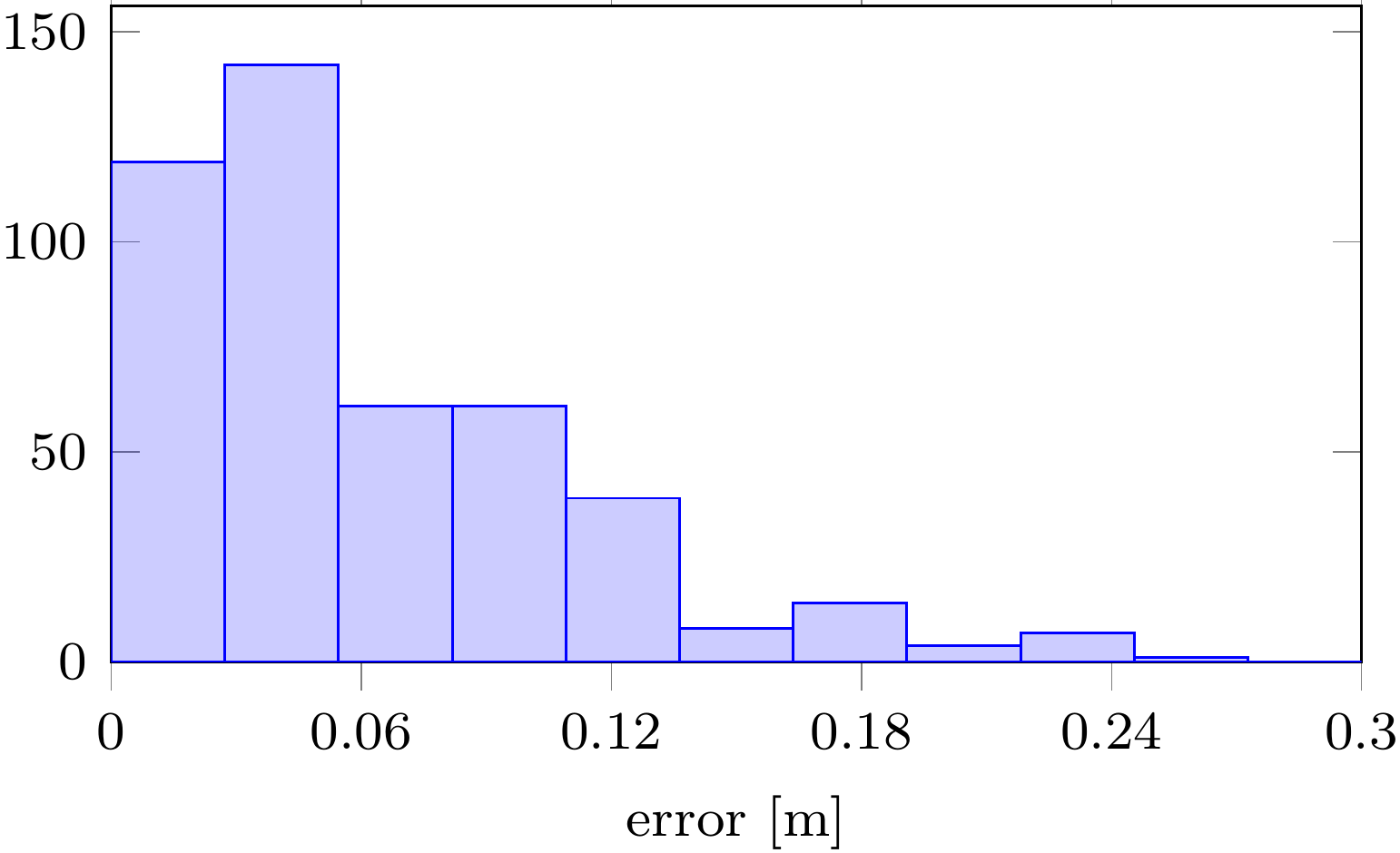}
\caption{Error histogram.}
\label{fig:hist_lidar_proj}
\end{subfigure}
\begin{subfigure}[t]{0.49\linewidth}
% \begin{tikzpicture}
% \begin{axis}[font=\footnotesize,width=\linewidth, height=5cm,
%   name={theaxis},
%   xlabel={distance \text{[m]}},
%   ylabel={error \text{[m]}},
%   ymin=0,
%   xmin=0.75,xmax=5.25,
%   ytick distance=0.1, xtick pos=left, xtick align=outside
% ]
% \addplot [only marks,color=blue,fill opacity = 0.2] table [x index={0},y index={1}]{\lidarProjDistErrD};
% \end{axis}
% \end{tikzpicture}
\includegraphics[width=1.0\linewidth]{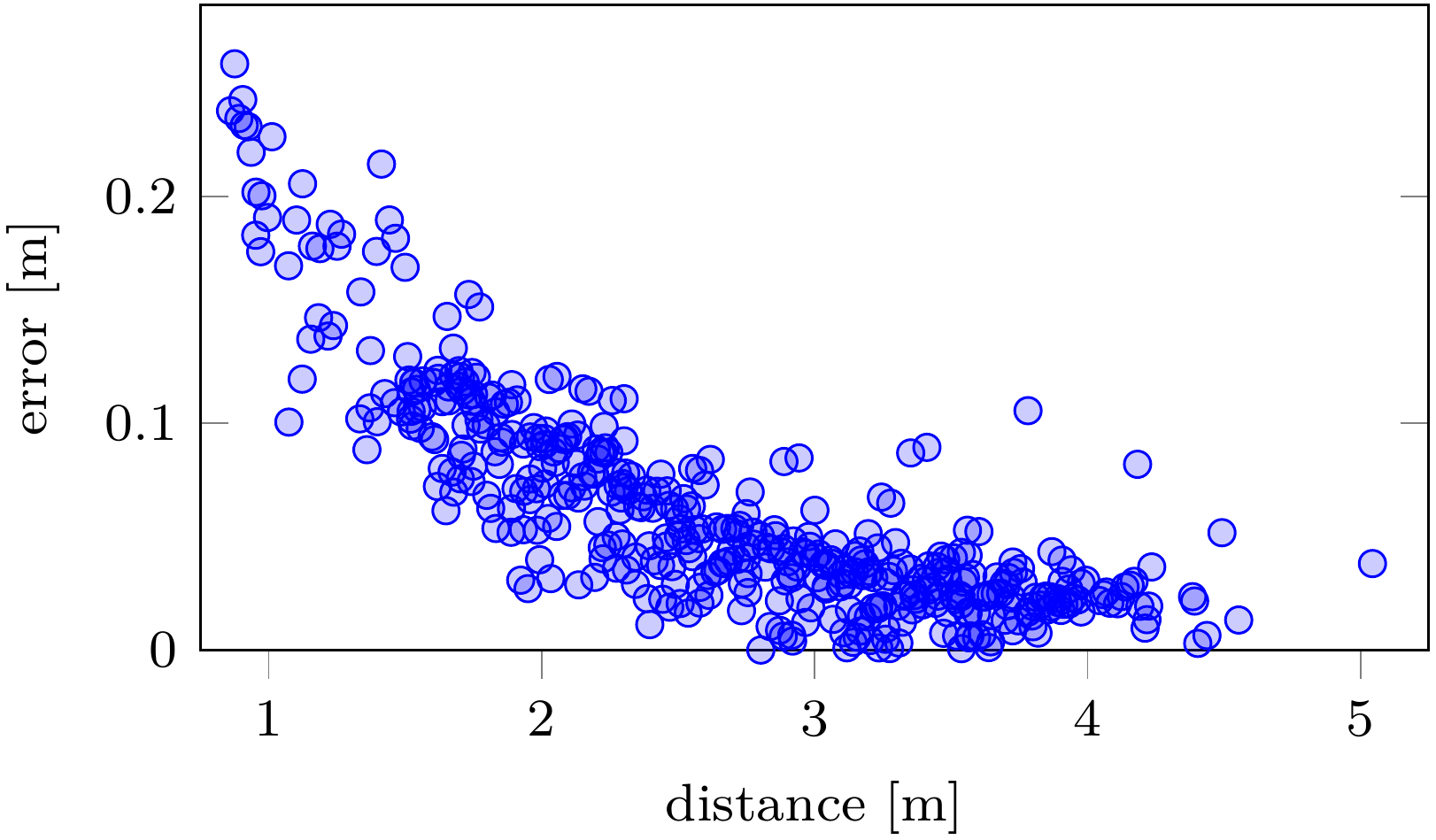}
\caption{Error over distance.}
\label{fig:dist_err_lidar_proj}
\end{subfigure}
\caption{Evaluation of distance for thermal detections with depth obtained by projecting LiDAR on the mockup target against ground truth from plane estimation. The error histogram shows sufficient accuracy to approach the target while the error increases at close range by partially measuring the backside through the hole.}
\label{fig:lidar_proj}
\end{figure}

In a separate test, we evaluate the accuracy for our estimated distance towards the heat source from projecting LiDAR into thermal detections at different ranges. We derive ground truth on the mockup from fitting a plane to its outside wall. The histogram in \reffig{fig:hist_lidar_proj} shows that the majority of all estimates has an error below 0.05\,m and less than 90\,\% have more then 0.12\,m error. Counterintuitively, close range measurements were least reliable, as visualized in \reffig{fig:dist_err_lidar_proj}. We found that more projected LiDAR measurements stem from the backside of the mockup and since we fuse measurements within the bounding box, the distance is overestimated and the error increases. In general, the accuracy is sufficient to approach the target until reliable hole detections are available. 

% \pgfplotstableread[col sep=comma]{./data/ch3/lidar_projection_distance_error.csv}\lidarProjDistErrD
% 
% \pgfplotstableread[col sep=comma]{./data/ch3/hole_2020_02_23_10_19_51_day1.csv}\holeDayOne
% \pgfplotstabletranspose{\holeDayOneT}{\holeDayOne}
% \pgfplotstableread[col sep=comma]{./data/ch3/hole_2020_02_24_19_25_50_day2.csv}\holeDayTwo
% \pgfplotstabletranspose{\holeDayTwoT}{\holeDayTwo}
% \pgfplotstableread[col sep=comma]{./data/ch3/hole_2020_07_17_18_XX_XX.csv}\holeHome
% \pgfplotstabletranspose{\holeHomeT}{\holeHome}
% \pgfplotstablevertcat{\holeDaysT}{\holeDayOneT}
% \pgfplotstablevertcat{\holeDaysT}{\holeDayTwoT}
% 
% \pgfplotstableread[col sep=comma]{./data/ch3/thermal_2020_02_23_10_19_51_day1.csv}\thermalDayOne
% \pgfplotstabletranspose{\thermalDayOneT}{\thermalDayOne}
% \pgfplotstableread[col sep=comma]{./data/ch3/thermal_2020_02_24_19_25_50_day2.csv}\thermalDayTwo
% \pgfplotstabletranspose{\thermalDayTwoT}{\thermalDayTwo}
% \pgfplotstableread[col sep=comma]{./data/ch3/thermal_2020_07_17_18_XX_XX.csv}\thermalHome
% \pgfplotstabletranspose{\thermalHomeT}{\thermalHome}
% 
% \pgfplotstablevertcat{\thermalDaysT}{\thermalDayOneT}
% \pgfplotstablevertcat{\thermalDaysT}{\thermalDayTwoT}

\begin{figure}
\begin{subfigure}[t]{0.49\linewidth}
% \begin{tikzpicture}
% \begin{axis}[width=\linewidth, height=5cm,
%   name={theaxis},
%   xlabel=distance,
%   ylabel=detections,
%   ymin=0,
%   xmin=0,xmax=18,
%   tick pos=left, xtick align=outside
% ]
% \addplot [color=green,fill,hist={bins=80,data min=0,data max=20}] table {\holeDaysT};
% \addplot [color=blue,fill,fill opacity = 0.2, hist={bins=80,data min=0,data max=20}] table {\thermalDaysT};
% \addlegendentry{Hole}
% \addlegendentry{Thermal}
% \end{axis}
% \end{tikzpicture}
\includegraphics[width=1.0\linewidth]{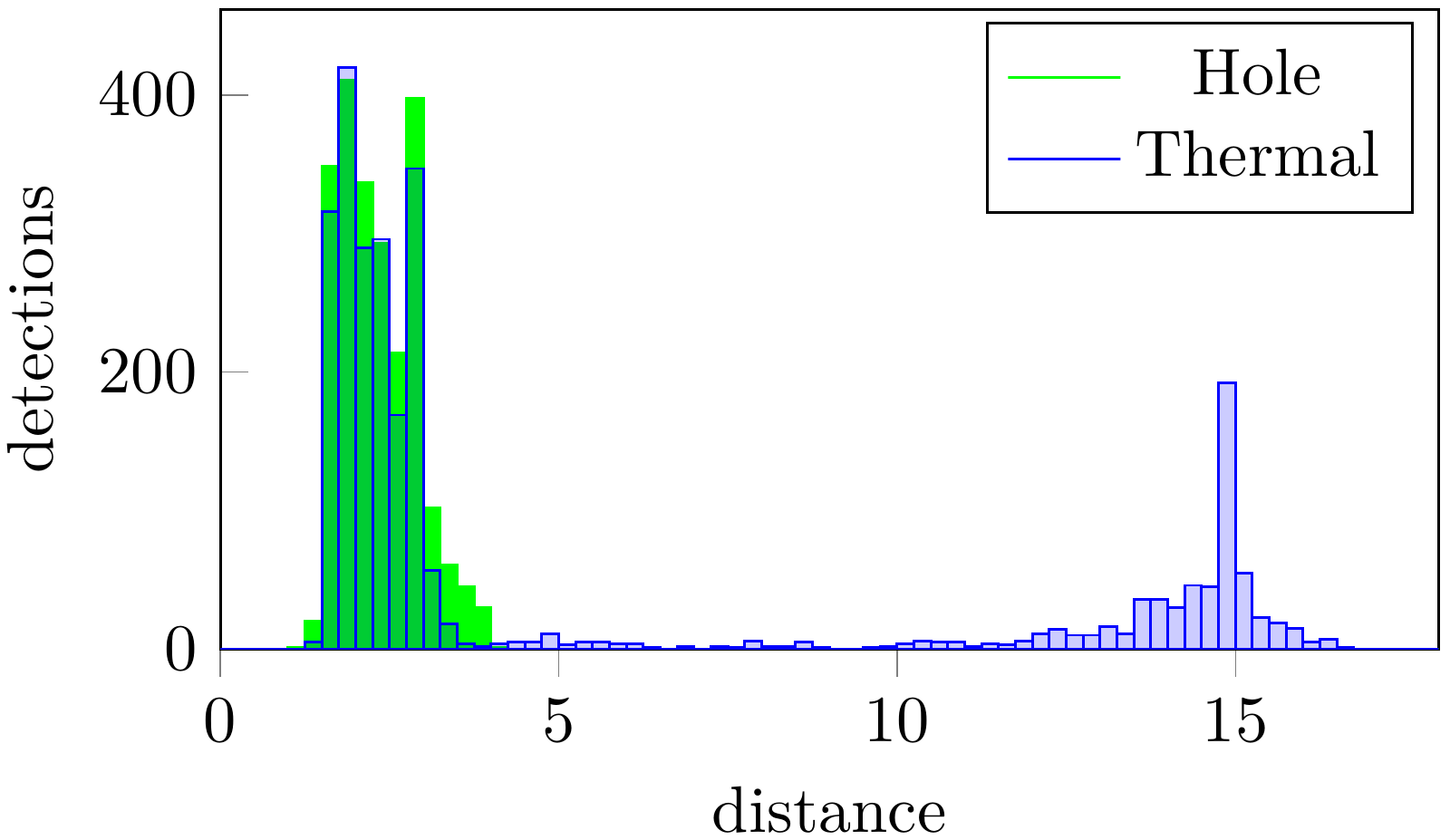}
\caption{MBZIRC Challenge Day~1 and 2.}
\label{fig:hist_det_mbzirc}
\end{subfigure}
\begin{subfigure}[t]{0.49\linewidth}
% \begin{tikzpicture}
% \begin{axis}[width=\linewidth, height=5cm,
%   name={theaxis},
%   xlabel=distance,
%   ylabel=detections,
%   ymin=0,
%   xmin=0,xmax=6,
%   tick pos=left, xtick align=outside
% ]
% \addplot [color=green,fill,hist={bins=20,data min=0,data max=6}] table {\holeHomeT};
% \addplot [color=blue,fill,fill opacity = 0.2, hist={bins=20,data min=0,data max=6}] table {\thermalHomeT};
% \addlegendentry{Hole}
% \addlegendentry{Thermal}
% \end{axis}
% \end{tikzpicture}
\includegraphics[width=1.0\linewidth]{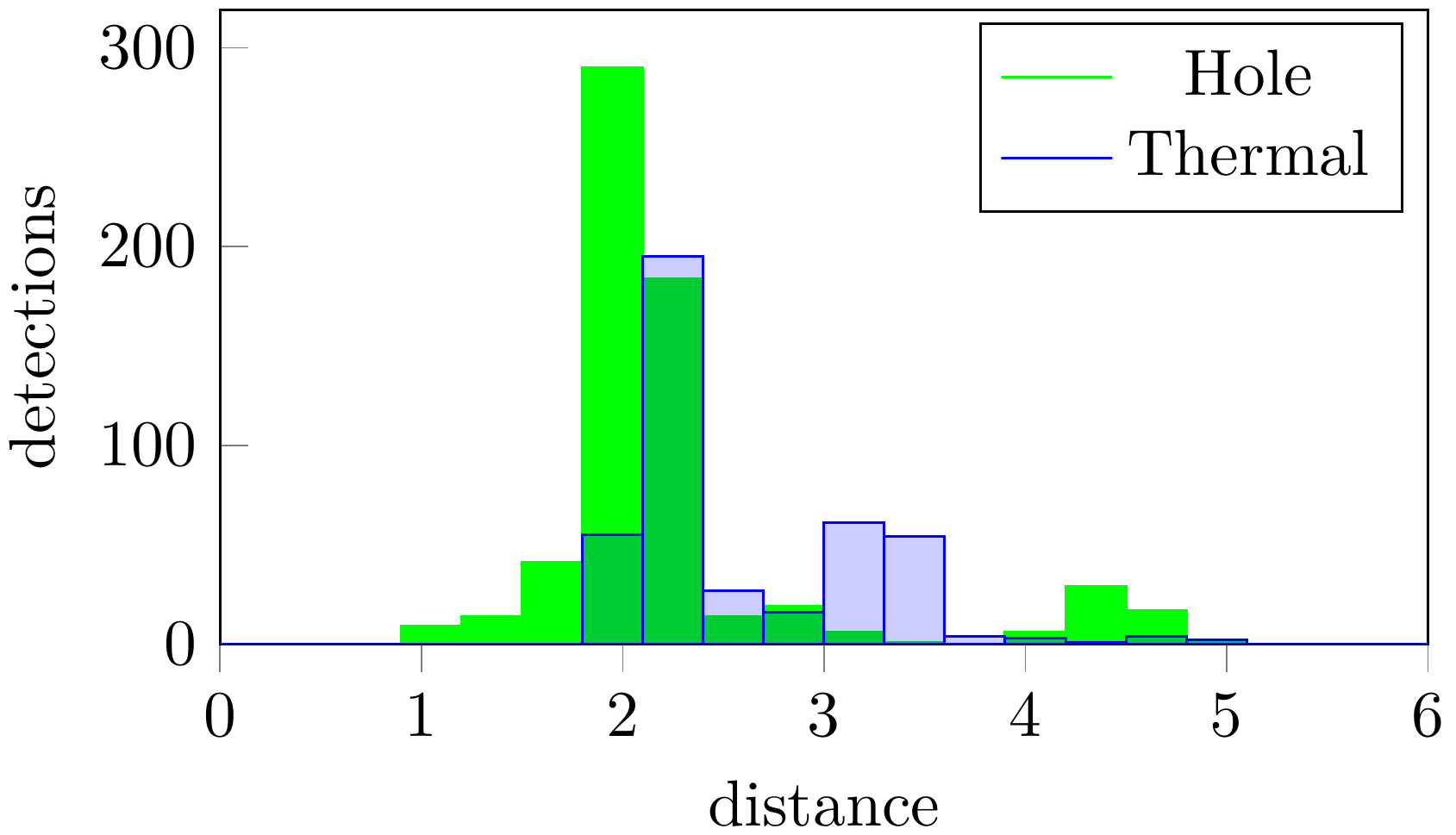}
\caption{Mockup trials.}
\label{fig:hist_det_home}
\end{subfigure}
\caption{Distance histogram for thermal and hole detections combined for MBZIRC Trial Days 1 and 2 (left) as well as the trials with the mockup target.}
\label{fig:hist_dets}
\end{figure}

\reffig{fig:hist_dets} compares the estimated distances of hole and thermal detections. We combined the trial runs on Day 1 and 2 of MBZIRC in \reffig{fig:hist_det_mbzirc}. The thermal detection provided first distance measures at up to 16.5\,m which is four times the maximal distance for the hole detection with 4.07\,m. In contrast, during our trials with the mockup target (\reffig{fig:uav_poppelsdorf}), both methods provided detections with a maximum distance of approx. 5\,m. We attribute the difference in behavior to the heat sources and sensor combination. The low thermal sensor resolution allowed us to detect the mockup's heating element at a similar range to the hole detection. During the challenge, the heat source was detectable further away due to the surrounding fire. The apparent difference in the absolute number of detections in \reffig{fig:hist_det_home} originates from the higher LiDAR scan frequency. Furthermore, we attribute for thermal detections the generally larger distance estimates at close range to measurements on the backside of the mockup.

\subsection{Autonomous Indoor Flight}
\label{sec:Autonomous_Indoor_Flight}
During the competition, we employed Splasher only for outdoor fire extinguishing and addressed indoor tasks using our UGV. This allowed us to use a combination of GPS and LiDAR localization.
However, we experienced severe localization errors due to limited GPS visibility and human errors. Although we later proved the viability of our approach on a mockup on an open field, we decided to increase the robustness of our localization method such that it can even be applied to challenging indoor scenarios. Thus, we developed a method for LiDAR-based odometry.

We model surfaces within LiDAR scans with normal distributions derived from measured points on a uniform sparse voxel grid. Our odometry \citep{quenzel2021iros} uses a sliding registration window to simultaneously register multiple surface element (surfel) maps against a local surfel map. A continuous-time Lie group B-Spline \citep{sommer2020cvpr} describes the UAV trajectory within the sliding registration window. After a certain traveled distance, we add the last scan in a keyframe-based sliding window approach to the local surfel map.

We exchanged our previous localization method with the new LiDAR-based odometry. An Extended Kalman Filter fuses the resulting position measurements with IMU data to generate estimates of the high-dimensional UAV state. Additionally, we use the method of \citet{schleich2021search} to plan a collision-free trajectory instead of manually defining waypoints.

\begin{figure}[t]
  \centering
  \begin{subfigure}[t]{0.45\linewidth}
  \centering
    \includegraphics[height=4cm]{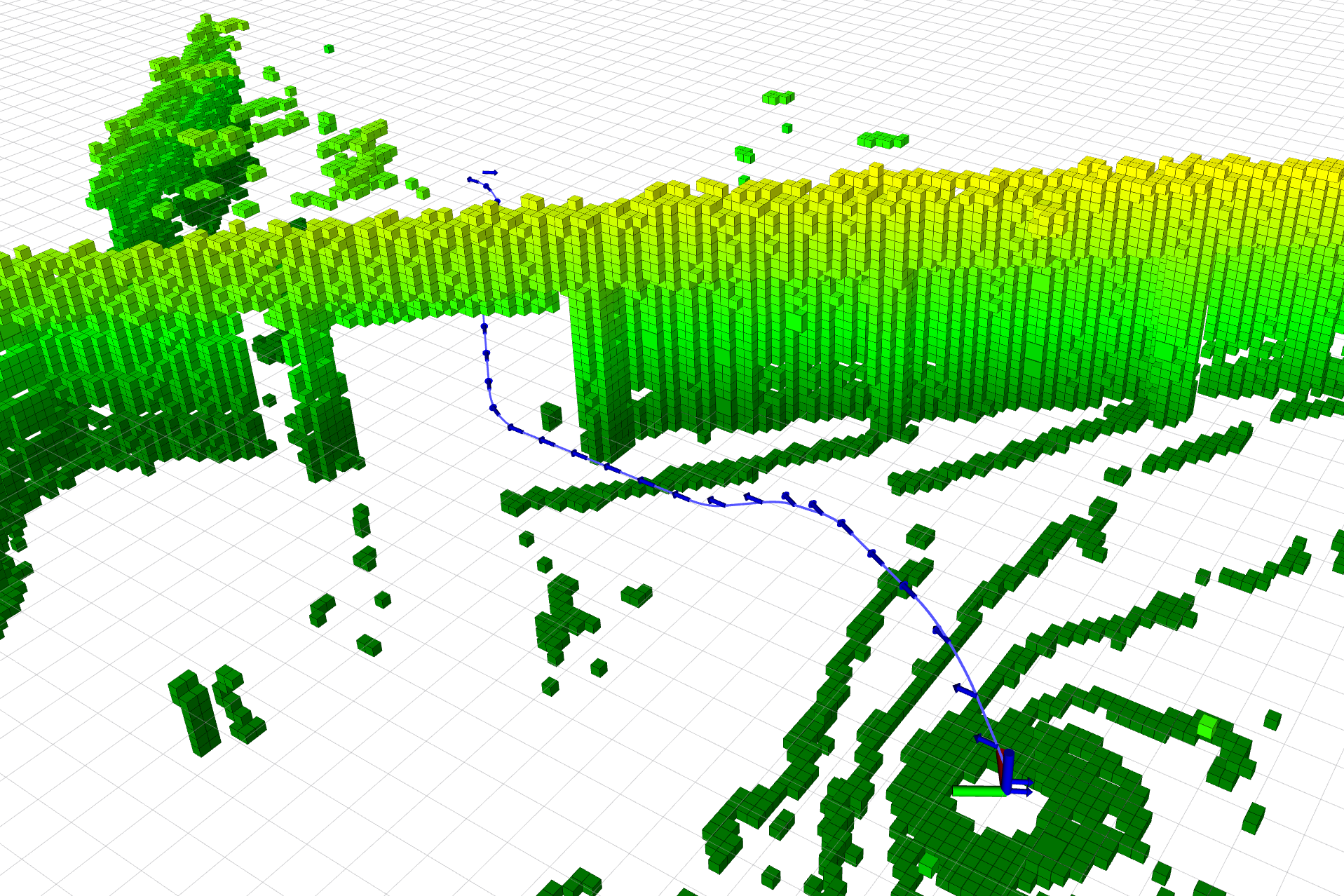}
    \caption{Initially planned trajectory.}
    \label{fig:lbh_plan}
  \end{subfigure}
  \begin{subfigure}[t]{0.45\linewidth}
  \centering
    \includegraphics[height=4cm]{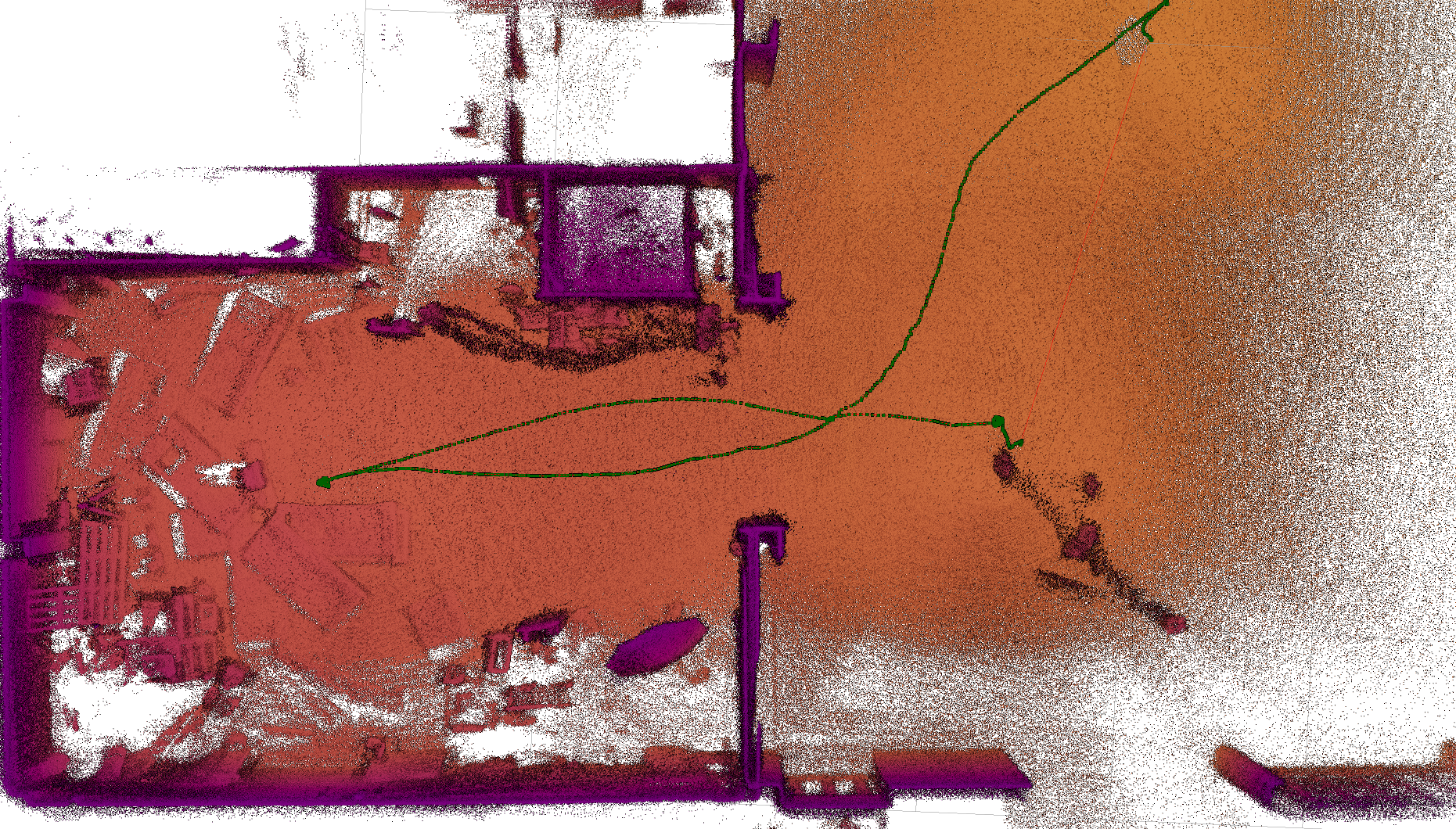}
    \caption{Top-down view of an aggregated point cloud and a trajectory (green) generated using our LiDAR-based odometry.}
    \label{fig:lbh_trajectory_top}
  \end{subfigure}
  \caption{Autonomous indoor flight experiment.}
  \label{fig:flight_lbh}
\end{figure}

\begin{figure}[t]
  \centering
  \begin{subfigure}[t]{0.45\linewidth}
  \centering
    \includegraphics[height=4cm]{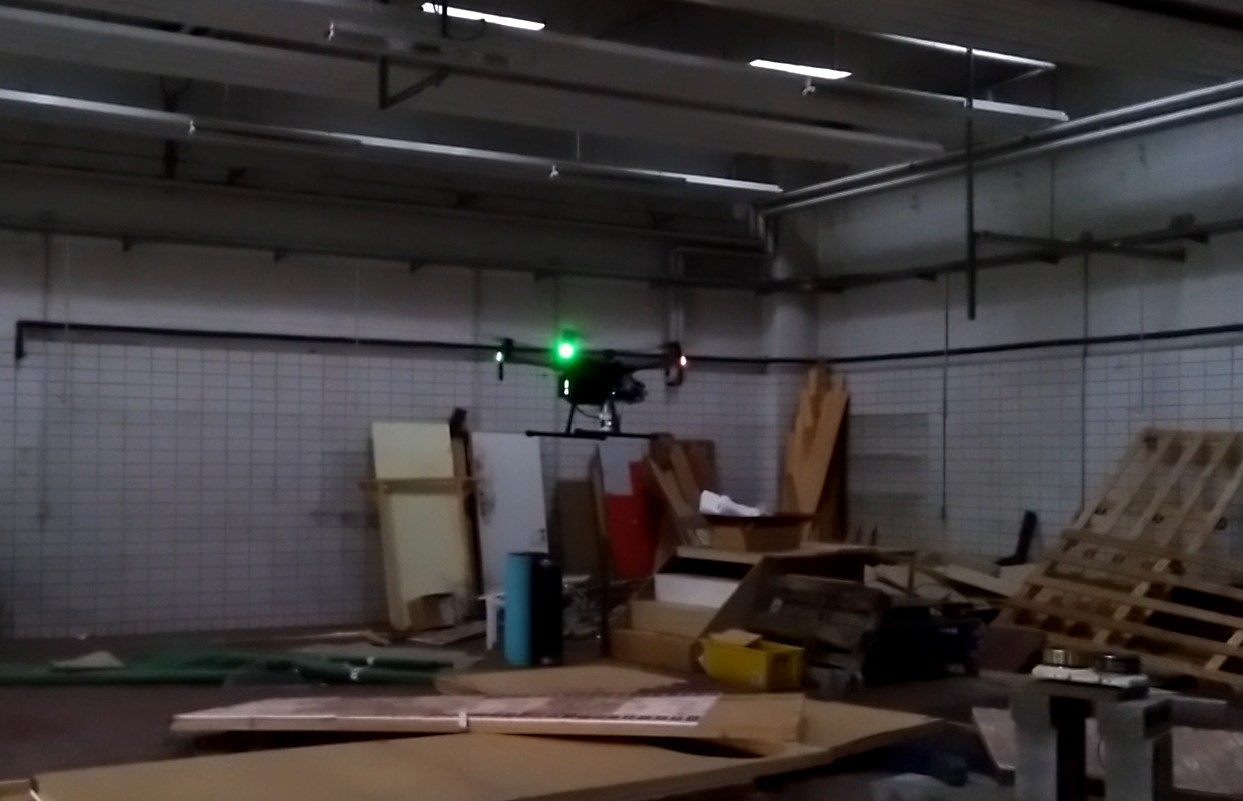}
    \caption{Our UAV at the indoor observation pose.}
  \label{fig:lbh_observation}
  \end{subfigure}
  \begin{subfigure}[t]{0.45\linewidth}
  \centering
    \includegraphics[height=4cm]{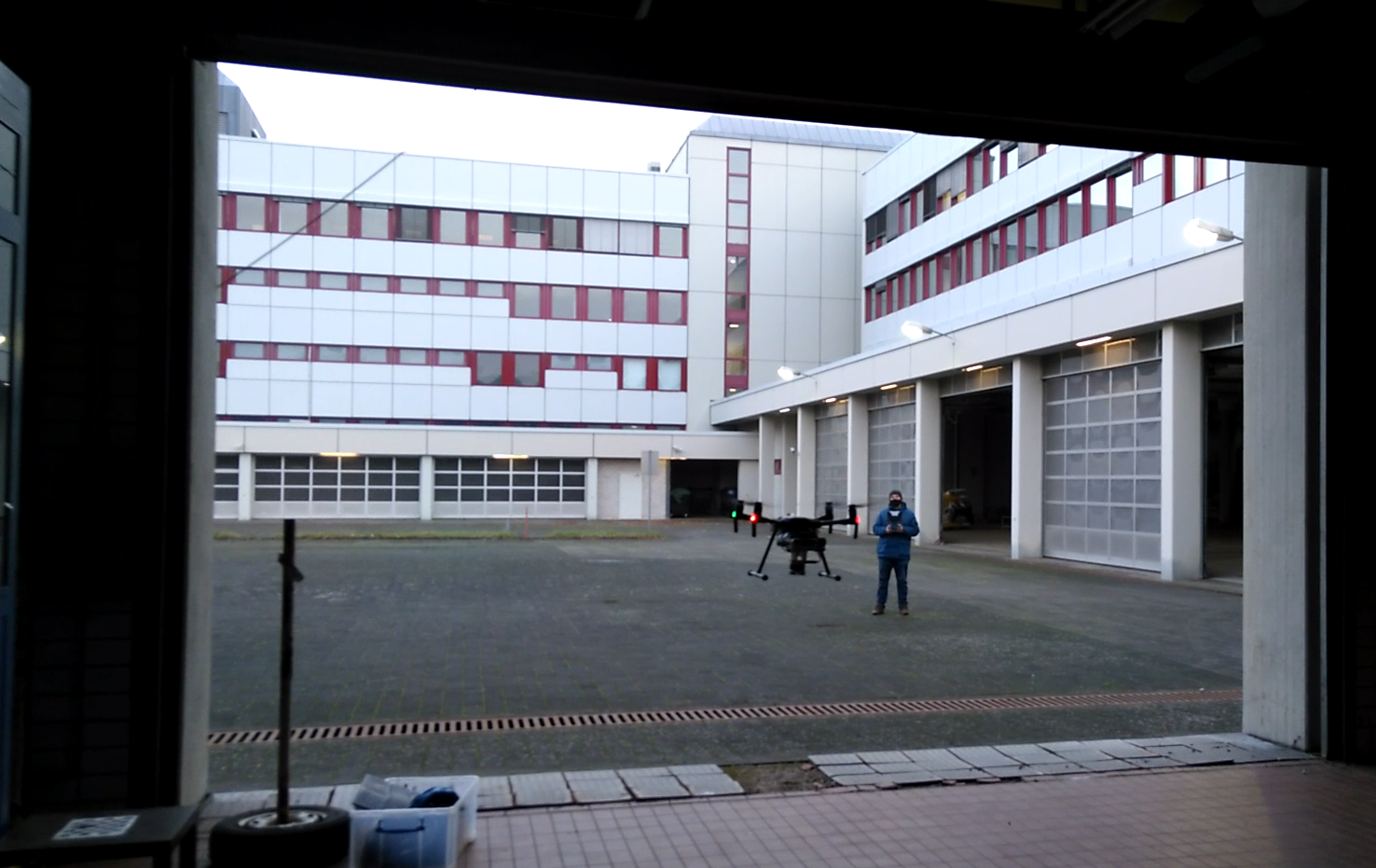}
    \caption{Our UAV autonomously exits the building while being supervised by a safety pilot.}
  \label{fig:lbh_exit}
  \end{subfigure}
  \caption{Our UAV during autonomous flight in a GNSS-denied environment.}
  \label{fig:img_lbh}
\end{figure}

We evaluated our updated system by autonomously exploring the inside of an industrial building. The UAV started outside and had to enter through a gate to reach a manually defined observation pose, as depicted in \reffig{fig:lbh_plan}. There, it rotated to generate an overview of the environment and left the building again. \reffig{fig:lbh_trajectory_top} shows an aggregated point cloud of the environment and a flight trajectory generated using our LiDAR-based odometry. Additional example images of our UAV during the flight are shown in \reffig{fig:img_lbh}.

\section{Lessons Learned}
\label{sec:Lessons_Learned}

\begin{table}
\small
\caption{Points (Ranks) at MBZIRC 2020 Grand Challenge.}
\label{tab:Scores_at_MBZIRC_2020_Grand_Challenge}
\vspace{-1.0em}
\begin{center}
\setlength{\tabcolsep}{1.9mm}
\begin{tabular}{lcccccc}
  \toprule
  Team                           & Ch.~1             & Ch.~2            & Ch.~3             & Time left   & Sum of Ranks & GC Rank    \\
  \midrule
  CTU Prague \& UPenn \& and NYU & \textbf{72.0 (1)} & 0.0 (8)          & 12.5   (2)        & 10          & \textbf{11}  & \textbf{1} \\
  Team NimbRo (Bonn)             & 30.0 (4)          & \textbf{0.0 (7)} & \textbf{12.5 (1)} & \textbf{20} & 12           & 2          \\
  UPM \& UPO \& PUT \& CNRS      & 40.5 (2)          & 0.0 (9)          & 0.1331 (5)        &  0          & 16           & 3          \\
  \bottomrule
\end{tabular}
\end{center}
\end{table}

To bring our performance in context, we depict the Grand Challenge scores of the best teams in \reftab{tab:Scores_at_MBZIRC_2020_Grand_Challenge}. One can see that no team was able to score in the wall-building challenge\footnote{Only one of 17 teams was able to score in this subchallenge.}. The rank was determined as a sum of the individual challenge ranks. While in Challenge~1, no tiebreaker was needed, our rank (7) in Challenge~2 and Challenge~3 (1) was mainly determined by the time left.

We take the opportunity to identify the key strengths and weaknesses of our system and development approach. We also want to identify aspects of the competition that could be improved to increase scientific usefulness in the future.

Our hardware designs proved themselves during the competition after small adaptions to the conditions at hand. For example, after solving initial problems with our magnets, especially the passive gripper turned out to be an advantage over other teams, who could not manipulate the heavier bricks.

The biggest issue shortly before and during the competition was unavailable testing time. Robust solutions require full-stack testing under competition constraints. Since we postponed many design decisions until the competition rules were settled, we could not test our intricate design thoroughly. In hindsight, simpler designs with fewer components, which would have required less thorough testing, could have been more successful in the short available time frame.

While we tested our UAVs on small mockups in our lab and on an open field, the competition environment was very different and led to system failures. For example, GPS visibility was severely limited near the mandatory start position, which led to initialization errors. An initialization-free or delayed initialization scheme, which we implemented later, or even a GPS drift detection would have improved robustness. Furthermore, relative navigation enables task fulfillment with unreliable pose information, and low-level obstacle avoidance is mandatory in such a scenario.

Also, improved visualization of the UAV state and perception could have helped to detect problems early on. This includes a thorough application of our supervision system for every component to catch errors like the unintentionally disabled laser localization subsystem.

The competition also placed an enormous strain on the involved personnel. For safe operation, one safety pilot was required per UAV, plus at least one team member supervising the internal state of the autonomous systems and being able to reconfigure the system during resets. For example, in the Grand Challenge, this led to the situation that a safety pilot had to run from one arena to another depending on which subchallenge was active, unnecessarily delaying the run. While reducing the number of required human operators per robot is an admirable research goal, this is not possible in the near future due to safety regulations. We thus feel that future competitions should keep the required human personnel in mind so that small teams can continue to participate.

What proved to be useful in the context of competitions is to prepare low-effort backup strategies in advance, like assistance functions for manual mode. Also, reducing high-level control to a bare minimum, instead of universal strategies, makes the systems easier to test and reduces potential errors.

For example, in the balloon hunting challenge, our approach could not incorporate the arena's non-convex shape. Instead of using an untested elaborated path planning approach, we used the low-effort strategy of inserting an intermediate waypoint in the middle of the arena, which required minimal testing.

This edition of the MBZIRC suffered from overall low team performance, to the extent that the Grand Challenge price money was not paid out on the jury's recommendation. This underperformance of all teams points to systematic issues with the competition. From a participants' perspective, we think late changes to the rules have certainly contributed to this situation. A pre-competition event such as the Testbed in the DARPA Robotics Challenge can help identify critical issues with rules and material early in the competition timeline.

Another issue was the required effort to participate in all the different sub-challenges. The MBZIRC 2020 defined seven different tasks. Ideally, one would develop specialized solutions for all of these. Focusing the competition more on general usability, \ie defining multiple tasks that can \emph{and should} be completed by one platform would lower the barrier for participants.

\section{Conclusion}
\label{sec:Conclusion}

We demonstrated successful hardware design, perception and control methods, high-level control, and system integration for a highly complex robotic challenge.
The lessons learned discussed above in \cref{sec:Lessons_Learned} already contain hints about individual improvements that
can be applied to sub-components, team strategy, or competition organization.

On a more global scale, we firmly believe that robotic competitions such as MBZIRC are key for our community, as they
force researchers to test their algorithms under real-world, integrated conditions. Public benchmarks are the only
way to properly evaluate systems in this direction.
Contrary to comparable challenges, MBZIRC provides very varied and complex challenges.
While this is highly interesting and provides research challenges, it would be good to have similar tasks
from one competition edition to the next to give the community time to learn from their promising approaches.

Finally, while we made technical contributions to various aspects in each sub-challenge, it is clear that all of them warrant further research to increase flexibility and applicability to other domains.

\subsubsection*{Acknowledgments}
We thank all members of our team NimbRo for their support before and during the competition. This work has been supported by a grant from the Mohamed Bin Zayed International Robotics Challenge (MBZIRC).

\bibliographystyle{apalike}

\begin{thebibliography}{}

\bibitem[Ando et~al., 2018]{Ando2018ral}
Ando, H., Ambe, Y., Ishii, A., Konyo, M., Tadakuma, K., Maruyama, S., and
  Tadokoro, S. (2018).
\newblock Aerial hose type robot by water jet for fire fighting.
\newblock {\em IEEE Robotics and Automation Letters (RA-L)}, 3(2):1128--1135.

\bibitem[Aydin et~al., 2019]{Aydin2019drones}
Aydin, B., Selvi, E., Tao, J., and Starek, M.~J. (2019).
\newblock Use of fire-extinguishing balls for a conceptual system of
  drone-assisted wildfire fighting.
\newblock {\em Drones}, 3(1):17.

\bibitem[Baca et~al., 2020]{baca2020arxiv}
Baca, T., Penicka, R., Stepan, P., Petrlik, M., Spurny, V., Hert, D., and
  Saska, M. (2020).
\newblock Autonomous cooperative wall building by a team of unmanned aerial
  vehicles in the mbzirc 2020 competition.
\newblock {\em arXiv e-prints}, arXiv:2012.05946.

\bibitem[Baca et~al., 2017]{Baca2017}
Baca, T., Stepan, P., and Saska, M. (2017).
\newblock Autonomous landing on a moving car with unmanned aerial vehicle.
\newblock In {\em Proc. of European Conf. on Mobile Robots (ECMR)}.

\bibitem[B{\"a}hnemann et~al., 2019]{bahnemann2019eth}
B{\"a}hnemann, R., Pantic, M., Popovi{\'c}, M., Schindler, D., Tranzatto, M.,
  Kamel, M., Grimm, M., Widauer, J., Siegwart, R., and Nieto, J. (2019).
\newblock The {ETH-MAV} team in the {MBZ} international robotics challenge.
\newblock {\em Jnl. of Field Robotics}, 36(1):78--103.

\bibitem[Bailon-Ruiz and Lacroix, 2020]{BailonRuiz2020icuas}
Bailon-Ruiz, R. and Lacroix, S. (2020).
\newblock Wildfire remote sensing with {UAVs}: A review from the autonomy point
  of view.
\newblock In {\em Proc. of Int. Conf. on Unmanned Aircraft Systems (ICUAS)}.

\bibitem[Battiato et~al., 2017]{Battiato2017}
Battiato, S., Cantelli, L., D'Urso, F., Farinella, G.~M., Guarnera, L.,
  Guastella, D., Melita, C.~D., Muscato, G., Ortis, A., Ragusa, F., and
  Santoro, C. (2017).
\newblock A system for autonomous landing of a {UAV} on a moving vehicle.
\newblock In {\em Image Analysis and Processing (ICIAP)}.

\bibitem[Beul and Behnke, 2016]{beul2016icuas}
Beul, M. and Behnke, S. (2016).
\newblock Analytical time-optimal trajectory generation and control for
  multirotors.
\newblock In {\em Proc. of Int. Conf. on Unmanned Aircraft Systems (ICUAS)}.

\bibitem[Beul and Behnke, 2017]{beul2017icuas}
Beul, M. and Behnke, S. (2017).
\newblock Fast full state trajectory generation for multirotors.
\newblock In {\em Proc. of Int. Conf. on Unmanned Aircraft Systems (ICUAS)}.

\bibitem[Beul et~al., 2020]{beul2020ssrr_mbzirc}
Beul, M., Bultmann, S., Rochow, A., Rosu, R.~A., Schleich, D., Splietker, M.,
  and Behnke, S. (2020).
\newblock Visually guided balloon popping with an autonomous {MAV} at {MBZIRC}
  2020.
\newblock In {\em Proceedings of the IEEE International Symposium on Safety,
  Security and Rescue Robotics (SSRR)}.

\bibitem[Beul et~al., 2017]{ecmr2017_c1}
Beul, M., Houben, S., Nieuwenhuisen, M., and Behnke, S. (2017).
\newblock Fast autonomous landing on a moving target at {MBZIRC}.
\newblock In {\em Proc. of European Conf. on Mobile Robots (ECMR)}.

\bibitem[Beul et~al., 2019]{beul2019team}
Beul, M., Nieuwenhuisen, M., Quenzel, J., Rosu, R.~A., Horn, J., Pavlichenko,
  D., Houben, S., and Behnke, S. (2019).
\newblock Team {NimbRo} at {MBZIRC} 2017: Fast landing on a moving target and
  treasure hunting with a team of micro aerial vehicles.
\newblock {\em Jnl. Field Rob.}, 36(1).

\bibitem[Borrmann et~al., 2013]{borrmann2013thermal}
Borrmann, D., Elseberg, J., and N{\"u}chter, A. (2013).
\newblock Thermal {3D} mapping of building fa{\c{c}}ades.
\newblock In {\em Proceedings of Int. Conf. on Intelligent Autonomous Systems
  (IAS)}, pages 173--182. Springer.

\bibitem[Cantelli et~al., 2017]{Cantelli2017}
Cantelli, L., Guastella, D., Melita, C.~D., Muscato, G., Battiato, S., D'Urso,
  F., Farinella, G.~M., Ortis, A., and Santoro, C. (2017).
\newblock Autonomous landing of a {UAV} on a moving vehicle for the {MBZIRC}.
\newblock In {\em Proceedings of the 20th International Conference on Climbing
  and Walking Robots and the Support Technologies for Mobile Machines
  (CLAWAR)}.

\bibitem[Carlson et~al., 2019]{inbook}
Carlson, A., Skinner, K., Vasudevan, R., and Johnson-Roberson, M. (2019).
\newblock {\em Modeling Camera Effects to Improve Visual Learning from
  Synthetic Data}, pages 505--520.

\bibitem[Cho et~al., 2015]{cho2015survey}
Cho, Y.~K., Ham, Y., and Golpavar-Fard, M. (2015).
\newblock {3D} as-is building energy modeling and diagnostics: A review of the
  state-of-the-art.
\newblock {\em Advanced Engineering Informatics}, 29(2):184 -- 195.

\bibitem[Delmerico et~al., 2019]{Delmerico2019jfr}
Delmerico, J., Mintchev, S., Giusti, A., Gromov, B., Melo, K., Horvat, T.,
  Cadena, C., Hutter, M., Ijspeert, A., Floreano, D., et~al. (2019).
\newblock The current state and future outlook of rescue robotics.
\newblock {\em J. of Field Robotics}, 36(7):1171--1191.

\bibitem[Demisse et~al., 2015]{demisse2015interpreting}
Demisse, G., Borrmann, D., and N{\"u}chter, A. (2015).
\newblock Interpreting thermal {3D} models of indoor environments for energy
  efficiency.
\newblock {\em J. Intell. Robot. Syst.}, 77(1):55--72.

\bibitem[Droeschel and Behnke, 2018]{DavidCTSLAM}
Droeschel, D. and Behnke, S. (2018).
\newblock Efficient continuous-time {SLAM} for {3D} lidar-based online mapping.
\newblock In {\em Proceedings of IEEE Int. Conf. on Robotics and Automation
  (ICRA)}.

\bibitem[Everingham et~al., 2010]{pascalvoc_2010}
Everingham, M., Van~Gool, L., Williams, C. K.~I., Winn, J., and Zisserman, A.
  (2010).
\newblock The {Pascal} {Visual} {Object} {Classes} ({VOC}) {Challenge}.
\newblock {\em Int J Comput Vis}, 88(2):303--338.

\bibitem[Ezair et~al., 2014]{Ezair2014}
Ezair, B., Tassa, T., and Shiller, Z. (2014).
\newblock Planning high order trajectories with general initial and final
  conditions and asymmetric bounds.
\newblock {\em The International Journal of Robotics Research}, 33(6):898--916.

\bibitem[Falanga et~al., 2017]{Falanga_SSRR2017}
Falanga, D., Zanchettin, A., Simovic, A., Delmerico, J., and Scaramuzza, D.
  (2017).
\newblock Vision-based autonomous quadrotor landing on a moving platform.
\newblock In {\em Proceedings of the IEEE International Symposium on Safety,
  Security and Rescue Robotics (SSRR)}.

\bibitem[{Fritsche} et~al., 2017]{fritsche2017fusion}
{Fritsche}, P., {Zeise}, B., {Hemme}, P., and {Wagner}, B. (2017).
\newblock Fusion of radar, {LiDAR} and thermal information for hazard detection
  in low visibility environments.
\newblock In {\em Proceedings of the IEEE International Symposium on Safety,
  Security and Rescue Robotics (SSRR)}, pages 96--101.

\bibitem[Ghamry et~al., 2016]{Ghamry2016icuas}
Ghamry, K.~A., Kamel, M.~A., and Zhang, Y. (2016).
\newblock Cooperative forest monitoring and fire detection using a team of
  {UAVs-UGVs}.
\newblock In {\em Proc. of Int. Conf. on Unmanned Aircraft Systems (ICUAS)},
  pages 1206--1211. IEEE.

\bibitem[Goessens et~al., 2018]{goessens2018feasibility}
Goessens, S., Mueller, C., and Latteur, P. (2018).
\newblock Feasibility study for drone-based masonry construction of real-scale
  structures.
\newblock {\em Automation in Construction}, 94.

\bibitem[He et~al., 2016]{he_deep_2016}
He, K., Zhang, X., Ren, S., and Sun, J. (2016).
\newblock Deep residual learning for image recognition.
\newblock In {\em {IEEE} {Conference} on {Computer} {Vision} and {Pattern}
  {Recognition} ({CVPR})}, pages 770--778.

\bibitem[Howard et~al., 2017]{mobilenet_2017}
Howard, A.~G., Zhu, M., Chen, B., Kalenichenko, D., Wang, W., Weyand, T.,
  Andreetto, M., and Adam, H. (2017).
\newblock {MobileNets: Efficient Convolutional Neural Networks for Mobile
  Vision Applications}.
\newblock {\em {CoRR}}, abs/1704.04861.

\bibitem[Huber et~al., 2013]{huber2013first}
Huber, F., Kondak, K., Krieger, K., Sommer, D., Schwarzbach, M., Laiacker, M.,
  Kossyk, I., Parusel, S., Haddadin, S., and Albu-Sch{\"a}ffer, A. (2013).
\newblock First analysis and experiments in aerial manipulation using fully
  actuated redundant robot arm.
\newblock In {\em Proceedings of IEEE/RSJ Int. Conf. on Intelligent Robots and
  Systems (IROS)}.

\bibitem[Jacob et~al., 2018]{quantization_2018}
Jacob, B., Kligys, S., Chen, B., Zhu, M., Tang, M., Howard, A., Adam, H., and
  Kalenichenko, D. (2018).
\newblock Quantization and training of neural networks for efficient
  integer-arithmetic-only inference.
\newblock In {\em {IEEE}/{CVF} {Conference} on {Computer} {Vision} and
  {Pattern} {Recognition} (CVPR)}, pages 2704--2713.

\bibitem[Jindal et~al., 2021]{jindal2021design}
Jindal, K., Wang, A., Thakur, D., Zhou, A., Spurny, V., Walter, V., Broughton,
  G., Krajnik, T., Saska, M., and Loianno, G. (2021).
\newblock Design and deployment of an autonomous unmanned ground vehicle for
  urban firefighting scenarios.
\newblock {\em arXiv e-prints}, arXiv:2107.03582.

\bibitem[Juh{\'a}sz et~al., 2008]{Juhasz2008}
Juh{\'a}sz, Z., Sipos, {\'A}., and Porkol{\'a}b, Z. (2008).
\newblock {\em Implementation of a Finite State Machine with Active Libraries
  in C++}, pages 474--488.
\newblock Springer Berlin Heidelberg, Berlin, Heidelberg.

\bibitem[Kim et~al., 2013]{kim2013aerial}
Kim, S., Choi, S., and Kim, H.~J. (2013).
\newblock Aerial manipulation using a quadrotor with a two {DOF} robotic arm.
\newblock In {\em Proceedings of IEEE/RSJ Int. Conf. on Intelligent Robots and
  Systems (IROS)}.

\bibitem[Krizmancic et~al., 2020]{krizmancic2020cooperative}
Krizmancic, M., Arbanas, B., Petrovic, T., Petric, F., and Bogdan, S. (2020).
\newblock Cooperative aerial-ground multi-robot system for automated
  construction tasks.
\newblock {\em IEEE Robotics and Automation Letters}, 5(2):798--805.

\bibitem[Lenz et~al., 2021]{lenz2021jfr}
Lenz, C., Quenzel, J., Periyasamy, A.~S., Razlaw, J., Rochow, A., Splietker,
  M., Schreiber, M., Schwarz, M., Süberkrüb, F., and Behnke, S. (2021).
\newblock Autonomous wall-building and firefighting: Team {NimbRo's} {UGV}
  solution for {MBZIRC} 2020.
\newblock Accepted for Field Robotics.

\bibitem[Lenz et~al., 2020]{lenz2020ssrr}
Lenz, C., Schwarz, M., Rochow, A., Razlaw, J., Periyasamy, A.~S., Schreiber,
  M., and Behnke, S. (2020).
\newblock Autonomous wall building with a {UGV-UAV} team at {MBZIRC} 2020.
\newblock In {\em Proceedings of the IEEE International Symposium on Safety,
  Security and Rescue Robotics (SSRR)}.

\bibitem[Lindsey et~al., 2012]{lindsey2012construction}
Lindsey, Q., Mellinger, D., and Kumar, V. (2012).
\newblock Construction with quadrotor teams.
\newblock {\em Autonomous Robots}, 33(3):323--336.

\bibitem[Liu et~al., 2016a]{Liu2016icac}
Liu, P., Yu, H., Cang, S., and Vladareanu, L. (2016a).
\newblock Robot-assisted smart firefighting and interdisciplinary perspectives.
\newblock In {\em Proceedings of the International Conference on Automation and
  Computing (ICAC)}, pages 395--401. IEEE.

\bibitem[Liu et~al., 2016b]{liu2016ssd}
Liu, W., Anguelov, D., Erhan, D., Szegedy, C., Reed, S., Fu, C.-Y., and Berg,
  A.~C. (2016b).
\newblock Ssd: Single shot multibox detector.
\newblock In {\em European conference on computer vision}, pages 21--37.
  Springer.

\bibitem[Loianno et~al., 2018]{8276269}
Loianno, G., Spurny, V., Thomas, J., Baca, T., Thakur, D., Hert, D., Penicka,
  R., Krajnik, T., Zhou, A., Cho, A., Saska, M., and Kumar, V. (2018).
\newblock Localization, grasping, and transportation of magnetic objects by a
  team of {MAV}s in challenging desert-like environments.
\newblock {\em IEEE Robotics and Automation Letters}, 3(3):1576--1583.

\bibitem[Maninis et~al., 2018]{conv_boundaries_2018}
Maninis, K.-K., Pont-Tuset, J., Arbel\'{a}ez, P., and Gool, L.~V. (2018).
\newblock {Convolutional Oriented Boundaries: From Image Segmentation to
  High-Level Tasks}.
\newblock {\em {IEEE Transactions on Pattern Analysis and Machine Intelligence
  (TPAMI)}}, 40(4):819 -- 833.

\bibitem[Michael et~al., 2011]{Cooperative_Manipulation}
Michael, N., Fink, J., and Kumar, V. (2011).
\newblock Cooperative manipulation and transportation with aerial robots.
\newblock In {\em Autonomous Robots}, volume~30, pages 73--86.

\bibitem[Michael et~al., 2012]{https://doi.org/10.1002/rob.21436}
Michael, N., Shen, S., Mohta, K., Mulgaonkar, Y., Kumar, V., Nagatani, K.,
  Okada, Y., Kiribayashi, S., Otake, K., Yoshida, K., Ohno, K., Takeuchi, E.,
  and Tadokoro, S. (2012).
\newblock Collaborative mapping of an earthquake-damaged building via ground
  and aerial robots.
\newblock {\em J. of Field Robotics}, 29(5):832--841.

\bibitem[Moore and Stouch, 2014]{moore2014ekf}
Moore, T. and Stouch, D. (2014).
\newblock A generalized extended {Kalman} filter implementation for the robot
  operating system.
\newblock In {\em Proceedings of Int. Conf. on Intelligent Autonomous Systems
  (IAS)}. Springer.

\bibitem[Nieuwenhuisen et~al., 2017]{ecmr2017_c3}
Nieuwenhuisen, M., Beul, M., Rosu, R.~A., Quenzel, J., Pavlichenko, D., Houben,
  S., and Behnke, S. (2017).
\newblock Collaborative object picking and delivery with a team of micro aerial
  vehicles at {MBZIRC}.
\newblock In {\em Proc. of European Conf. on Mobile Robots (ECMR)}.

\bibitem[Quenzel and Behnke, 2021]{quenzel2021iros}
Quenzel, J. and Behnke, S. (2021).
\newblock Real-time multi-adaptive-resolution-surfel {6D LiDAR} odometry using
  continuous-time trajectory optimization.
\newblock In {\em Proceedings of IEEE/RSJ Int. Conf. on Intelligent Robots and
  Systems (IROS)}.

\bibitem[Quigley et~al., 2009]{ROS}
Quigley, M., Conley, K., Gerkey, B.~P., Faust, J., Foote, T., Leibs, J.,
  Wheeler, R., and Ng, A.~Y. (2009).
\newblock {ROS}: An open-source robot operating system.
\newblock In {\em ICRA Workshop on Open Source Software}.

\bibitem[Real et~al., 2021]{real2021access}
Real, F., \'{A}ngel R.~Casta\~{n}o, Torres-Gonz\'{a}lez, A., Capit\'{a}n, J.,
  Sanchez-Cuevas, P.~J., Fernandez, M.~J., Villar, M., and Ollero, A. (2021).
\newblock Experimental evaluation of a team of multiple unmanned aerial
  vehicles for cooperative construction.
\newblock {\em IEEE Access}, 9:6817--6835.

\bibitem[Rodriguez et~al., 2019]{RoboCup_2019}
Rodriguez, D., Farazi, H., Ficht, G., Pavlichenko, D., Brandenburger, A.,
  Hosseini, M., Kosenko, O., Schreiber, M., Missura, M., and Behnke, S. (2019).
\newblock {RoboCup} 2019 {AdultSize} winner {NimbRo}: Deep learning perception,
  in-walk kick, push recovery, and team play capabilities.
\newblock {\em RoboCup 2019: Robot World Cup XXIII}, pages 631--645.

\bibitem[Romero-Ramirez et~al., 2018]{romero2018speeded}
Romero-Ramirez, F.~J., Mu{\~n}oz-Salinas, R., and Medina-Carnicer, R. (2018).
\newblock Speeded up detection of squared fiducial markers.
\newblock {\em Image and Vision Computing}, 76:38--47.

\bibitem[Rosu and Behnke, 2020]{rosu2021grapp}
Rosu, R.~A. and Behnke, S. (2020).
\newblock Easypbr: A lightweight physically-based renderer.
\newblock In {\em Proceedings of the International Conference on Computer
  Graphics Theory and Applications (GRAPP)}.

\bibitem[Rosu et~al., 2019]{Rosu2019ssrr}
Rosu, R.~A., Quenzel, J., and Behnke, S. (2019).
\newblock Reconstruction of textured meshes for fire and heat source detection.
\newblock In {\em Proceedings of the IEEE International Symposium on Safety,
  Security and Rescue Robotics (SSRR)}, pages 235--242. IEEE.

\bibitem[Ruggiero et~al., 2018]{ruggiero2018aerial}
Ruggiero, F., Lippiello, V., and Ollero, A. (2018).
\newblock Aerial manipulation: A literature review.
\newblock {\em Rob. and Automation Letters}, 3(3).

\bibitem[Sandler et~al., 2018]{mobilenetv2_2018}
Sandler, M., Howard, A., Zhu, M., Zhmoginov, A., and Chen, L.-C. (2018).
\newblock {MobileNetV2}: {Inverted} {Residuals} and {Linear} {Bottlenecks}.
\newblock In {\em {IEEE}/{CVF} {Conference} on {Computer} {Vision} and
  {Pattern} {Recognition} (CVPR)}, pages 4510--4520.

\bibitem[Schleich and Behnke, 2021]{schleich2021search}
Schleich, D. and Behnke, S. (2021).
\newblock Search-based planning of dynamic {MAV} trajectories using local
  multiresolution state lattices.
\newblock In {\em Proceedings of IEEE Int. Conf. on Robotics and Automation
  (ICRA)}.

\bibitem[Sch{\"o}nauer et~al., 2013]{schonauer2013live}
Sch{\"o}nauer, C., Vonach, E., Gerstweiler, G., and Kaufmann, H. (2013).
\newblock {3D} building reconstruction and thermal mapping in fire brigade
  operations.
\newblock In {\em Proceedings of the 4th Augmented Human International
  Conference}, pages 202--205. ACM.

\bibitem[Sommer et~al., 2020]{sommer2020cvpr}
Sommer, C., Usenko, V., Schubert, D., Demmel, N., and Cremers, D. (2020).
\newblock Efficient derivative computation for cumulative {B-Splines} on {Lie}
  groups.
\newblock In {\em Proceedings of the IEEE Conference on Computer Vision and
  Pattern Recognition (CVPR)}.

\bibitem[Spurn{\`y} et~al., 2019]{spurny2019cooperative}
Spurn{\`y}, V., B{\'a}{\v{c}}a, T., Saska, M., P{\v{e}}ni{\v{c}}ka, R.,
  Krajn{\'\i}k, T., Thomas, J., Thakur, D., Loianno, G., and Kumar, V. (2019).
\newblock Cooperative autonomous search, grasping, and delivering in a treasure
  hunt scenario by a team of {UAVs}.
\newblock {\em Jnl. Field Rob.}, 36(1).

\bibitem[Spurny et~al., 2021]{spurny2021access}
Spurny, V., Pritzl, V., Walter, V., Petrlik, M., Baca, T., Stepan, P., Zaitlik,
  D., and Saska, M. (2021).
\newblock Autonomous firefighting inside buildings by an unmanned aerial
  vehicle.
\newblock {\em IEEE Access}, 9:15872--15890.

\bibitem[Suarez~Fernandez et~al., 2020]{SkyeyeMBZIRC}
Suarez~Fernandez, R., Rodríguez~Ramos, A., Alvarez, A., Rodríguez-Vázquez,
  J., Bavle, H., Lu, L., Fernandez, M., Rodelgo, A., Cobano, A., Alejo, D.,
  Acedo, D., Rey, R., Martinez-Rozas, S., Molina, M., Merino, L., Caballero,
  F., and Campoy, P. (2020).
\newblock The {SkyEye} team participation in the 2020 {Mohamed Bin Zayed
  International Robotics Challenge}.
\newblock In {\em Mohamed Bin Zayed International Robotics Competition (MBZIRC)
  Symposium}.

\bibitem[Tran et~al., 2021]{9124698}
Tran, V.~P., Santoso, F., Garratt, M.~A., and Anavatti, S.~G. (2021).
\newblock Distributed artificial neural networks-based adaptive strictly
  negative imaginary formation controllers for unmanned aerial vehicles in
  time-varying environments.
\newblock {\em IEEE Transactions on Industrial Informatics}, 17(6):3910--3919.

\bibitem[Yang et~al., 2016]{object_contour_2016}
Yang, J., Price, B., Cohen, S., Lee, H., and Yang, M.-H. (2016).
\newblock Object {Contour} {Detection} with a {Fully} {Convolutional}
  {Encoder}-{Decoder} {Network}.
\newblock In {\em 2016 {IEEE} {Conference} on {Computer} {Vision} and {Pattern}
  {Recognition} ({CVPR})}, pages 193--202, Las Vegas, NV, USA. IEEE.

\end{thebibliography}

\end{document}